\documentclass[11pt]{article}

\usepackage{geometry}
\usepackage{setspace}
\usepackage{amsmath, amssymb, amsfonts, bm, mathtools}
\usepackage{amsthm}
\usepackage[dvipsnames]{xcolor}
\definecolor{darkblue}{rgb}{0,0,.5}
\usepackage{graphicx}
\usepackage{subfigure}

\usepackage[numbers]{natbib}

\usepackage[colorlinks=true,allcolors=darkblue]{hyperref}       
\usepackage{url}            
\usepackage{booktabs}       
\usepackage{amsfonts}       
\usepackage{nicefrac}       
\usepackage{microtype}      

\allowdisplaybreaks

\usepackage{float}
\usepackage{multirow}
\usepackage{footnote}
\usepackage{dsfont}

\usepackage{algorithm}
\usepackage{algorithmic}
\usepackage{nicefrac}

\usepackage{tikz}
\usepackage{overpic}

\usepackage{dsfont}
\usepackage{hyperref}
\usepackage[capitalize]{cleveref}
\usepackage{crossreftools}

\newcommand{\ind}{\mathds{1}}

\newcommand{\brc}[1]{\left\{{#1}\right\}}
\newcommand{\paren}[1]{\left({#1}\right)} 

\DeclareMathOperator*{\argmax}{argmax}
\DeclareMathOperator*{\argmin}{argmin}

\def\gA{{\mathcal{A}}}
\def\gB{{\mathcal{B}}}

\def\gN{{\mathcal{N}}}

\def\gS{{\mathcal{S}}}

\def\gX{{\mathcal{X}}}

\def\RB{{\mathbb R}}
\def\EB{{\mathbb E}}

\def\PB{{\mathbb P}}

\makeatletter
\long\def\@makecaption#1#2{
  \vskip 0.8ex
  \setbox\@tempboxa\hbox{\small {\bf #1:} #2}
  \parindent 1.5em  
  \dimen0=\hsize
  \advance\dimen0 by -3em
  \ifdim \wd\@tempboxa >\dimen0
  \hbox to \hsize{
    \parindent 0em
    \hfil 
    \parbox{\dimen0}{\def\baselinestretch{0.96}\small
      {\bf #1.} #2
    } 
    \hfil}
  \else \hbox to \hsize{\hfil \box\@tempboxa \hfil}
  \fi
}
\makeatother

\newcommand{\Var}{\mathrm{Var}}

\usepackage{enumitem}
\newtheorem{claim}{Claim}[section]
\newtheorem{lemma}[claim]{Lemma}
\newtheorem{assumption}{Assumption}
\newtheorem{theorem}{Theorem}

\newtheorem{corollary}{Corollary}
\theoremstyle{definition}
\newtheorem{definition}{Definition}
\newtheorem{example}{Example}
\theoremstyle{remark}
\newtheorem{remark}{Remark}

\usepackage{xr}

\title{Semi-Infinitely Constrained Markov Decision Processes and Efficient Reinforcement Learning}

\author{Liangyu Zhang\thanks{Academy of Advanced Interdisciplinary Studies, Peking University; email: \texttt{zhangliangyu@pku.edu.cn}.} 
\and
Yang Peng\thanks{School of Mathematical Sciences, Peking University; email: \texttt{pengyang@pku.edu.cn}.}
\and
Wenhao Yang\thanks{Academy of Advanced Interdisciplinary Studies, Peking University; email: \texttt{yangwenhaosms@pku.edu.cn}.}
\and
Zhihua Zhang\thanks{School of Mathematical Sciences, Peking University; email: \texttt{zhzhang@math.pku.edu.cn}.}
}

\begin{document}
\maketitle

\begin{abstract}
  We propose a novel generalization of constrained Markov decision processes (CMDPs) that we call the \emph{semi-infinitely constrained Markov decision process} (SICMDP).
Particularly, 
we consider a continuum of constraints instead of a finite number of constraints as in the case of ordinary CMDPs.
We also devise two reinforcement learning algorithms
for SICMDPs that we call SI-CRL and SI-CPO.
SI-CRL is a model-based reinforcement learning algorithm.
Given an estimate of the transition model, we first transform the reinforcement learning problem into a linear semi-infinitely programming (LSIP) problem and then use the dual exchange method in the LSIP literature to solve it.
SI-CPO is a policy optimization algorithm.
Borrowing the ideas from the cooperative stochastic approximation approach, we make alternative updates to the policy parameters to maximize the reward or minimize the cost.
To the best of our knowledge, we are the first to apply tools from semi-infinitely programming (SIP) to solve constrained reinforcement learning problems.
We present theoretical analysis for SI-CRL and SI-CPO, identifying their iteration complexity and sample complexity.
We also conduct extensive numerical examples to illustrate the SICMDP model and demonstrate that our proposed algorithms are able to solve complex sequential decision-making tasks leveraging modern deep reinforcement learning techniques.
\end{abstract}

\section{Introduction}
Reinforcement learning has achieved great success in areas such as Game-playing \citep{silver2018general,vinyals2019grandmaster}, robotics \cite{kober2013reinforcement}, large language models \citep{ouyang2022training}, etc.
However, due to safety concerns or physical limitations, in some real-world reinforcement learning problems, we must consider additional constraints that may influence the optimal policy and the learning process \citep{garcia2015comprehensive}.
A standard framework to handle such cases is the constrained Markov Decision Process (CMDP) \citep{altman1999constrained}.
Within the CMDP framework, the agent has to maximize
the expected cumulative reward while
obeying a finite number of constraints, which are usually in the form of expected cumulative cost criteria.

However, we are sometimes concerned with the problem with a continuum of constraints.
For example,
the constraints we meet might be time-evolving or subject to uncertain parameters, which
cannot be formulated as an ordinary CMDP
(see Examples \ref{Example_Time_Evolving} and  \ref{Example_Uncertain}).
In this paper we would study a generalized CMDP  
to address the above problem.  Because the constraints are not only infinite-number but also lie
in a continuous set,
the generalization is not trivial. Fortunately, we find that we can borrow the idea behind semi-infinite programming (SIP) \citep{remez1934determination, hettich1993semi} to deal with the semi-infinite constraints.
Accordingly, we propose \emph{semi-infinitely constrained Markov decision processes} (SICMDPs)
as a novel complement to the ordinary CMDP framework.


We also present two reinforcement learning algorithms to solve SICMDPs called SI-CRL and SI-CPO, respectively.
SI-CRL is a model-based reinforcement learning algorithm designed for tabular cases, and SI-CPO is a policy optimization algorithm for non-tabular cases.
The main challenge is that we need to deal with a continuum of constraints, thus reinforcement learning algorithms for ordinary CMDPs do not work anymore.
In SI-CRL, we tackle this difficulty by first transforming the reinforcement learning problem to an equivalent LSIP problem, which can then be solved using methods in the LSIP literature like the dual exchange methods \citep{Hu1990,reemtsen1998numerical}.
In SI-CPO, we resort to the idea of cooperative stochastic approximation developed in \cite{lan2020algorithms, wei2020comirror}.
As far as we know, we are the first to introduce tools from semi-infinitely programming (SIP) into the reinforcement learning community for solving constrained reinforcement learning problems.

Furthermore, we give theoretical analysis for both SI-CRL and SI-CPO.
We decompose the error of SI-CRL into two parts: the statistical error from approximating the true SICMDP with an offline dataset and the optimization error due to the fact that the solution of the LSIP problem obtained by the dual exchange method is inexact.
On the optimization side, we show that the iteration complexity of SI-CRL is $O\left(\left\{\mathrm{diam}(Y)L\sqrt{|\gS|^2|\gA|m}/\left[(1-\gamma)\epsilon\right]\right\}^m\right)$.
On the statistical side, we show that the sample complexity of SI-CRL is $\widetilde O\left(\frac{|S|^2|A|^2}{\epsilon^2(1-\gamma)^3}\right)$ if the offline dataset is generated by a generative model, and $\widetilde O\left(\frac{|S||A|}{\nu_{\min} \epsilon^2(1-\gamma)^3}\right)$ if the dataset is generated by a probability measure $\nu$ as considered in \cite{chen2019information}.
Here $\widetilde O$ means that all logarithm terms are discarded.
For SI-CPO, things become a little more complicated because other than the statistical error and the optimization error, we also need to consider the function approximation error, which comes from imperfect policy parametrizations.
It is shown if the function approximation error can be controlled to $O(\epsilon)$ order, the iteration complexity of SI-CPO is $\widetilde{O}\left(\frac{1}{\epsilon^2(1-\gamma)^6}\right)$ and the sample complexity of SI-CPO is $\widetilde{O}(\frac{1}{\epsilon^4(1-\gamma)^{10}})$.
Here our iteration complexity bound is equivalent to a typical $\widetilde O(1/\sqrt{T})$ global convergence rate.

We perform a set of numerical experiments to illustrate the SICMDP model and validate our proposed algorithms.
Specifically, we examine two numerical examples, namely the discharge of sewage and ship route planning.
Through the discharge of sewage example, we show the advantage of the SICMDP framework over the CMDP baseline obtained by naive discretization in modeling realistic sequential decision-making problems.
Moreover, we demonstrate the effectiveness of the SI-CRL and SI-CPO algorithms in such tabular environments. 
In the ship route planning example, we illustrate the benefits of the SICMDP framework and the ability of the SI-CPO algorithm to address complex continuous control tasks involving continuous state spaces with modern deep reinforcement learning techniques.


\section{Related work}
Constrained Markov decision processes (CMDPs) have been extensively applied in areas like robotics \citep{ono2015chance}, communication and networks \citep{mastronarde2011fast, singh2018throughput} and finance \citep{abe2010optimizing}.
For a detailed treatment of CMDPs readers may refer to \cite{altman1999constrained}.
A number of reinforcement learning algorithms for CMDPs are proposed, which can be divided into model-based methods and model-free methods.
For model-based methods, \citet{wachi2020safe, zheng2020constrained} considered the case where the reward and cost are random but the transition dynamics are known.
\citet{efroni2020explorationexploitation, amani2021safe, ijcai2021-347} considered the case where the transition dynamics are unknown and need to be estimated, which is a more common setting in the literature of reinforcement learning.
And \citet{vaswani2022near} gave a near minimax optimal sample complexity bound of learning CMDPs.
Most model-free methods can indeed be categorized as policy optimization methods.
\citet{tessler2018reward, ding2020natural} utilized a primal-dual approach that transforms the constrained problem into an unconstrained one by considering the Lagrange functions.
which include Lagrangian methods, actor-critic methods \citet{achiam2017constrained, yang2020projection, liu2020ipo} addressed the constrained problem by adding constraints to the sub-problems used to compute the updating direction in each iteration step.
\citet{xu2021crpo} proposed to solve the CMDP problem by performing alternating updates to maximize the reward or minimize the cost.
Our SI-CRL algorithm uses a similar strategy as in \cite{efroni2020explorationexploitation} in the sense that they all use the optimistic method to transform the reinforcement learning problem into a linear (semi-infinitely) programming problem, which resolves the feasibility issue and makes the theoretical analysis easier as well.
However, our work and \cite{efroni2020explorationexploitation} are very different at the technical level: 1) our theoretical guarantees
are in the form of sample complexity bounds, while the results in \citep{efroni2020explorationexploitation}
are in the form of online regret bounds; the proof techniques are quite different. 2) \citet{efroni2020explorationexploitation} considered the episodic MDPs, while we consider the infinite-horizon case.

The origination of semi-infinitely programming (SIP) can date back to \cite{remez1934determination}.
From then on, SIP has been widely used in quantum physics \citep{2021quantum}, signal processing \citep{moulin1997role, nordebo2001semi}, finance \citep{daum2011novel}, environment science, and engineering \citep{hettich1993semi}.
One may refer to \cite{hettich1993semi, goberna2018recent} for a detailed overview of SIP as well as its recent advances.
One important class of SIP problems is called linear semi-infinitely programming (LSIP).
\citet{GOBERNA2002390} provided a thorough survey about the LSIP theory.
Various numerical methods are proposed to solve SIP problems, including discretization methods \citep{still2001discretization, 2004discretization}, exchange methods \citep{Hu1990, 2010exchange}, and local reduction methods \citep{1970moment, coope1985projected}.
In SI-CRL, we choose to use the dual exchange method in \cite{Hu1990} to solve the LSIP problem therein for its conceptual simplicity as well as concrete theoretical guarantees.
Recently, \citet{wei2020comirror} proposed to solve convex SIP problems via the cooperative stochastic approximation method, which is first developed in \cite{lan2020algorithms} to solve stochastic optimization problems with functional or expectation constraints.


\section{The SICMDP Model}
A semi-infinitely constrained MDP (SICMDP) is defined by a tuple $M=\langle \gS,\gA,Y,P,r,c,u,\mu,\gamma\rangle$.
Here $\gS, \gA, P, r, \mu, \gamma$ are defined in a similar manner as in common infinite-horizon discounted MDPs.
Specifically, $\gS$ and $\gA$ are the finite sets of states and actions, respectively. 
$P$ is the transition dynamics and $P(s^\prime|s,a)$ represents the probability of transitioning to state $s^\prime$ when playing action $a$ at state $s$. And
$r\colon \gS\times \gA\to[0,1]$ is the reward function,  
$\mu$ is the fixed initial distribution, and $\gamma$ is the discount factor.
$Y$ is the set of constrains, which we define as a compact set in $\RB^m$, and $\mathrm{diam}(Y)<\infty$ denotes its diameter.
That is, $\mathrm{diam}(Y):=\sup_{y,y^\prime\in Y}{\|y-y^\prime\|_\infty}$.
In addition, $c\colon Y\times \gS\times \gA\to[0,1]$ is used to denote a continuum of cost functions and the value for constraints (bounds that must be satisfied) is determined by function $u\colon Y\to \RB$. 
Note that when $Y$ is finite, we get an ordinary constrained MDP, which is indeed a special case of SICMDP.

For a given policy $\pi$, we define the value function ${V^\pi_r(s)=\EB\left(\sum_{t=0}^\infty \gamma^tr(s_t,a_t)|s_0=s\right)}$, the state-action value function ${Q^\pi_r(s,a)=\EB\left(\sum_{t=0}^\infty \gamma^tr(s_t,a_t)|s_0=s,a_0=a\right)}$, and the advantage function ${A^\pi_r(s,a)=Q^\pi_r(s,a)-V^\pi_r(s)}$. Here
$V_{c_y}^\pi(s)$, $Q_{c_y}^\pi(s,a)$ and $A_{c_y}^\pi(s,a)$ are defined in a similar manner.
Let the occupancy measure on $\gS\times \gA$ introduced by policy $\pi$ be $\nu_\pi\in\Delta(\gS\times \gA)$ and ${\nu_\pi(s,a)=(1-\gamma)\sum_{t=0}^\infty\gamma^t \PB_\pi(s_t=s,a_t=a)}$.

The general SICMDP problem is to find a stationary policy $\pi\colon \gS\to \Delta(\gA)$, where $\Delta(\gA)$ is the set of probability measure supported on $\gA$, to maximize the value function while complying with a continuum of constraints. 
In other words, we  consider the following optimization problem:
\begin{equation}\label{Problem_SICMDP} \tag{M}
\begin{aligned}
\max_{\pi}\ V_r^\pi(\mu)\quad
\text{s.t.}\ V_{c_y}^\pi(\mu) \leq u_y,\ \forall y\in Y. 
\end{aligned}
\end{equation}

Let us see two concrete examples of SICMDPs.  
\begin{example}[Spatial-temporal Constraints]\label{Example_Time_Evolving}
Consider an ordinary CMDP problem with a single constraint:
\begin{equation}\label{Problem_CMDP_Single_Constraint}
\begin{aligned}
\max_{\pi}\ V_r^\pi(\mu)\quad
\text{s.t.}\ V_c^\pi(\mu) \leq u.
\end{aligned}
\end{equation}
In some cases the constraint would be spatial-temporal, i.e., the cost function $c(s,a)$ and the value for constraints $u$ are no longer constant functions and would change with time $\tau \in[0,T]$ or location $x\in \gX\subset\RB^3$.
Then we should use the SICMDP model with $Y=[0,T]$ or $Y=\gX$ rather than the ordinary CMDP framework to model such problems:
\begin{equation}\label{Problem_SICMDP_Time_Evolving}
\begin{aligned}
\max_{\pi}\ V_r^\pi(\mu)\quad
\text{s.t.}\ V_{c_\tau}^\pi(\mu) \leq u_\tau, \; \forall \tau \in [0,T],
\end{aligned}
\end{equation}
or
\begin{equation}\label{Problem_SICMDP_Space}
\begin{aligned}
\max_{\pi}\ V_r^\pi(\mu)\quad
\text{s.t.}\ V_{c_x}^\pi(\mu) \leq u_x, \; \forall x \in \gX.
\end{aligned}
\end{equation}

\textit{Load Balancing}: Suppose an RL agent needs to balance the load between multiple cell sites using some policy $\pi$.
The objective is to minimize the cost $V_r^\pi(\mu)$ and the constraint is that at every place $x$ in the region $\gX$ the cumulative communication capacity $V_{c_x}^\pi(\mu)$ is above some adaptive threshold $u_x$.

\textit{Ship Route Planning}: Suppose we need to navigate a ship using some policy $\pi$.
Our objective is to minimize the voyage.
The constraint is that at every place $x$ in the region $\gX$ the cumulative environmental pollution $V_{c_x}^\pi(\mu)$ is below some adaptive threshold $u_x$.
\end{example}

\begin{example}[Constraints with Uncertainty]\label{Example_Uncertain}
Again we consider a problem like Problem (\ref{Problem_CMDP_Single_Constraint}).
In many application scenarios the cost function $c(s,a)$ is handcrafted and the construction of $c(s,a)$ is not guaranteed to be correct.
Hence it may be helpful to include an additional parameter $\epsilon\in E$ representing our uncertainty in the construction of the cost function $c(s,a)$ as well as the value of constraints $u$.
Even if the constraint is not handcrafted and has clear physical meaning, it may still subject to uncertain parameters $\epsilon\in E$ that cannot be observed in advance.
Therefore, we should use the SICMDP model with $Y=E$ rather than the ordinary CMDP framework to model such problems:
\begin{equation}\label{Problem_SICMDP_Uncertain}
\begin{aligned}
\max_{\pi}\ V_r^\pi(\mu)\quad
\text{s.t.}\ V_{c_\epsilon}^\pi(\mu) \leq u_\epsilon, \forall \epsilon\in E.
\end{aligned}
\end{equation}

\textit{Underwater Drone}: Suppose an underwater drone needs to maximize $V_r^\pi(\mu)$ to accomplish some tasks.
When the unknown environment feature (salinity, temperature, ocean current, etc,) is $\epsilon\in E$, for state-action pair $(s,a)$ the energy consumption is $c_\epsilon(s,a)$, and the constraint is that total energy consumption $V_{c_\epsilon}^\pi(\mu)$ cannot be larger than its battery capacity $u_\epsilon$.
\end{example}

\begin{remark}\label{Remark_Baseline}
An alternative approach to solving problems such as Examples~\ref{Example_Time_Evolving} and ~\ref{Example_Uncertain} is to naively discretize the constraint set $Y$, and then the discretized problem can be fit into the conventional CMDP framework.
We call this strategy naive discretization.
The problem with this naive method is that the prior knowledge, i.e., the constraint function is continuous w.r.t.\ $y$, would be lost, which makes the method extremely inefficient.
In Section~\ref{Section_Experiment} we demonstrate this issue via numerical examples.

\end{remark}

When a SICMDP $M$ is known to us, we may do the planning by solving a linear semi-infinite programming (LSIP) problem.
Problem (\ref{Problem_SICMDP}) can be reformulated as the following LSIP problem:
\begingroup
\small
\begin{equation}\label{Problem_SICMDP_LSIP}
\begin{aligned}
\max_{\nu}\ &\nu^\top r \\
\text{s.t.}\ &\frac{1}{1-\gamma}\nu^\top c_y\leq u_y, \; \forall y\in Y, \\
& \sum_{s^\prime ,a}  \nu(s^\prime,a)(\mathbf{1}_{\{s^\prime=s\}} {-} \gamma P(s|s^\prime,a))=(1{-}\gamma)\mu(s), \; \forall s\in \gS, \\
&\nu \succeq 0.
\end{aligned}
\end{equation}
\endgroup
Here $\nu$ represents the occupancy measure on $\gS\times\gA$ induced by some policy $\pi$.
And ${\pi(a|s)=\frac{\nu_\pi(s,a)}{\sum_{a^\prime\in \gA}\nu_\pi(s,a^\prime)}}$.
Therefore, when $M$ is already known the optimal policy $\pi^*$ can be found by solving Problem (\ref{Problem_SICMDP_LSIP}).
And we always assume such a policy $\pi^*$ exists.
\begin{assumption}\label{Assumption_Feasible}
Problem (\ref{Problem_SICMDP}) is feasible with an optimal solution $\pi^*$, or equivalently, Problem (\ref{Problem_SICMDP_LSIP}) is feasible with an optimal solution $\nu^*$.
\end{assumption}
\section{Algorithms}
In this section, we present two reinforcement learning algorithms called semi-infinitely constrained reinforcement learning (SI-CRL) and semi-infinitely constrained policy optimization (SI-CPO), respectively.
SI-CRL is a model-based reinforcement learning algorithm that can solve tabular SICMDP in a sample-efficient way.
The SI-CPO algorithm is a policy optimization algorithm and it works for large-scale SICMDPs where we can use complex function approximators such as deep neural networks to approximate the policy and the value function.
\subsection{The SI-CRL Algorithm}\label{Section_SICRL}
From a high-level point of view, the SI-CRL algorithm is a semi-infinite version of the algorithms proposed in \cite{ijcai2021-347, efroni2020explorationexploitation}.
In the first stage, SI-CRL takes an offline dataset $\{(s_i, a_i, s_i^\prime)|i=1, 2, \ldots, m\}$ as input and generates an empirical estimate $\widehat P$ of the true transition dynamic $P$.
Then the algorithm constructs a confidence set (the optimistic set) according to $\widehat P$ that would cover the true SICMDP with high probability.
For each policy $\pi$ we would only view its return as the largest possible return in SICMDPs in the confidence set.
This method is also called the optimistic approach.
In the second stage, we reformulate the problem as an LSIP problem and find the optimistic policy $\hat \pi$ using an LSIP solver.
It can be shown that the resulting policy $\hat\pi$ is guaranteed to be nearly optimal, and the theoretical analysis can be found in Section \ref{Section_Theory_SICRL}.

Now we give a more detailed description of SI-CRL.
First, the empirical estimate $\widehat P$ is calculated as:
$\widehat P(s^\prime|s, a):=\frac{n(s, a, s^\prime)}{\max\paren{1,n(s,a)}}$,
where $n(s,a,s^\prime) :=\sum_{i=1}^m \mathbf{1}\{s_i=s, a_i=a, s_i^\prime=s^\prime\}$ and $n(s,a)=\sum_{s^\prime} n(s,a,s^\prime)$.
The reason why we do not directly plug $\widehat P$ into Problem \eqref{Problem_SICMDP_LSIP} and solve the resulting LSIP problem is due to the fact that there is no guarantee that the LSIP problem w.r.t.\ $\widehat P$ is feasible.
To address this issue, we construct an optimistic set $M_\delta$ such that with high probability the true SICMDP $M$ lies in $M_\delta$.
In particular, $M_\delta$ is defined via the empirical Bernstein's bound and the Hoeffding's bound \citep{LATTIMORE2014125}:
\begin{align*}
M_\delta :=& \Big\{\langle \gS,\gA,Y,P^\prime,r,c,u,\mu, \gamma \rangle\colon  |P^\prime(s^\prime|s,a)-\widehat P(s^\prime|s,a)| \leq d_\delta(s,a,s^\prime), \forall s, s^\prime\in \gS, a\in \gA \Big\},
\end{align*}
where 
\begin{align*}
d_\delta(s,a,s^\prime):=&\min\left\{\sqrt{\frac{2\widehat P(s^\prime|s,a)(1 {-} \widehat P(s^\prime|s,a))\log(4/\delta)}{n(s,a,s^\prime)}}+\frac{4\log (4/\delta)}{n(s,a,s^\prime)}, \; \sqrt{\frac{\log (2/\delta)}{2n(s,a,s^\prime)}}\right\}.
\end{align*}

The next step is to solve the optimistic planning problem:
\begin{equation}\label{Problem_Optimistic}
\begin{aligned}
\max_{M^\prime\in M_\delta,\pi}\ V_r^{\pi,M^\prime}(\mu),\quad
\text{s.t.}\ V_{c_y}^{\pi,M^\prime}(\mu) \leq u_y,\ \forall y\in Y,
\end{aligned}
\end{equation}
where the superscript $M^\prime$ denotes that the expectation is taken w.r.t.\ SICMDP $M^\prime$.
\begin{theorem}\label{Theorem_Feasible}
Suppose $n\geq 3$. With probability at least $1-2|\gS|^2|\gA|\delta$, we have that $M\in M_\delta$, and Problem (\ref{Problem_Optimistic}) is feasible.
\end{theorem}

\proof {Proof of Theorem~\ref{Theorem_Feasible}}
See Appendix~\ref{Appendix_Proofs_4}.
\endproof

Note that the optimization variables include both $M^\prime$ and $\pi$, and LSIP reformulations like Problem (\ref{Problem_SICMDP_LSIP}) would no longer be possible. 
Instead, we shall introduce the state-action-state occupancy measure $z(s,a,s^\prime)$.
In particular, assuming $z_{P,\pi}(s,a,s^\prime):=P(s^\prime|s,a)q_\pi(s,a)$, we have $P(s^\prime|s,a)=\frac{z_{P,\pi}(s,a,s^\prime)}{\sum_{x\in \gS}z_{P,\pi}(s,a,x)}$, and $\pi(a|s)=\frac{\sum_{s^\prime\in \gS}z_{P,\pi}(s,a,s^\prime)}{\sum_{s^\prime\in \gS,a^\prime\in \gA}z_{P,\pi}(s,a^\prime,s^\prime)}$. 
Problem (\ref{Problem_Optimistic}) can be reformulated as the following extended LSIP problem:

\begingroup
\small
\begin{equation}\label{Problem_Optimistic_ELSIP}
\begin{aligned}
    \max_{z}\ &\sum_{s, a,s^\prime}z(s,a,s^\prime)r(s,a) \\
    \text{s.t.}\ &\frac{1}{1-\gamma}\sum_{s, a,s^\prime}z(s,a,s^\prime)c_y(s,a)\leq u_y,\ \forall y\in Y, \\
    &z(s,a,s^\prime)\leq (\widehat P(s^\prime|s,a)+d_\delta(s,a,s^\prime))\sum_{x\in \gS} z(s,a,x), \forall s,s^\prime,\ a\in \gA, \\
    &z(s,a,s^\prime)\geq (\widehat P(s^\prime|s,a)-d_\delta(s,a,s^\prime))\sum_{x\in \gS} z(s,a,x), \forall s,s^\prime\in \gS,\ a\in \gA, \\
    &\sum_{x\in \gS,b\in \gA}z(s,b,x)=(1-\gamma)\mu(s)+\gamma\sum_{x\in \gS,b\in \gA}z(x,b,s), \forall s\in \gS, \\
    &z\succeq 0.
\end{aligned}
\end{equation}
\endgroup

However, compared to LP problems, LSIP problems are typically harder to solve and there are no all-purpose LSIP solvers.
Here, we choose the simple yet effective dual exchange methods \citep{Hu1990,reemtsen1998numerical} to solve Problem~\ref{Problem_Optimistic_ELSIP}.
The SI-CRL algorithm can be summarized in Algorithm~\ref{Algorithm_SICRL}.
A key ingredient of Algorithm~\ref{Algorithm_SICRL} is solving the inner-loop optimization problem 
$$
\max_{y\in Y} \sum_{s, a,s^\prime}z(s,a,s^\prime)c_y(s,a)-u_y.
$$
We can obtain different versions of SI-CRL algorithm by choosing different optimization subroutines to solve the inner-loop problem above. 
If $c_y$ and $u_y$ satisfy conditions like concavity and smoothness, then the inner problem can be solved using methods like projected subgradient ascent \citep{bubeck2015convex}.
If the inner problem is ill-posed, we may still solve it using methods like random search \citep{solis1981minimization, andradottir2015review}.
\begin{algorithm}[htb]
   \caption{SI-CRL}
   \label{Algorithm_SICRL}
\begin{algorithmic}
   \STATE {\bfseries Input:} state space $\gS$, action space $\gA$, dataset $\{(s_i,a_i,s_i^\prime)|i=1,2,...,m\}$, reward function $r$, a continuum of cost function $c$, index set $Y$, value for constraints $u$, discount factor $\gamma$, tolerance $\eta$, maximum iteration number $T$.
   \FOR{each $(s,a,s^\prime)$ tuple}
   \STATE Set $\widehat P(s^\prime|s,a):=\frac{\sum_{i=1}^m \ind\{s_i=s,a_i=a,s_i^\prime=s^\prime\}}{\max\paren{1,\sum_{i=1}^m \ind\{s_i=s,a_i=a\}}}$
   \ENDFOR
   \STATE Initialize $Y_0=\{y_0\}$
   \FOR{$t=1$ {\bfseries to} $T$}
   \STATE Use an LP solver to solve a finite version of Problem (\ref{Problem_Optimistic_ELSIP}) by only considering constraints in $Y_0$ and store the solution as $z^{(t)}$.
   \STATE Find $y^{(t)}\approx\argmax_{y\in Y} \sum_{s, a,s^\prime}z^{(t)}(s,a,s^\prime)c_y(s,a)-u_y$.
   \IF {$\sum_{s, a,s^\prime}z(s,a,s^\prime)c_{y^{(t)}}(s,a)-u_{y^{(t)}} \leq\eta$}
   \STATE  Set $z^{(T)}=z^{(t)}$.
   \STATE  {\bfseries BREAK}
   \ENDIF
   \STATE Add $y^{(t)}$ to $Y_0$.
   \ENDFOR
   \FOR{each $(s,a)$ pair}
   \STATE Set $\hat\pi(a|s)=\frac{\sum_{s^\prime}z^{(T)}(s,a,s^\prime)}{\sum_{s^\prime,a^\prime}z^{(T)}(s,a^\prime,s^\prime)}$.
   \ENDFOR
   \STATE {\bfseries RETURN} $\hat{\pi}$.
\end{algorithmic}
\end{algorithm}

\subsection{The SI-CPO Algorithm}\label{Section_SICPO}
In SI-CPO, we borrow ideas from the cooperative stochastic approximation \citep{lan2020algorithms, wei2020comirror} to deal with the infinitely many constraints.
At a certain iteration, the SI-CPO algorithm first determines whether the constraint violation is below some tolerance or not.
It then performs a single step of policy optimization along the direction of maximizing the value of reward if the constraint violation is below some tolerance;
or performs a single step of policy optimization along the direction of minimizing the value of some cost corresponding to a violated constraint.

We now describe the SI-CPO algorithm in more detail.
We follow the convention to define the parameterized policy class as $\{\pi_\theta,\theta\in\Theta\subset\RB^d\}$ and use $\pi^{(t)}$ in short of $\pi_{\theta^{(t)}}$, $V_\diamond^{(t)}$ in short of $V_\diamond^{\pi^{(t)}}$ for ease of notation.
Here $\diamond$ represents either the reward $r$ or some cost $c_y$.
Suppose at the $t$-th iteration our policy parameter is $\theta^{(t)}$, then we first construct an estimate $\widehat V^{(t)}_{c_y}(\mu)$ using some policy evaluation subroutine.
Next, we are to solve a subproblem using some optimization subroutine
$$
y^{(t)}=\argmax_y\ \widehat V_{c_y}^{\pi^{(t)}}(\mu)-u_y.
$$
If ${\widehat V_{c^{(t)}}^{\pi^{(t)}}(\mu)-u_{y^{(t)}}\leq \eta}$, where $c^{(t)}:=c_{y^{(t)}}$ and $\eta> 0$ is a threshold of tolerance, we say the constraint violation is small and add the time index $t$ to the ``good set" $\gB$.
Then we perform a step of update with a policy optimization subroutine to maximize the value of reward $V_r^{(t)}(\mu)$ to get $\theta^{(t+1)}$.
Else, we first add the time index $t$ to the ``bad set" $\gN$.
Next, we find the violated constraint ${V_{c^{(t)}}^{\pi^{(t)}}(\mu)-u_{y^{(t)}}>\eta}$, and perform a step of update with a policy optimization subroutine to minimize the value of cost $V_{c^{(t)}}^{\pi^{(t)}}(\mu)$ to get $\theta^{(t+1)}$.
After $T$ iterations, we draw $\hat\theta$ uniformly from the set ${\{\theta^{(t)},t\in\gB\}}$, as return the policy ${\hat\pi=\pi_{\hat\theta}}$.
The procedure of SI-CPO  is summarized in Algorithm~\ref{Algorithm_SICPO}.

We can get different instances of the SI-CPO algorithms by making different choices of the subroutines aforementioned.
Specifically, the policy optimization subroutine can be any policy optimization algorithm like policy gradient~(PG) \citep{sutton1999policy}, natural policy gradient~(NPG) \citep{kakade2001natural}, trust-region policy gradient~(TRPO) \cite{schulman2015trust}, or proximal policy optimization~(PPO) \citep{schulman2017proximal}.
The policy evaluation subroutine can be chosen as Monte-Carlo policy evaluation algorithms \citep{curtiss1954theoretical} or various TD-learning algorithms \citep{sutton1988learning, dann2014policy}.
We may also integrate the policy optimization subroutine and the policy evaluation subroutine into actor-critic-type algorithms \citep{konda1999actor}.
The optimization subroutine can be any optimization algorithm suitable for the problem instance, like the case in Algorithm~\ref{Algorithm_SICRL}.

\begin{algorithm}[htb]
   \caption{SI-CPO}
   \label{Algorithm_SICPO}
\begin{algorithmic}
   \STATE {\bfseries Input:} state space $\gS$, action space $\gA$, reward function $r$, a continuum of cost function $c$, index set $Y$, value for constraints $u$, discount factor $\gamma$, learning rate $\alpha$, tolerance $\eta$, maximum iteration number $T$.
   \STATE Initialize $\gB=\emptyset$, $\gN=\emptyset$, $\theta^{(0)}=\theta_0\in\Theta$.
   \FOR{$t=0,...,T-1$}
   \STATE Obtain $\widehat V_{c_y}^{\pi^{(t)}}(\mu)$ via a policy evaluation subroutine.
   \STATE Use an optimization subroutine to solve ${\max_y\ \widehat V_{c_y}^{\pi^{(t)}}(\mu)-u_y}$, and set ${y^{(t)}\approx\argmax_y \widehat V_{c_y}^{\pi^{(t)}}(\mu)-u_y}$, $c^{(t)}=c_{y^{(t)}}$.
   \IF {$\widehat V_{c^{(t)}}^{\pi^{(t)}}(\mu)-u_{y^{(t)}}\leq \eta$}
   \STATE  Perform a step of policy update to maximize $V_r^{\pi^{(t)}}(\mu)$ to get $\pi^{(t+1)}$. Specifically,
   ${\theta^{(t+1)}=\theta^{(t)}+\alpha\hat w^{(t)}}.$
   \STATE Add $t$ to $\gB$
   \ELSE 
   \STATE  Perform a step of policy update to minimize $V_{c^{(t)}}^{\pi^{(t)}}(\mu)$ to get $\pi^{(t+1)}$. Specifically, ${\theta^{(t+1)}=\theta^{(t)}-\alpha\hat w^{(t)}}.$
   \STATE Add $t$ to $\gN$
   \ENDIF
   \ENDFOR
   \STATE {\bfseries RETURN} $\hat\pi=\pi_{\hat\theta}$, where $\hat\theta\sim\mathrm{Unif}\left(\{\theta^{(t)}, t\in\gB\}\right)$.
\end{algorithmic}
\end{algorithm}

\section{Theoretical Analysis}
\subsection{Theoretical Analysis of SI-CRL}\label{Section_Theory_SICRL}
We give PAC-type bounds for SI-CRL under different settings.
The error of SI-CRL is decomposed into two parts: the optimization error from the fact that the solution of (\ref{Problem_Optimistic}) obtained by the dual exchange method is inexact and the statistical error from approximating Problem (\ref{Problem_SICMDP}) with Problem (\ref{Problem_Optimistic}).
On the optimization side, we show that if the inner maximization problem w.r.t.\ $y$ is solved via random search or projected subgradient ascent, the dual exchange method would produce an $\epsilon$-optimal solutions (see Definition \ref{Definition_Optimal_Solution})
when the number of iterations $T=O\left(\left[\frac{\mathrm{diam}(Y)|\gS|^2|\gA|}{(1-\gamma)\epsilon}\right]^m \right)$.

On the statistical side, our goal is to determine how many samples are required to make SI-CRL an $(\epsilon, \delta)$-optimal~(see Definition \ref{Definition_PAC}) when Problem (\ref{Problem_Optimistic}) can be solved exactly, i.e., we want to find the sample complexity of SI-CRL  (see Definition \ref{Definition_PAC}).
We show that the sample complexity of SI-CRL is $\widetilde O\left(\frac{|\gS|^2|\gA|^2}{\epsilon^2(1-\gamma)^3}\right)$ if the dataset we use is generated by a generative model, and $\widetilde O\left(\frac{|\gS||\gA|}{\nu_{\min} \epsilon^2(1-\gamma)^3}\right)$ if the dataset we use is generated by a probability measure $\nu$ defined on the space $\gS\times \gA$ and $P(\cdot|s,a)$ as considered in \cite{chen2019information}.
Here $\widetilde O$ means that all logarithm terms are discarded, and $\nu_{\min}:=\min_{\nu(s,a)>0}\nu(s,a)$.
We will present our theoretical analysis in more detail in the following part of this section.

\subsubsection{Preliminaries}

Let $\pi^*$ denote the optimal policy.
An $(\epsilon,\delta)$-optimal policy is defined as follows. 
\begin{definition}\label{Definition_PAC}
An RL algorithm is called $(\epsilon,\delta)$-optimal for $\epsilon,\delta>0$ if with probability at least $1-\delta$ it can return a policy $\pi$ such that
$$
\begin{aligned}
V_r^{\pi^*}(\mu)-V_r^{\pi}(\mu) &\leq \epsilon;\quad
V_{c_y}^{\pi}(\mu) - u_y  \leq \epsilon, \forall y\in Y.
\end{aligned}
$$
\end{definition}
An $\epsilon$-optimal solution of Problem (\ref{Problem_Optimistic}) is defined as \begin{definition}\label{Definition_Optimal_Solution}
A stationary policy $\hat\pi$ is called an $\epsilon$-optimal solution of Problem (\ref{Problem_Optimistic}) for $\epsilon>0$ if 
$$
\begin{aligned}
|V_r^{\hat\pi}(\mu)-V_r^{\tilde\pi}(\mu)| &\leq \epsilon \quad \mbox{and} \quad
|V_{c_y}^{\hat\pi}(\mu) - u_y|  \leq \epsilon,  \forall y\in Y \\
\end{aligned}
$$
hold simultaneously.
\end{definition}

Unless otherwise specified, we assume that $\forall (s,a)\in \gS\times \gA$, $c_y(s,a)$ is $L_y$-Lipschitz in $y$ w.r.t.\ $\|\cdot\|_\infty$.
We also assume that $u_y$ is $L_y$-Lipschitz in $y$ w.r.t.\ $\|\cdot\|_\infty$.
The assumptions can be formally stated as:
\begin{assumption}\label{Assumption_Lipschitz}
$c_y(s,a)$ and $u_y$ are Lipschitz in $y$ w.r.t.\ $\|\cdot\|_\infty$, i.e., $\exists L_y>0$ s.t. $\forall y,y^\prime\in Y, (s,a)\in \gS\times \gA, |c_y(s,a)-c_{y^\prime}(s,a)|\leq L_y\|y-y^\prime\|_\infty, 
|u_y-u_{y^\prime}|\leq L_y\|y-y^\prime\|_\infty$.
\end{assumption}
The Lipschitz assumption is usually necessary when dealing with a semi-infinitely constrained problem \citep{still2001discretization,Hu1990}.
And this assumption is indeed quite mild because $Y$ is a compact set.

We say an offline dataset $\{(s_i,a_i,s_i^\prime)|i=1, 2, \ldots, n\}$ to be generated by a generative model if we sample according to $P(\cdot|s,a)$ for each $(s,a)$-pair $n_0=n/|\gS||\gA|$ times and record the results in the dataset.
We say an offline dataset to be generated by probability measure $\nu$ and $P(\cdot|s,a)$ if $(s_i,a_i)\stackrel{i.i.d.}{\sim} \nu$ and $s_i^\prime\sim P(\cdot|s_i,a_i)$.

We solve the inner-loop problem in Algorithm~\ref{Algorithm_SICRL} with random search or projected gradient ascent.
The idea of random search is simple.
For an objective $f(y)$ defined on domain $Y$, we form a random grid of $Y$ consisting of $M$ grid points and select the grid point with the largest objective value.
The precise definition can be found in Algorithm~\ref{Algorithm_random_search} in Appendix~\ref{Appendix_Algorithm}.
The projected subgradient ascent is defined in a standard way \citep{bubeck2015convex}.
The precise definition can be found in Algorithm~\ref{Algorithm_projected_GD} in Appendix~\ref{Appendix_Algorithm}.

\subsubsection{Iteration Complexity of SI-CRL}

We give the iteration complexity of SI-CRL, i.e., how many iterations are required to output an $\epsilon$-optimal solution of Problem (\ref{Problem_Optimistic}) when near-optimal solutions of the inner-loop optimization problems can be obtained.
Our result is similar to Theorem 4 in \cite{Hu1990}.
Specifically, we consider two different cases: 1) we make no assumption of the constraint and use random search to solve the inner-loop problem; 2) we assume the constraint is concave and use projected subgradient ascent to solve the inner-loop problem.

Before we give the iteration complexity of the case of random search, we make the following assumption to ensure technical rigor.
\begin{assumption}\label{Assumption_regular_maxima}
     For any $(s,a)\in\gS\times\gA$ and weight $v\in \RB^{\gS\times\gA}$, let $y_0\in\arg\max_{y\in Y} (v^\top  c_y-u_y)$. Then $\exists y_0$ such that
    $$
    \{y:\|y-y_0\|_\infty\leq \epsilon_0\}\subset Y.
    $$
\end{assumption}
Assumption~\ref{Assumption_regular_maxima} guarantees any possible solution of the inner-loop problem lies in the interior of $Y$.
\begin{theorem}\label{Theorem_Iteration_Complexity_Random_Search}
Suppose we use random search to solve the inner-loop problem of the SI-CRL algorithm, then if we set the size of random grid $M=O\left(\frac{\log(\delta/T)}{\log \left(1-((1-\gamma)\epsilon/|\gS|^2|\gA|\mathrm{diam}(Y))^m\right)}\right)$, $T=O\left(\left[\frac{\mathrm{diam}(Y)|\gS|^2|\gA|}{(1-\gamma)\epsilon}\right]^m \right)$, SI-CRL would output a $\epsilon$-optimal solution of Problem~\ref{Problem_Optimistic_ELSIP} with probability at least $1-\delta$.
Here we require $\epsilon\leq \frac{2|\gS|^2|\gA|L_y\epsilon_0}{1-\gamma}$.
\end{theorem}

\proof{Proof of Theorem~\ref{Theorem_Iteration_Complexity_Random_Search}.}
See Appendix~\ref{Appendix_Proofs_SICRL}.
\endproof

To derive theoretical guarantees for the case of projected subgradient ascent, we need the following assumption of concavity.

\begin{assumption}\label{Assumption_concave_constraint}
     For any $(s,a)\in\gS\times\gA$, $c_y(s,a)$ is concave in $y$. In addition, $u_y$ is convex in $y$.
\end{assumption}

\begin{theorem}\label{Theorem_Iteration_Complexity_Projected_GD}
Suppose we use projected gradient ascent to solve the inner-loop problem of the SI-CRL algorithm, then if we set the iteration number of the projected subgradient ascent $T_{PGA}=O\left(\frac{|\gS|^4|\gA|^2\mathrm{diam}(Y)^2}{(1-\gamma)^2\epsilon^2}\right)$, $T=O\left(\left[\frac{\mathrm{diam}(Y)|\gS|^2|\gA|}{(1-\gamma)\epsilon}\right]^m \right)$, SI-CRL would output a $\epsilon$-optimal solution of Problem~\ref{Problem_Optimistic_ELSIP}.
\end{theorem}

\proof{Proof of Theorem~\ref{Theorem_Iteration_Complexity_Projected_GD}.}
See Appendix~\ref{Appendix_Proofs_SICRL}.
\endproof

The most crucial part of our proof is a $\epsilon$-packing argument.
Suppose we can get a $\epsilon/2$-optimal solution to the inner-loop problem by either random search of projected subgradient ascent and set the tolerance $\eta=\epsilon/2$.
By the assumption of Lipschitzness and the construction of the SI-CRL algorithm, for any $t\leq T$, either the SI-CRL algorithm has already terminated and we obtain a $\epsilon$-optimal solution to Problem~\ref{Problem_Optimistic_ELSIP}, or $\{B^{(t^\prime)},t=1,...,t\}$ forms a packing of $Y$.
Here $B^{(t^\prime)}:=\{y:\|y-y^{(t^\prime)}\|_\infty\leq \epsilon/2\beta\}$, and $\beta$ is some Lipschitz coefficient.
Then we may draw the conclusion by noting that the maximum iteration number of SI-CRL is no larger than the $\epsilon/2\beta$-packing number of $Y$.
We find that \cite{Hu1990} also used similar techniques to derive their convergence rate, although they assume the inner-loop problem can always be solved exactly.

\begin{remark}
    The iteration complexity of the SI-CRL algorithm grows with $m$ in an exponential manner.
    Thus from a theoretical viewpoint, the SI-CRL algorithm is no better than the naive discretization method mentioned in Remark~\ref{Remark_Baseline}.
    However, we find SI-CRL is far more efficient than the naive method in empirical evaluations.
    Perhaps it is because our bound of iteration complexity is obtained by the packing argument and not tight enough.
    Hopefully, the bound can be tightened by a refined analysis of the dynamics of $\{(y^{(t)}, z^{(t)}),t=1,...,T\}$.
\end{remark}

\subsubsection{Sample Complexity of SI-CRL}
We consider the case where the offline dataset we use is generated by a generative model.
First, we consider a restricted setting as in \cite{LATTIMORE2014125} where for each $(s,a)$-pair in the true SICMDP there are at most two possible next-states and provide the sample complexity bound.
Then we will drop Assumption \ref{Assumption_Two_Nonzero} using the same strategy as in \cite{LATTIMORE2014125} and derive the sample complexity bound of the general case.
\begin{assumption}\label{Assumption_Two_Nonzero}
The true unknown SICMDP $M$ satisfies $P(s^\prime|s,a)=0$ for all but two $s^\prime\in \gS$ denoted as $sa^+$ and  $sa^-\in \gS$.
\end{assumption}

Although Assumption \ref{Assumption_Two_Nonzero} seems quite restrictive, we argue that it is necessary to establish sharp sample complexity bound, as shown in \cite{LATTIMORE2014125}.
Specifically, without this assumption the ``quasi-Bernstein bound'' (Lemma \ref{Lemma_Quasi_Bernstein}) will not hold, thus we may not be able to get the $\widetilde O((1-\gamma)^{-3})$ bound.

\begin{lemma}\label{Lemma_Bound_on_V}
Suppose Assumption \ref{Assumption_Two_Nonzero} holds, and the dataset we use is generated by a generative model with $n/|\gS||\gA|=n_0>\max\left\{\frac{36\log4/\delta}{(1-\gamma)^2}, \frac{4\log4/\delta}{(1-\gamma)^3}\right\}$. Then with probability $1-2|\gS|^2|\gA|\delta$, we have that
$$\begin{aligned}
V_r^{\pi^*}(\mu)-V_r^{\tilde\pi}(\mu)\leq 24\sqrt{\frac{\log 4/\delta}{{n_0}(1-\gamma)^3}};\quad
V_{c_y}^{\tilde \pi}(\mu) - u_y \leq 12\sqrt{\frac{\log 4/\delta}{{n_0}(1-\gamma)^3}}, \; \forall y\in Y.
\end{aligned}
$$
Here $\tilde\pi$ is the exact solution of Problem~\ref{Problem_Optimistic}.
\end{lemma}
\proof{Proof of Lemma~\ref{Lemma_Bound_on_V}.}
See Appendix~\ref{Appendix_Proofs_SICRL}.

\begin{theorem}\label{Theorem_Sample_Complexity}
Suppose Assumption \ref{Assumption_Two_Nonzero} holds, the dataset we use is generated by a generative model and Problem \ref{Problem_Optimistic} can be solved exactly. Then when $n=O\left(\frac{|\gS||\gA|\log \paren{8|\gS|^2|\gA|/\delta}}{\epsilon^2(1-\gamma)^3}\right)$, SI-CRL is $(\epsilon,\delta)$-optimal.
\end{theorem}
\proof{Proof of Theorem~\ref{Theorem_Sample_Complexity}.}
Theorem~\ref{Theorem_Sample_Complexity} is a direct consequence of Lemma~\ref{Lemma_Bound_on_V}.

\begin{theorem}\label{Theorem_Sample_Complexity_General}
Suppose the dataset we use is generated by a generative model and Problem \ref{Problem_Optimistic} can be solved exactly. Then when $n=O\left(\frac{|\gS|^2|\gA|^2\paren{\log|\gS|}^3\log \paren{8|\gS|^4|\gA|^3/\delta}}{\epsilon^2(1-\gamma)^3}\right)$, a modification of SI-CRL is $(\epsilon,\delta)$-optimal.
\end{theorem}
\proof{Proof of Theorem~\ref{Theorem_Sample_Complexity_General}.}
See Appendix~\ref{Appendix_Proofs_SICRL}.

Our proof strategy is similar to \cite{LATTIMORE2014125}. 
However, to get a $\widetilde O((1-\gamma)^{-3})$ bound \cite{LATTIMORE2014125} used a tedious recursion argument.
We greatly simplify the proof and achieve improvements in log terms (by a factor of $(\log(\frac{|\gS|}{\epsilon(1-\gamma)}))^2$) using sharper bounds on local variances of MDPs developed in \cite{pmlr-v125-agarwal20b}.


\begin{remark}\label{Remark_Sample_Complexity_General_Dependence_on_Constraints}
It can be noted that our sample complexity bound does not rely on the constraint set $Y$.
This is because we consider the setting where $r$ and $c_y$ are known deterministic functions and the only source of randomness comes from estimating the unknown transition dynamic using an offline dataset.
In other words, the constraints do not make the problem more difficult in the statistical sense.
\end{remark}

\begin{remark}\label{Remark_Modification}
Here ``a modification of SI-CRL" stands for the following procedure: first we transform the original SICMDP to a new SICMDP satisfying Assumption~\ref{Assumption_Two_Nonzero}, then we run SI-CRL to solve the new SICMDP.
One may refer to the proof in Appendix~\ref{Appendix_Proofs_SICRL} for more details.
\end{remark}

Now we generalize our results to the case where the offline dataset is generated by a probability measure.
\begin{theorem}\label{Theorem_Sample_Complexity_General_Measure}
Suppose the dataset we use is generated by a probability measure $\nu$ and Problem \ref{Problem_Optimistic} can be solved exactly. Then when $m=O\left(\frac{|\gS||\gA|\paren{\log|\gS|}^3\log \paren{8|\gS|^4|\gA|^3/\delta}}{\nu_{\min} \epsilon^2(1-\gamma)^3}\right)$, a modification of SI-CRL is $(\epsilon,\delta)$-optimal.
\end{theorem}
\proof{Proof of Theorem~\ref{Theorem_Sample_Complexity_General_Measure}.}
See Appendix~\ref{Appendix_Proofs_SICRL}.
\subsection{Theoretical Analysis of SI-CPO}\label{Section_Theory_SICPO}
In this section, we present theoretical guarantees of SI-CPO.
We consider a version of the SI-CPO algorithm, where we use sample-based NPG \citep{agarwal2021theory} as the policy optimization subroutine, a finite-horizon Monte-Carlo estimator as the policy evaluation subroutine, and either random search or projected subgradient ascent as the optimization subroutine.
It is shown that when the function approximation error $\epsilon_{bias}$ is of the same order with $\epsilon$, our proposed algorithm takes $\widetilde{O}\left(\frac{1}{\epsilon^2(1-\gamma)^6}\right)$ iterations and make $\widetilde{O}\left(\frac{1}{\epsilon^4(1-\gamma)^{10}}\right)$ interactions with the environment to achieve an $\epsilon$-level of averaged suboptimality with high probability.
This corresponds to a $\widetilde{O}(1/\sqrt{T})$ globally convergence rate, which is typical for NPG-based policy optimization algorithms.
We will give a detailed description of the considered version of the SI-CPO algorithm as well as our technical assumptions in Section~\ref{Subsection_SICPO_Prem} and present the theoretical results in Sections~\ref{Subsection_SICPO_Iteration_Complexity} and~\ref{Subsection_SICPO_Sample_Complexity}.

\subsubsection{Preliminaries}\label{Subsection_SICPO_Prem}
Recall the policy $\pi$ is parameterized by $\theta\in\Theta\subset\RB^d$ (denoted by $~\pi_\theta$).
We make the following assumptions about the parameterized policy class.
\begin{assumption}[Differentiable policy class]\label{Assumption_differentiable}
$\Pi$ can be parametrized as $\Pi_\theta=\{\pi_\theta|\theta\in\RB^d\}$, such that for all $s\in\gS$,~$a\in\gA$, $\log_\theta\pi(s|a)$ is a differentiable function of $\theta$.
\end{assumption}

\begin{assumption}[Lipschitz policy class]\label{Assumption_Lipschitz_policy}
For all $s\in\gS$,~$a\in\gA$, $\log\pi_\theta(s|a)$ is a $L_\pi$-Lipschitz function of $\theta$, i.e.,
$$
\|\nabla_\theta\log\pi_\theta(s|a)\|_{2}\leq L_\pi,\forall s\in\gS,a\in\gA,\theta\in \RB^d.
$$
\end{assumption}

\begin{assumption}[Smooth policy class]\label{Assumption_smooth}
For all $s\in\gS$,~$a\in\gA$, $\log\pi_\theta(s|a)$ is a $\beta$-smooth function of $\theta$, i.e.,
$$
\|\nabla_\theta\log\pi_\theta(s|a)-\nabla_\theta\log\pi_{\theta^\prime}(s|a)\|_{2}\leq \beta\|\theta-\theta^\prime\|_2,\forall s\in\gS,a\in\gA,\theta,\theta^\prime\in \RB^d.
$$
\end{assumption}

\begin{assumption}[Positive semidefinite Fisher information]\label{Assumption_PSD_Fisher}
    For all $\theta\in\RB^d$,
    $$F(\theta):=\EB_{(s,a)\sim\nu_\theta}[\nabla_\theta\log\pi_\theta(a|s)\nabla_\theta\log\pi_\theta(a|s)^\top]\succeq\mu_F I_d.
    $$
\end{assumption}
The assumptions above are standard in the literature of policy optimizations \citep{agarwal2021theory}.
We also assume the parametrization realizes good function approximation in terms of transferred compatible function approximation errors, which is first introduced by \cite{agarwal2021theory}.
The error term can be close to zero if the policy class is rich \citep{wang2019neural} or the underlying MDP has low-rank structures \citep{jiang2017contextual}.
\begin{assumption}[Bounded function approximation error]\label{Assumption_func_approx_err}
The transferred compatible function approximation errors satisfies that $\forall t\in\{1,...,T\}$
$$
\begin{aligned}
\min_w E^{\nu^{(t)}}(r,\theta^{(t)},w)&\leq \epsilon_{\text{bias}}\\
\min_w E^{\nu^{(t)}}(c_y,\theta^{(t)},w)&\leq \epsilon_{\text{bias}}\ \forall y\in Y,\\
\end{aligned}
$$
where $\nu^{(t)}$ denotes the state-action occupancy measure induced by policy $\pi^{(t)}$. 
The transferred compatible function approximation errors are defined as:
$$
E^{\nu}(\diamond,\theta,w):=\EB_{(s,a)\sim\nu}(A^{\pi_\theta}_\diamond(s,a)-w^\top\nabla_\theta\log\pi_\theta(a|s))^2.
$$
\end{assumption}

Besides, we also assume the weights to minimize the transferred compatible function approximation errors are bounded.

\begin{assumption}[Bounded Weight]\label{Assumption_est_err}
For any $t\in\{1,...,T\}$, $\forall y\in Y$,
$$
\left\|\underset{w}{\mathrm{argmin}} E^{\nu^{(t)}}(r,\theta^{(t)},w)\right\|_2^2\leq W^2, \left\|\underset{w}{\mathrm{argmin}} E^{\nu^{(t)}}(c_y,\theta^{(t)},w)\right\|_2^2\leq W^2.
$$
\end{assumption}

In the theoretical analysis of SI-CPO, we consider an instance of SI-CPO where we use a sample-based version of NPG \citep{agarwal2021theory} as the policy optimization subroutine, a fixed-horizon Monte-Carlo estimator as the policy evaluation subroutine, and either random search or projected subgradient ascent as the optimization subroutine.
In the NPG algorithm, we use the following natural policy gradient $w^{(t)}$ to update the policy parameters:
$$
w^{(t)}:=F(\theta^{(t)})^\dagger\EB_{(s,a)\sim\nu^{(t)}}(A^{\pi^{(t)}}_\diamond(s,a)\nabla_\theta\pi_\theta(a|s)).
$$
Here $\diamond$ can be either the reward $r$ or some cost function $c_y$.
However, for most RL problems it is computationally prohibitive to evaluate $F(\theta)^\dagger$, and $\EB_{(s,a)\sim\nu^{(t)}}(A^{\pi^{(t)}}_\diamond(s,a)\nabla_\theta\pi_\theta(a|s))$ are usually unknown to the algorithm.
Therefore, we instead use a sample-based estimate of $w^{(t)}$, which can be obtained by solving the following optimization problem by running $K_{sgd}$ steps of stochastic gradient descent:
$$
\hat w^{(t)}\approx\frac{1}{1-\gamma}\arg\min_w E^{\nu^{(t)}}(b,\theta^{(t)},w),
$$
recall that $E^{\nu^{(t)}}(\diamond,\theta^{(t)},w)$ is the transferred function approximation error defined in Assumption~\ref{Assumption_func_approx_err}.
The precise definition of sample-based NPG can be found in Algorithm~\ref{Algorithm_sample_based_NPG} in Appendix~\ref{Appendix_Algorithm}.

As for policy evaluation, we choose to use a Monte-Carlo estimator with a fixed horizon $H$.
The idea is very simple, in each episode we run the target policy $\pi$ for $H$ steps, and record the return
$$
G_i=\sum_{k=0}^{H-1}\gamma^k c_y(s_k,a_k).
$$
The procedure is repeated for $K_{eval}$ times and we take the average as an estimate of $V_{c_y}^{\pi^{(t)}}(\mu)$.
Compared with the more commonly used unbiased Monte-Carlo estimate
$$\widetilde{G}_i=\sum_{k=0}^{H^\prime-1}c_y(s_k,a_k)
$$
where $H^\prime$ is no longer fixed and drawn from an exponential distribution $\mathrm{Exp(\lambda)}$, $G_i$ does introduce bias, but it also has the advantage of being sub-gaussian.
Moreover, the bias term is always bounded by $\frac{\gamma^H}{1-\gamma}$, which decays exponentially as we choose larger $H$s.

\subsubsection{Iteration complexity of SI-CPO}\label{Subsection_SICPO_Iteration_Complexity}

The following two theorems give the iteration complexity of the SI-CPO algorithm when we use either random search or projected subgradient ascent to solve the inner-loop problem.

\begin{theorem}\label{Theorem_random_search_SICPO}
Suppose we use random search to solve the inner-loop problem and Assumption~\ref{Assumption_regular_maxima} holds. 
If we set $\alpha=1/\sqrt{T}$, $\eta=\epsilon+\frac{1}{(1-\gamma)^{3/2}}\sqrt{\left\|\frac{\nu^*}{\nu_0}\right\|_\infty\epsilon_{bias}}$, and 
${K_{sgd}=\widetilde{O}\left(\frac{1}{\epsilon_{bias}^2(1-\gamma)^4}\right)}$ , $ H=O\left(\frac{\log(1-\gamma)+\min\{\log(\epsilon_{bias}),\log(\epsilon)\}}{\log\gamma}\right)$, $K_{eval}=\widetilde{O}\left(\frac{1}{\epsilon^2(1-\gamma)^2}\right)$, ${M=\widetilde{O}\left(\frac{(\mathrm{diam}(Y))^m}{\epsilon^m(1-\gamma)^m}\right)}$, ${T=\widetilde{O}\left(\frac{1}{\epsilon^2(1-\gamma)^6}\right)}$, then we have with probability $1-2\delta$,
$$
    \frac{1}{|\gB|}\sum_{t\in\gB}(V_r^*(\mu)-V_r^{(t)}(\mu))\leq \epsilon+\frac{1}{(1-\gamma)^{3/2}}\sqrt{\left\|\frac{\nu^*}{\nu_0}\right\|_\infty\epsilon_{bias}},
$$
and $\forall t\in\gB$
$$ \sup_{y\in Y}\left[V_{c_y}^{(t)}(\mu)-u_y\right]\leq 2\epsilon+\frac{1}{(1-\gamma)^{3/2}}\sqrt{\left\|\frac{\nu^*}{\nu_0}\right\|_\infty\epsilon_{bias}}.
$$
for any $\epsilon<2\epsilon_0 L_y/(1-\gamma)$.
\end{theorem}
\proof {Proof of Theorem~\ref{Theorem_random_search_SICPO}}
See Appendix~\ref{Appendix_Proofs_SICPO}.
\endproof

\begin{theorem}\label{Theorem_PGA_SICPO}
Suppose we use projected subgradient ascent to solve the inner-loop problem and Assumption~\ref{Assumption_concave_constraint} holds. 
If we set $\alpha=1/\sqrt{T}$, $\eta=\epsilon+\frac{1}{(1-\gamma)^{3/2}}\sqrt{\left\|\frac{\nu^*}{\nu_0}\right\|_\infty\epsilon_{bias}}$, and 
${K_{sgd}=\widetilde{O}\left(\frac{1}{\epsilon_{bias}^2(1-\gamma)^4}\right)}$ , $ H=O\left(\frac{\log(1-\gamma)+\min\{\log(\epsilon_{bias}),\log(\epsilon)\}}{\log\gamma}\right)$, $K_{eval}=\widetilde{O}\left(\frac{1}{\epsilon^2(1-\gamma)^2}\right)$, ${T_{PGA}=O\left(\frac{[\mathrm{diam}(Y)]^2}{\epsilon^2(1-\gamma)^2}
    \right)}$, ${T=\widetilde{O}\left(\frac{1}{\epsilon^2(1-\gamma)^6}\right)}$, then we have with probability $1-\delta$,
$$
    \frac{1}{|\gB|}\sum_{t\in\gB}(V_r^*(\mu)-V_r^{(t)}(\mu))\leq \epsilon+\frac{1}{(1-\gamma)^{3/2}}\sqrt{\left\|\frac{\nu^*}{\nu_0}\right\|_\infty\epsilon_{bias}},
$$
and $\forall t\in\gB$
$$ \sup_{y\in Y}\left[V_{c_y}^{(t)}(\mu)-u_y\right]\leq 2\epsilon+\frac{1}{(1-\gamma)^{3/2}}\sqrt{\left\|\frac{\nu^*}{\nu_0}\right\|_\infty\epsilon_{bias}}.
$$
\end{theorem}
\proof {Proof of Theorem~\ref{Theorem_random_search_SICPO}}
See Appendix~\ref{Appendix_Proofs_SICPO}.
\endproof

In our proof, we focus on the event that the policy evaluation subroutine returns accurate estimates of $V^{(t)}_{c_y}(\mu)$ and the sample-based NPG generates a near-optimal solution of $\min_w E^{\nu^{(t)}}(\diamond,\theta^{(t)},w)$.
We show that this event happens with high probability.
When it happens, with carefully chosen tolerance threshold $\eta$, either the "good set" $\gB$ is large or the policies in $\gB$ perform equally well to the optimal policy $\pi^*$ on average, \textit{i.e.} $\sum_{t\in\gB}(V^{(t)}_r(\mu)-V^*_r(\mu))\geq 0$.
As long as $\gB$ is large enough, we may further conclude that  $\frac{1}{|\gB|}\sum_{t\in\gB}|V^{(t)}_r(\mu)-V^*_r(\mu)|$ is small by typical analysis techniques of NPG \citep{agarwal2021theory}.
Recalling that the constraint violations of policies in $\gB$ are small as long as the inner-loop optimization problems are effectively solved, we complete our proof.

Our ideas of proof are similar to \cite{wei2020comirror, xu2021crpo}.
However, \cite{wei2020comirror} focused on the semi-infinitely constrained convex problems and we focus on the semi-infinitely constrained RL problems.
Moreover, their theoretical results are in the form of bounds on expectations, while ours are in the form of high probability bounds.
Our work is different with \cite{xu2021crpo} in the sense that they address finitely constrained RL problems, and restrict their analysis to two specific forms of policy parametrizations, whereas we consider general policy parametrizations.

\begin{remark}
The error terms of SI-CPO can be attributed to three sources: the function approximation error, the statistical error, and the optimization error.
When we say SI-CPO converges to the globally optimal policy $\pi^*$ at a $\widetilde {O}(1/\sqrt{T})$ rate we mean that if we use a near-perfect parameterized policy class, estimate $V^{(t)}_{c_y}(\mu)$ and the natural policy gradient with adequate data and solve the inner-loop problem with sufficient accuracy, then the averaged error term of SI-CPO has a $\widetilde O(1/\sqrt{T})$ order with high probability.
\end{remark}

\begin{remark}\label{Remark_random_search_vs_fixed_search}
    When solving the inner-loop problem, an alternative approach to random search is to search over a fixed grid of $Y$.
    This is equivalent to a version of naive discretization: we first transform the SICMDP to a finitely constrained MDP by discretizing $Y$, and then solve the resulting problem with CRPO \citep{xu2021crpo}.
    From a theoretical viewpoint, random search is no better than the gird search since both need to search over a $\widetilde{O}((\mathrm{diam}(Y)/\epsilon)^m)$-sized grid to ensure $\epsilon$-optimality.
    However, in numerical experiments we find that the approach based on random search is far more efficient that the approach based on grid search.
    The reasons can be two-fold: 1) in the theoretical analysis we must give guarantees for the hardest problem instances, but real-world problem settings may contain structures that can be exploited by random search \citep{bergstra2012random}; 2) in random search the random grid are generated in an independent way in each iteration, which can reduce the bias introduced by replacing the constraint set $Y$ with a fixed finite grid.
\end{remark}

\subsubsection{Sample complexity of SI-CPO}\label{Subsection_SICPO_Sample_Complexity}

\begin{corollary}\label{Corollary_Sample_Complexity_SICPO}
SI-CPO need to make $\widetilde O{\left(\frac{1}{\epsilon^2\min\{\epsilon^2,\epsilon_{bias}^2\}(1-\gamma)^{10}}\right)}$ interactions with the environments to ensure with high probability
$$
    \frac{1}{|\gB|}\sum_{t\in\gB}(V_r^*(\mu)-V_r^{(t)}(\mu))\leq \epsilon+\frac{1}{(1-\gamma)^{3/2}}\sqrt{\left\|\frac{\nu^*}{\nu_0}\right\|_\infty\epsilon_{bias}},
$$
and $\forall t\in\gB$
$$ \sup_{y\in Y}\left[V_{c_y}^{(t)}(\mu)-u_y\right]\leq 2\epsilon+\frac{1}{(1-\gamma)^{3/2}}\sqrt{\left\|\frac{\nu^*}{\nu_0}\right\|_\infty\epsilon_{bias}}.
$$
\end{corollary}
\proof{Proof of Corollary~\ref{Corollary_Sample_Complexity_SICPO}}
This corollary is a direct consequence of Theorem~\ref{Theorem_random_search_SICPO} as the sample complexity is of the order $T \cdot H\cdot(K_{eval}+K_{sgd})$.
Note that the sample complexity bound is independent of how we solve the inner-loop problem.
\endproof

Our sample complexity bound is of the order $\widetilde{O}\left(\frac{1}{\epsilon^4(1-\gamma)^{10}}\right)$.
This is worse than typical sample complexity bounds for sample-based NPG such as the $\widetilde{O}\left(\frac{1}{\epsilon^4(1-\gamma)^8}\right)$ in \cite{agarwal2021theory}.
This does not mean that the constrained problem is statistically harder or our bounds are loose.
The difference comes from that their goal is to ensure accuracy in expectation, while our goal is to ensure accuracy with high probability.
Therefore, in our algorithm, we need to run more SGD iterations (corresponding to larger $K_{sgd}$) to find a better estimate of the natural gradient.

\section{Numerical Experiments}\label{Section_Experiment}
We design two numerical examples: discharge of sewage and ship route planning.
Through a set of numerical experiments, we illustrate the SICMDP model and validate the efficacy of our proposed algorithms.
In particular, we find that the SICMDP framework greatly outperforms the CMDP baseline obtained by naively discretizing the original problem in modeling problems like Example~\ref{Example_Time_Evolving}, ~\ref{Example_Uncertain}.
We highlight that in the example of ship route planning the SI-CPO algorithm is adept at efficiently solving complex reinforcement learning tasks using modern deep reinforcement learning approaches.
\subsection{Discharge of Sewage}
\label{Experiment_Discharge_of_Sewage}
We consider a tabular sequential decision-making problem called discharge of sewage that is adapted from the literature of environmental science \citep{gorr1972optimal}.
Assume there are $|\gS|$ sewage outfalls in a region $[0, 1]^2$, and at each time point only one single outfall is active.
The active outfall would cause pollution in nearby areas, and the impact would decrease with Euclidean distance. 
Hence our state is the current active outfall. 
Given the current active outfall, the available actions are to switch to $|\gA|$ neighboring outfalls or do nothing.
Each switch would receive a negative reward representing the switching cost.
We need to figure out a switching policy to avoid over-pollution at each location of the region while minimizing the switching cost.
Clearly, this problem can be formulated as a SICMDP model with $Y=[0,1]^2$ and corresponding $c_y$ and $u_y$.
Specifically, we use $c_y(s, a)=c_y(s)=1/(1+\|y-s\|^2_2)$, where $s$ represents the position of the state (outfall).
Given a target state-occupancy measure $d$ we define $u_y=(1+\Delta)\sum_{s\in S} d(s)c_y(s)$, where $\Delta$ is a small positive number. 
The SICMDP would be nontrivial if we choose a suitable $\Delta$.
In the following numerical experiments, we assume that an offline dataset generated by a generative model is available.
\begin{figure}[htb]
\begin{minipage}[htb]{0.45\linewidth}
    \centering
    \includegraphics[height=60mm, width=60mm]{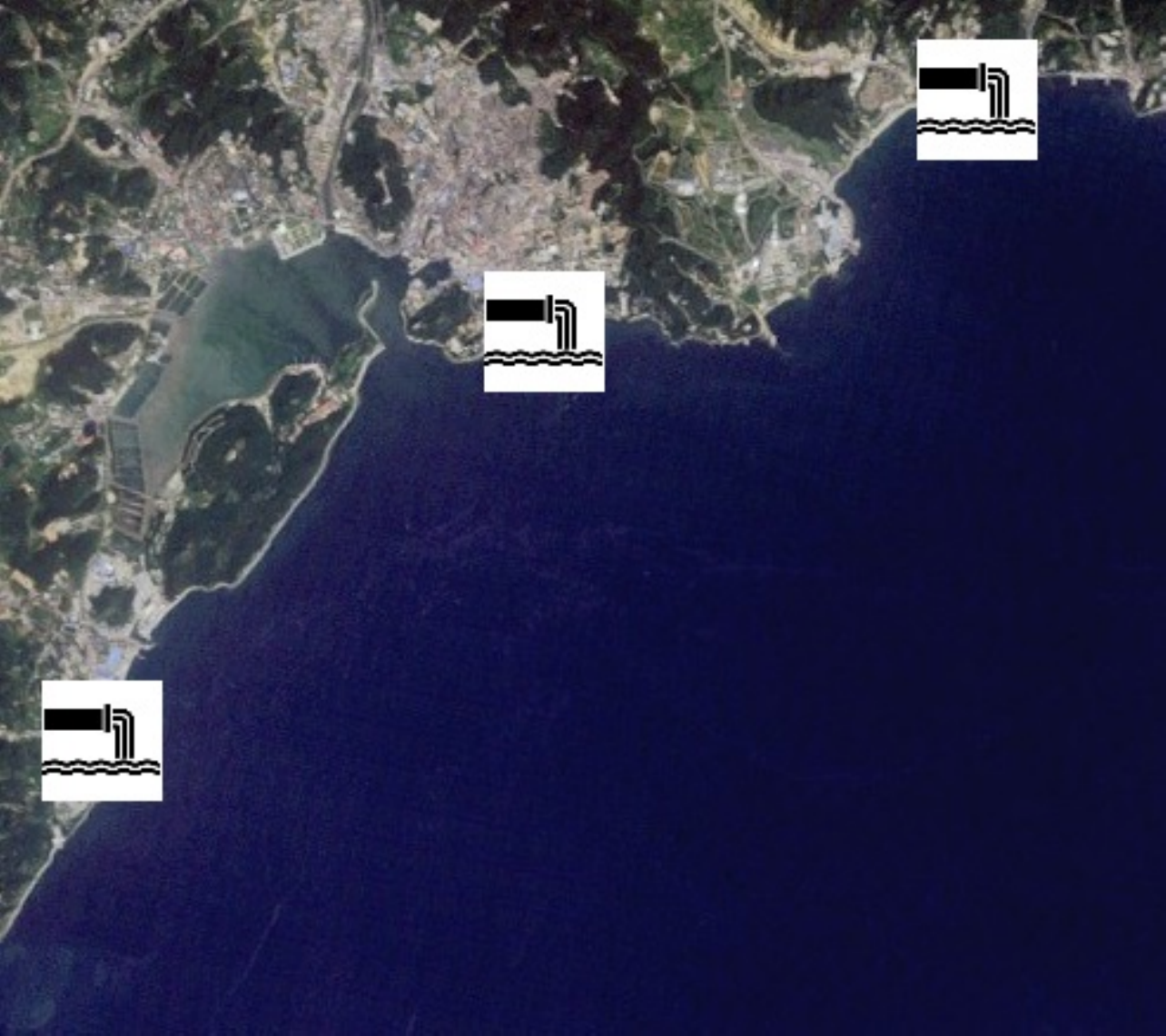}
    \caption{(Discharge of Sewage) The icons represent locations of the sewage outfalls.
    The satellite image is from NASA and only for illustrative purpose.}
    \label{Figure_Sewage_Env}  
\end{minipage}
\hspace{.15in}
\begin{minipage}[htb]{0.45\linewidth}
    \centering
    \vspace{2.9cm}
    \includegraphics[height=60mm, width=60mm]{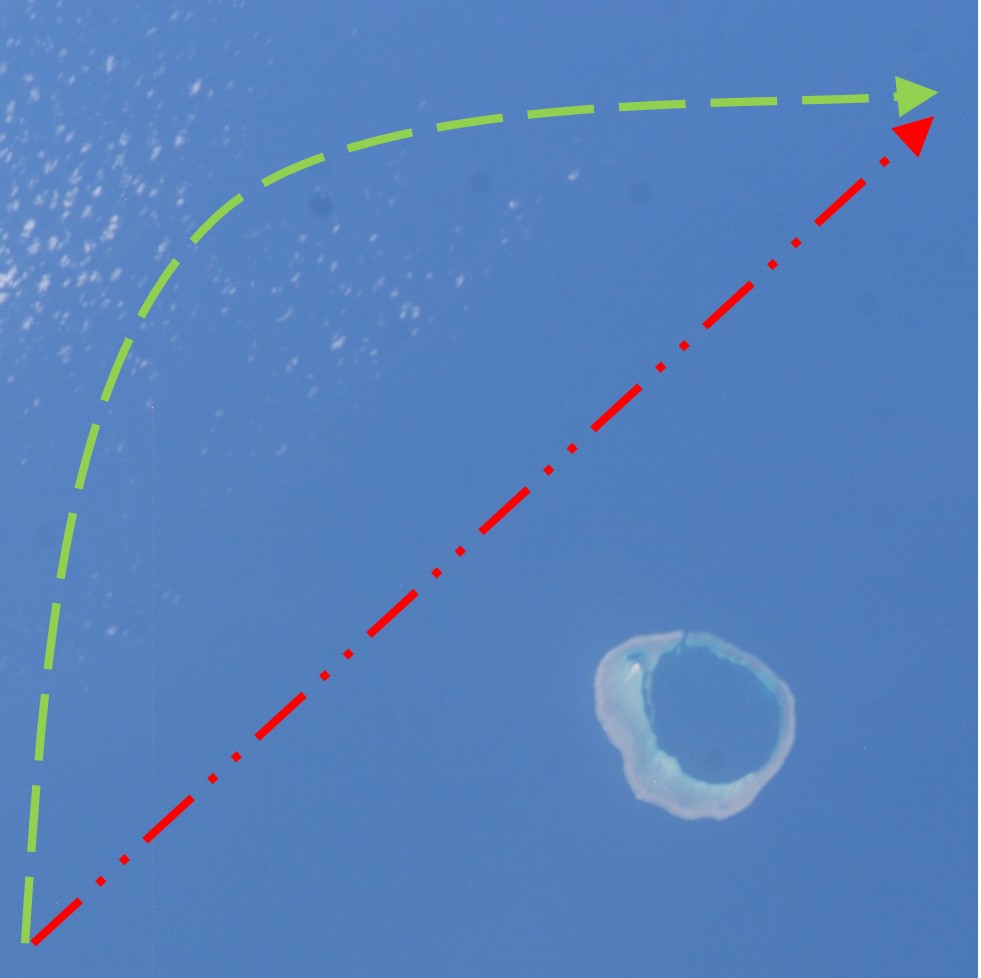}
    \caption{(Ship Route Planning) The island represents the ecological critical point. The green dashed line represents a feasible route, while the red dash-dot-dot line represents a more efficient but ecologically infeasible route.
    The satellite image is from NASA and only for illustrative purpose.}
    \label{Figure_Route_Env}  
\end{minipage}
\end{figure}

\begin{figure}[htbp]
\begin{minipage}[t]{0.45\linewidth}
    \centering
    \includegraphics[height=6cm,width=6cm]{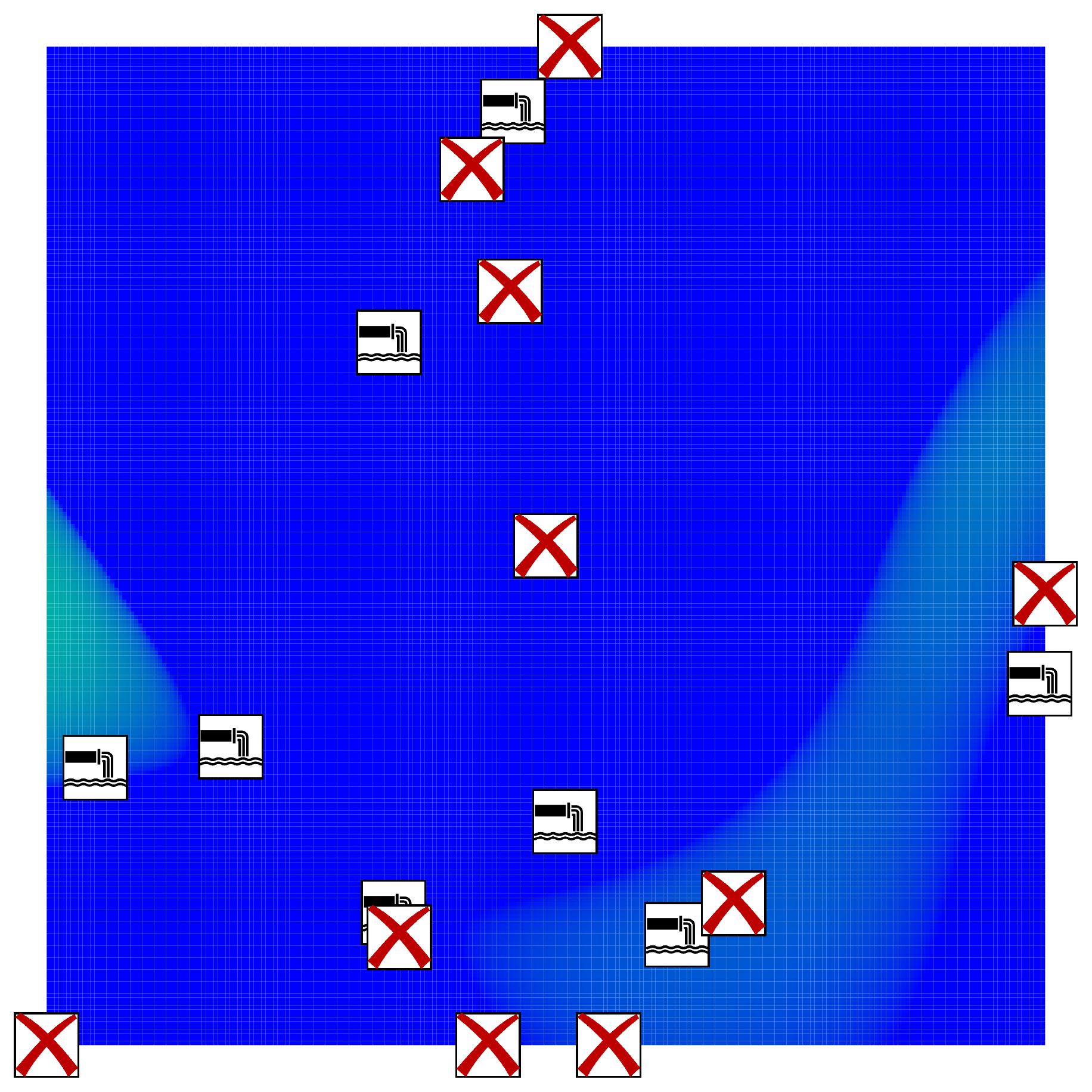}
\end{minipage}
\begin{minipage}[t]{0.45\linewidth}
    \centering
    \includegraphics[height=6cm,width=6cm]{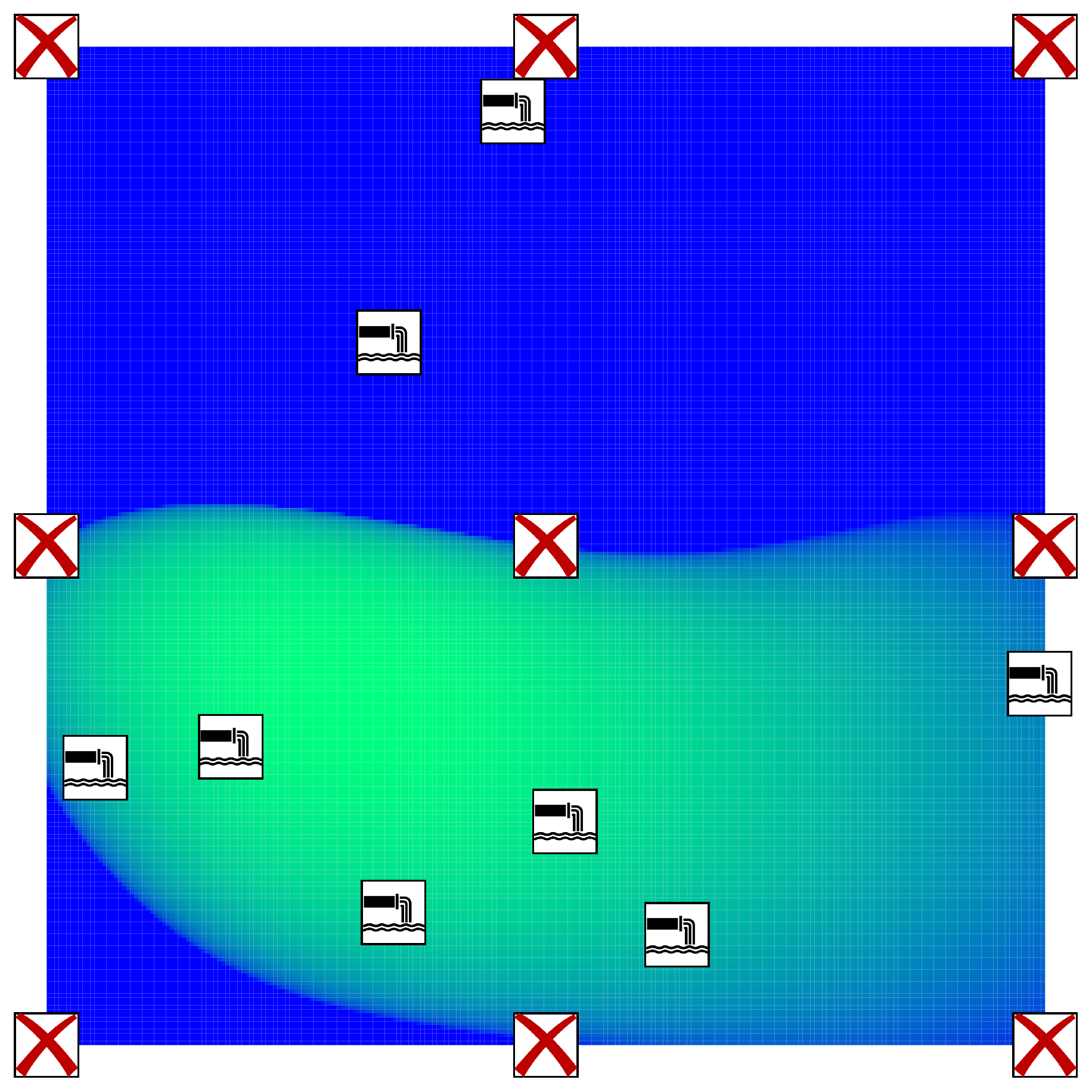}
\end{minipage}
\begin{minipage}[t]{0.08\linewidth}
    \centering
    \includegraphics[height=5.8cm,width=0.725cm]{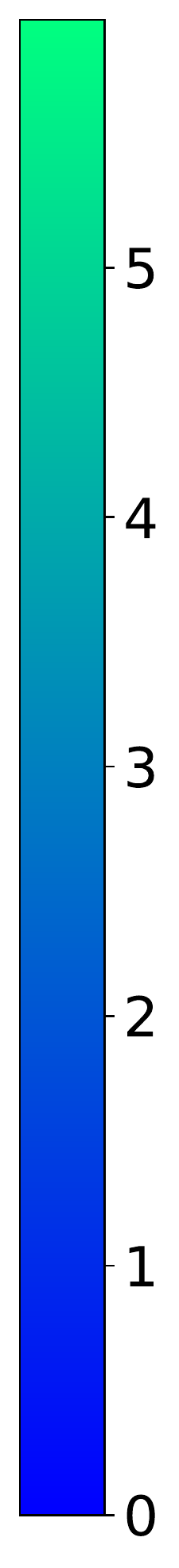}
\end{minipage}
    \caption{(Discharge of Sewage) Visualization of violation of constraints using SI-CRL (left) and baseline (right). The heat refers to the number $\log\paren*{(V^{\hat\pi}_{c_y}(\mu)-u_y)_++5\times10^{-6}}-\log(5\times10^{-6})$. Larger numbers mean a more serious violation of constraints. The red cross icons in the left two subfigures represent the $T=N_{\text{baseline}}=9$ checkpoints selected by the algorithms.}
    \label{Figure_Sewage_Heat}    
\end{figure}

First, we compare our SI-CRL algorithm with a naive discretization baseline~\ref{Remark_Baseline}.
In the baseline method, we only consider the constraints on a grid of $Y$ containing $N_{\text{baseline}}$ points, which allows us to model Discharge of Sewage as a standard CMDP problem with $N_{\text{baseline}}$ constraints.
The CMDP problem is then solved by the algorithm proposed in \cite{efroni2020explorationexploitation}.
Details of our implementation can be found in Appendix \ref{Appendix_Detials_of_Experiments}.
We visualize the quality of solutions of our proposed method and baseline method in Figure \ref{Figure_Sewage_Heat}.
It can be found that when $T=N_{\text{baseline}}$, the policy obtained by our proposed methods is of far better quality than the policy obtained by the baseline methods.

\begin{figure}[htb]
\begin{minipage}[htb]{0.48\linewidth}
    \centering
    \includegraphics[height=70mm, width=70mm]{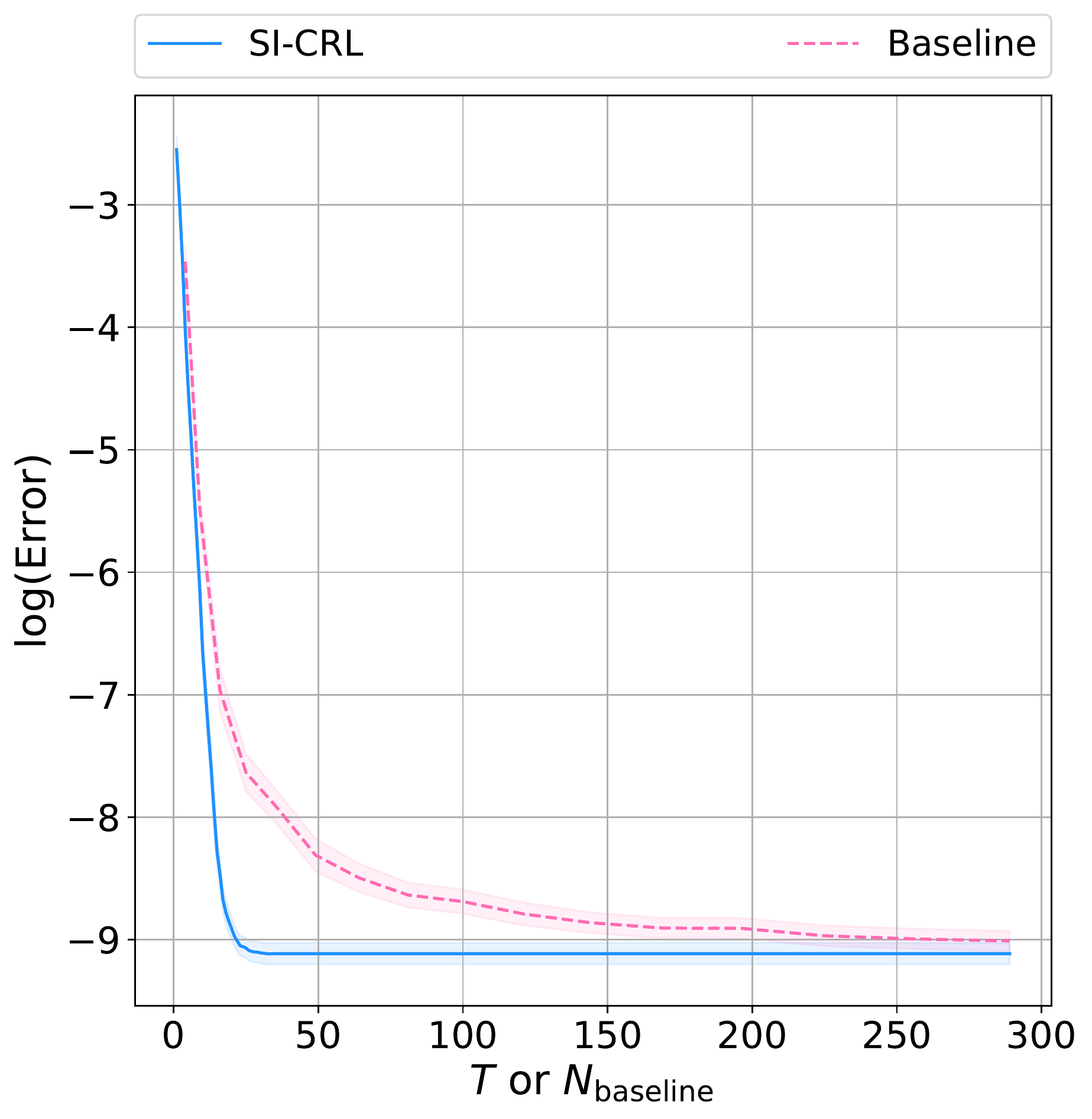}
    \caption{(Discharge of Sewage) Averaged error term of our proposed method and the baseline method over 100 seeds when $T$ and $N_{\text{baseline}}$ vary. ($\delta=\frac{0.005}{|S|^2|A|}$, $m$ sufficiently large)}
    \label{Figure_Sewage_Node}  
\end{minipage}
\hspace{.15in}
\begin{minipage}[htb]{0.48\linewidth}
    \centering
    \vspace{-0.1cm}
    \includegraphics[height=70mm, width=70mm]{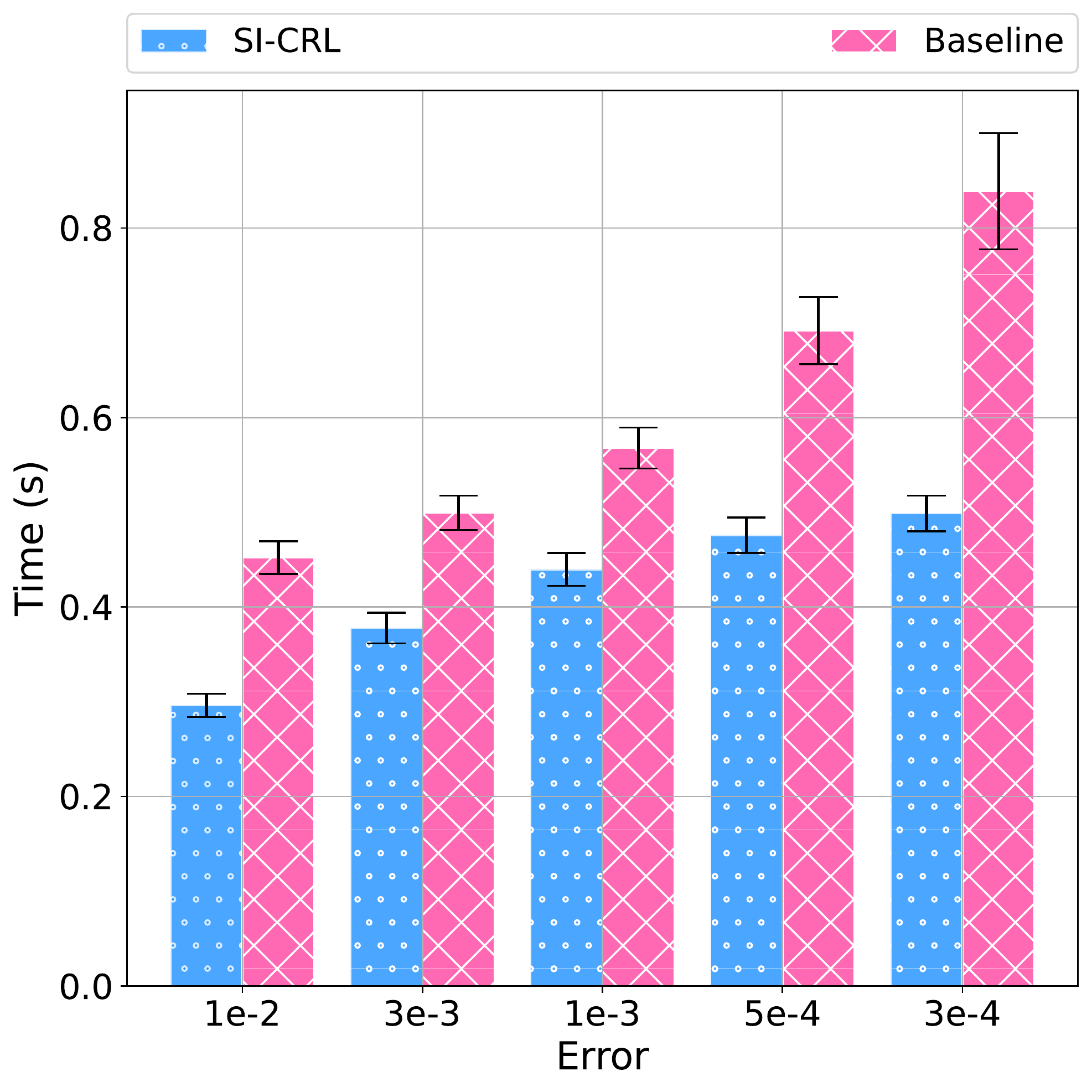}
    \caption{(Discharge of Sewage) Averaged time consumption of our method and the CMDP baseline to get a solution of given accuracy over 100 seeds. ($\delta=\frac{0.005}{|S|^2|A|}$, $m$ sufficiently large)}
    \label{Figure_Sewage_time}  
\end{minipage}
\end{figure}

An anti-intuitive phenomenon is that although in our method we need to deal with multiple LP problems while in the baseline we only solve one single LP problem, our method is still more time-efficient than the CMDP baseline.
Figure \ref{Figure_Sewage_time} indicates that our method takes less time to get a solution of given accuracy, which is evaluated by the error term $\max\brc{V^{\pi^*}_r(\mu)-V^{\hat\pi}_r(\mu),\sup_{y\in Y}V_{c_y}^{\hat \pi}(\mu)-u_y}$.
The reason is that in SI-CRL we can solve LP problems with a dual simplex method, thus re-optimization after adding a new constraint is much faster than re-solving the LP problem from scratch~\citep{koberstein2005dual}.
And our method needs far fewer active constraints to attain the same accuracy as the baseline methods, see Figure \ref{Figure_Sewage_Node}.

\begin{figure}[htbp]
\begin{minipage}[t]{0.45\linewidth}
    \centering
        \vspace{0cm}
    \includegraphics[height=6cm,width=6cm]{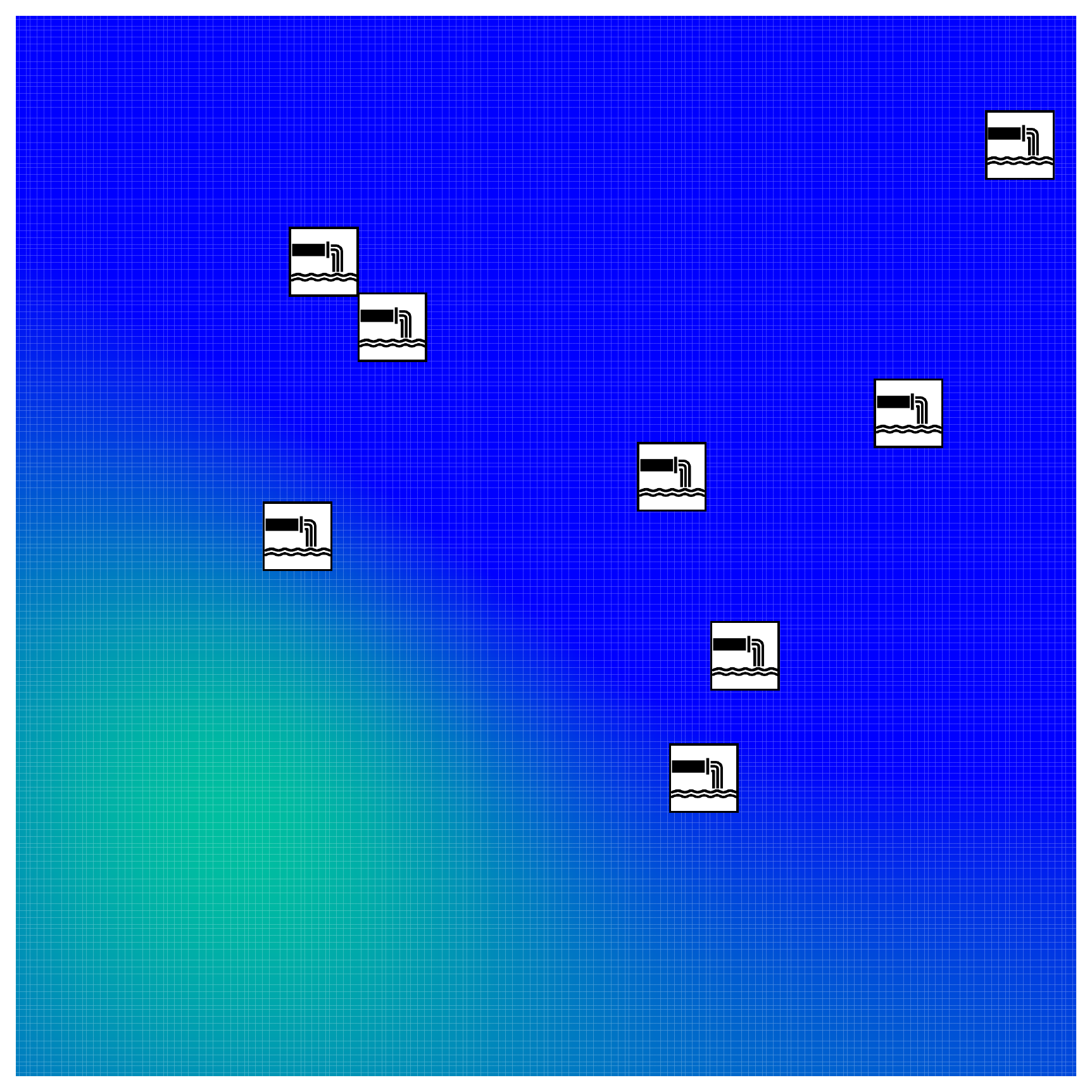}
\end{minipage}
\begin{minipage}[t]{0.45\linewidth}
    \centering
        \vspace{0cm}
    \includegraphics[height=6cm,width=6cm]{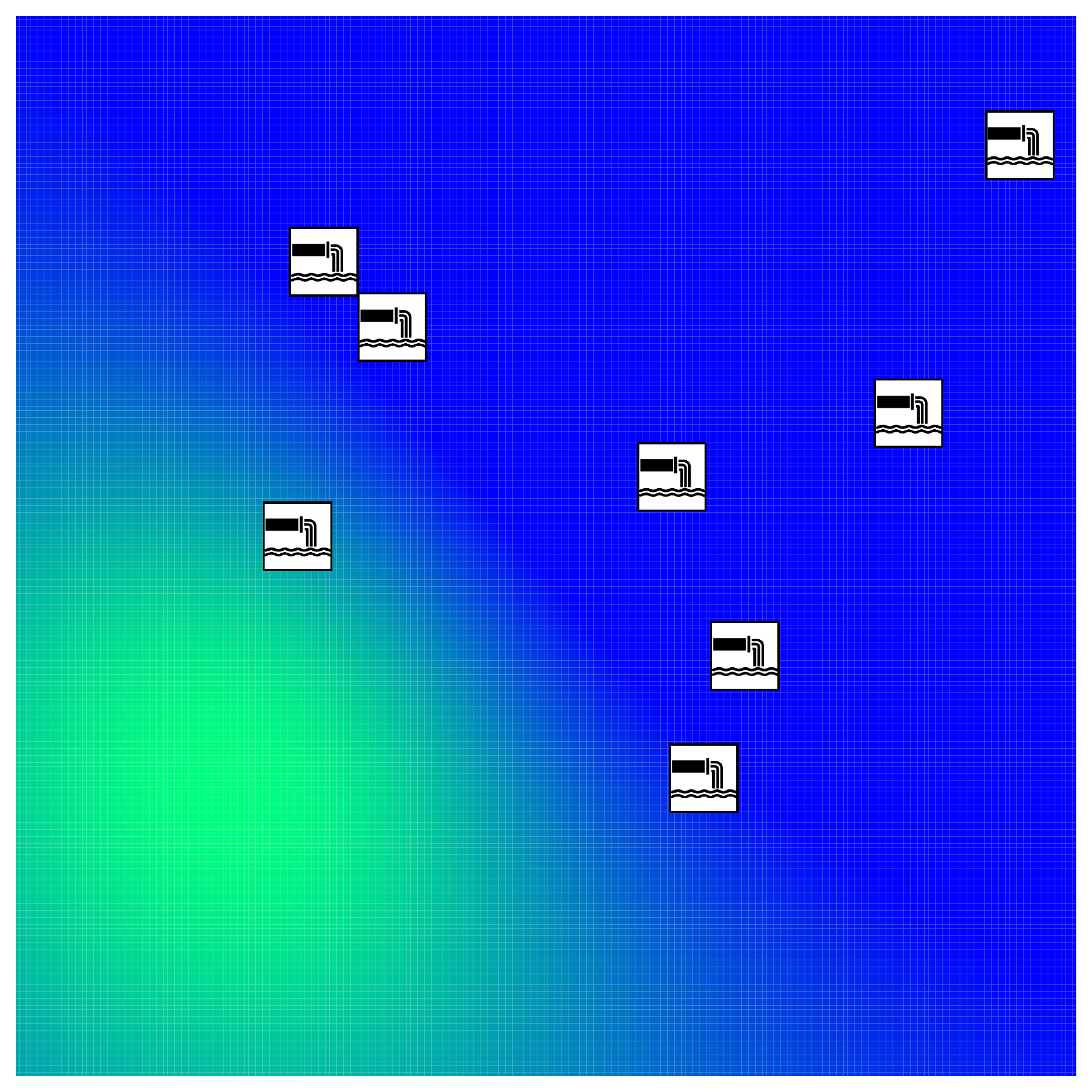}
\end{minipage}
\begin{minipage}[t]{0.08\linewidth}
    \centering
        \vspace{0cm}
    \includegraphics[height=6.1cm,width=0.7625cm]{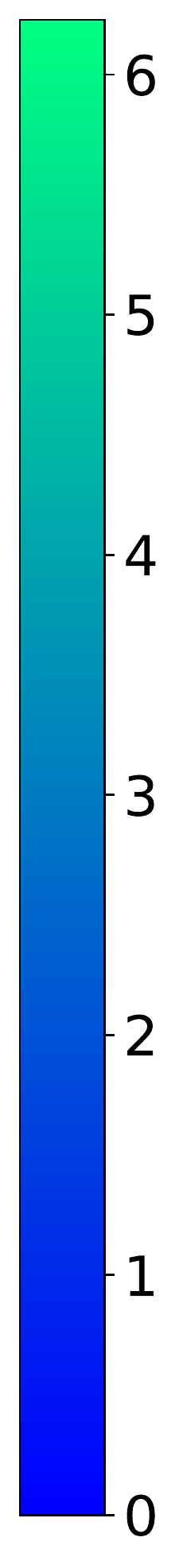}
\end{minipage}
    \caption{(Discharge of Sewage) Visualization of violation of constraints using SI-CPO (left) and the naive discretization baseline with $N_{\text{baseline}}=500$ solved by CRPO (right). The heat refers to the number $18(V^{\hat\pi}_{c_y}(\mu)-u_y)_+$. Larger numbers mean a more serious violation of constraints.}
    \label{Figure_Sewage_Heat_PG}    
\end{figure}

\begin{figure}[htb]
\begin{minipage}[htb]{0.96\linewidth}
    \centering
    \includegraphics[height=80mm, width=80mm]{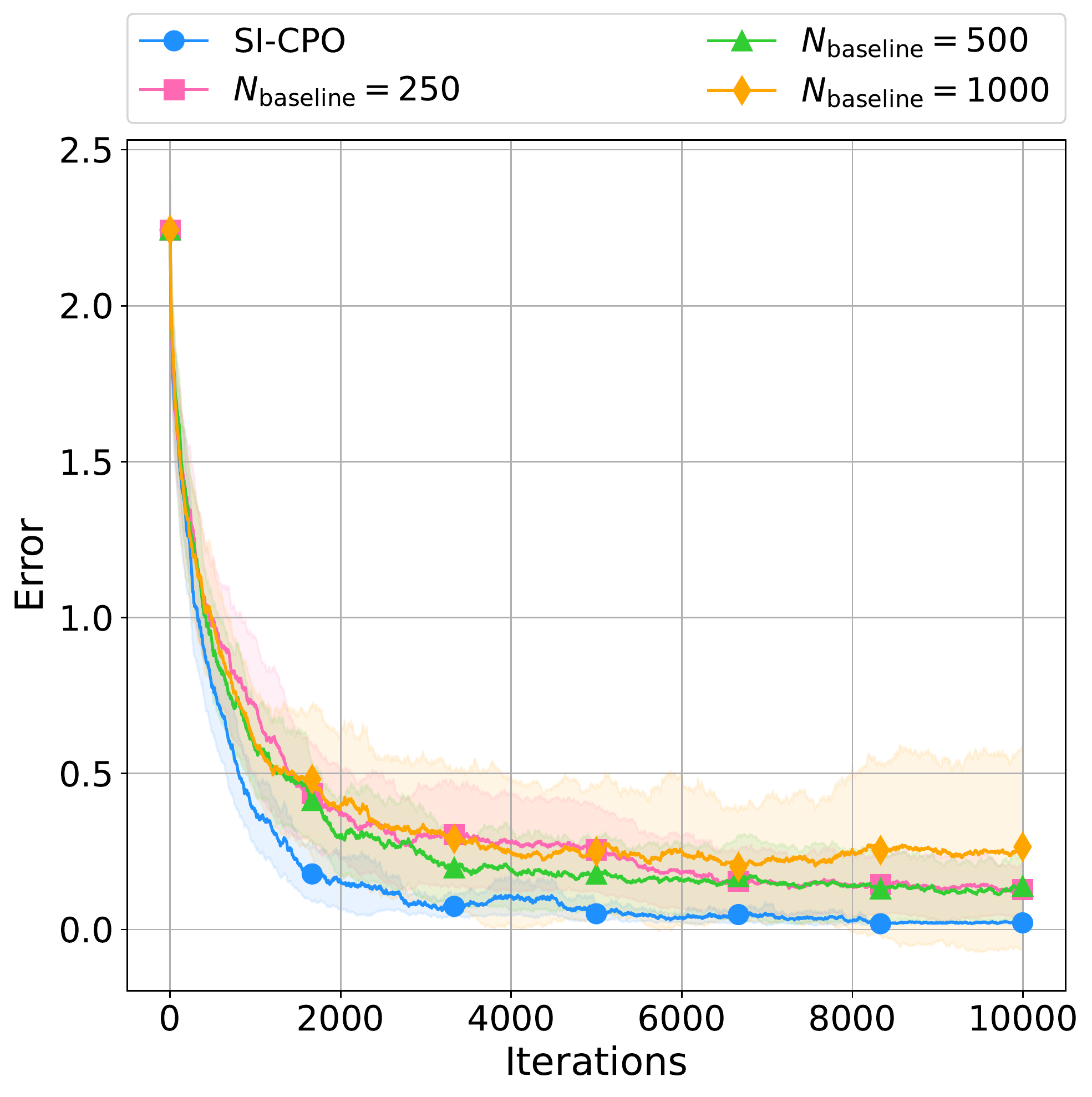}
    \caption{(Discharge of Sewage) Error term of SI-CPO and baselines versus the number of iterations.
    The solid line is the error term averaged over 20 random seeds.
    And we also provide the according error bars.}
    \label{Figure_Sewage_SICPO_Node}  
\end{minipage}
\end{figure}

We also compare our SI-CPO algorithm with the pre-mentioned discretization baseline.
In this numerical experiment, we use softmax policy parametrization.
Our SI-CPO algorithm is instantiated with sample-based NPG as the policy optimization subroutine, a finite-horizon Monte-Carlo estimator as the policy evaluation subroutine, and random search with a grid of 100 points as the optimization subroutine.
Here the size of the random grid in each iteration is 100. 
For a fairer comparison, here the CMDP resulting from discretizing $Y$ is solved by CRPO \citep{xu2021crpo}.
This is also equivalent to a naive version of our SI-CPO algorithm where the inner problem is solved by searching over a fixed grid. (See Remark~\ref{Remark_random_search_vs_fixed_search}).
One may find the details of the implementation of our methods as well as the baselines in Appendix \ref{Appendix_Detials_of_Experiments}.

\begin{table}[t]
    \centering
\begin{tabular}{ccccc}
\hline
 &SI-CPO  & $N_{\text{baseline}}=250$ & $N_{\text{baseline}}=500$ & $N_{\text{baseline}}=1000$ \\
 \hline
time per iteration (s) & $0.19\pm 0.02$ & $0.23\pm 0.02$ & $0.30\pm 0.03$ & $0.45\pm 0.04$\\
\hline
\end{tabular}
\caption{(Discharge of Sewage) Time consumption of each iteration in SI-CPO and baselines.}
\label{Table_time_discharge}
\end{table}

The visualization of the solutions' quality can be found in Figure~\ref{Figure_Sewage_Heat_PG}, which shows that the policy obtained by SI-CPO is better than the policy obtained by the baseline solved by CRPO.
In Figure~\ref{Figure_Sewage_SICPO_Node} we compare the convergence performance of SI-CPO to baselines that naively discretize $Y$ into grids with different sizes (different $N_{\text{baseline}}$s).
The resulting CMDPs are also solved by CRPO.
We may observe that the SI-CPO algorithm achieves a more rapid convergence measured by the number of iterations than all the naive discretization baselines no matter how large the grid is.
Also, Table~\ref{Table_time_discharge} suggests that the time consumption of a single iteration of SI-CPO is comparable to baseline methods.

\subsection{Ship Route Planning}
To demonstrate the power of the SICMDP model and our proposed algorithms, we design a more complex continuous control problem with continuous state space named ship route planning.
This numerical example tackles a challenging task in maritime science \citep{wan2016four, wan2016pollution}: planning ship routes while ensuring their negative environmental impacts under an adaptive threshold.
Consider a ship sailing in a 2-dimensional area represented by the unit square $[0,1]^2$. 
At each time step $t$, the state of the ship is represented by its current position $s_t\in[0,1]^2$ and the action it takes is represented by the next heading angle $a_t\in[0,2\pi)$.
Given an outset $O\in [0,1]^2$ and a destination $D\in[0,1]^2$, at each time step $t$, we receive a negative reward $r(s_t)=-0.1 \times (\|s_t-D \|_2+1)$, and after we arrive at $D$ we will receive a large positive reward $5$.
The most efficient route is apparently a straight line.
However, we must take into account additional environmental concerns.
Specifically, the ship positioned at $s$ would cause pollution $c_y(s)=e^{-20\|y-s\|_2}$ to position $y$.
$c_y$ is designed to account for the greater pollution impact on areas closer to the ship.
The adaptive threshold of pollution is defined by $u_y=0.015+0.005\times e^{20\|y-\text{MPA}\|_2}$, where $\text{MPA}\in [0, 1]^2$ is an environmentally critical point that has special ecological significance, such as a habitat of endangered species or a natural heritage priority site.
The design of $u_y$ reflects the principle that we implement more strict pollution restrictions for nearer positions from the environmentally critical point $\text{MPA}$.
We would like to complement that due to the existence of a terminal state (the ship's destination), we set the discount factor $\gamma=1$.
Figure~\ref{Figure_Route_Env} provides a visual explanation of this numerical example.

We study the performance of an actor-critic version of SI-CPO called SI-CPPO in this example.
In SI-CPPO, the policy optimization subroutine is PPO, the policy evaluation subroutine is TD-learning and the optimization subroutine is a trust-region method \citep{conn2000trust}.
Both the policy and the value estimator are parametrized by deep neural networks.
We still consider the naive discretization baseline where the CMDPs resulting from discretization are solved by CRPO.
The implementation details of SI-CPPO and the baseline can be found in Appendix \ref{Appendix_Detials_of_Experiments}.
Figure~\ref{Figure_Route_Heat} is a visualization of the solutions attained via SI-CPPO and the discretization baseline.
While the baseline fails to generate a feasible route, SI-CPPO manages to plan a route that is both feasible and efficient.
We demonstrate the convergence performance of SI-CPPO and baselines with various $N_{\text{baseline}}$ in Figure~\ref{Figure_Route_Reward} and Figure~\ref{Figure_Route_Violat}.
It is shown that the convergence of baselines is very slow and the curves oscillate a lot.
And simply increasing $N_{\text{baseline}}$ does not help.
In contrast, our SI-CPPO algorithm rapidly converges to the optimal solution.
Also, Table~\ref{Table_time_route} shows that a single iteration of SI-CPPO consumes a similar amount of time compared with baseline methods.

\begin{table}[t]
    \centering
\begin{tabular}{ccccc}
\hline
 &SI-CPPO  & $N_{\text{baseline}}=250$ & $N_{\text{baseline}}=500$ & $N_{\text{baseline}}=1000$ \\
 \hline
time per iteration (s) & $11.39\pm 1.43$ & $8.91\pm 1.50$ & $10.05\pm 1.72$ & $11.78\pm 1.74$\\
\hline
\end{tabular}
\caption{(Discharge of Sewage) Time consumption of each iteration in SI-CPO and baselines.}
\label{Table_time_route}
\end{table}

\begin{figure}[htbp]
\begin{minipage}[t]{0.45\linewidth}
    \centering
    \vspace{0cm}
    \includegraphics[height=6cm,width=6cm]{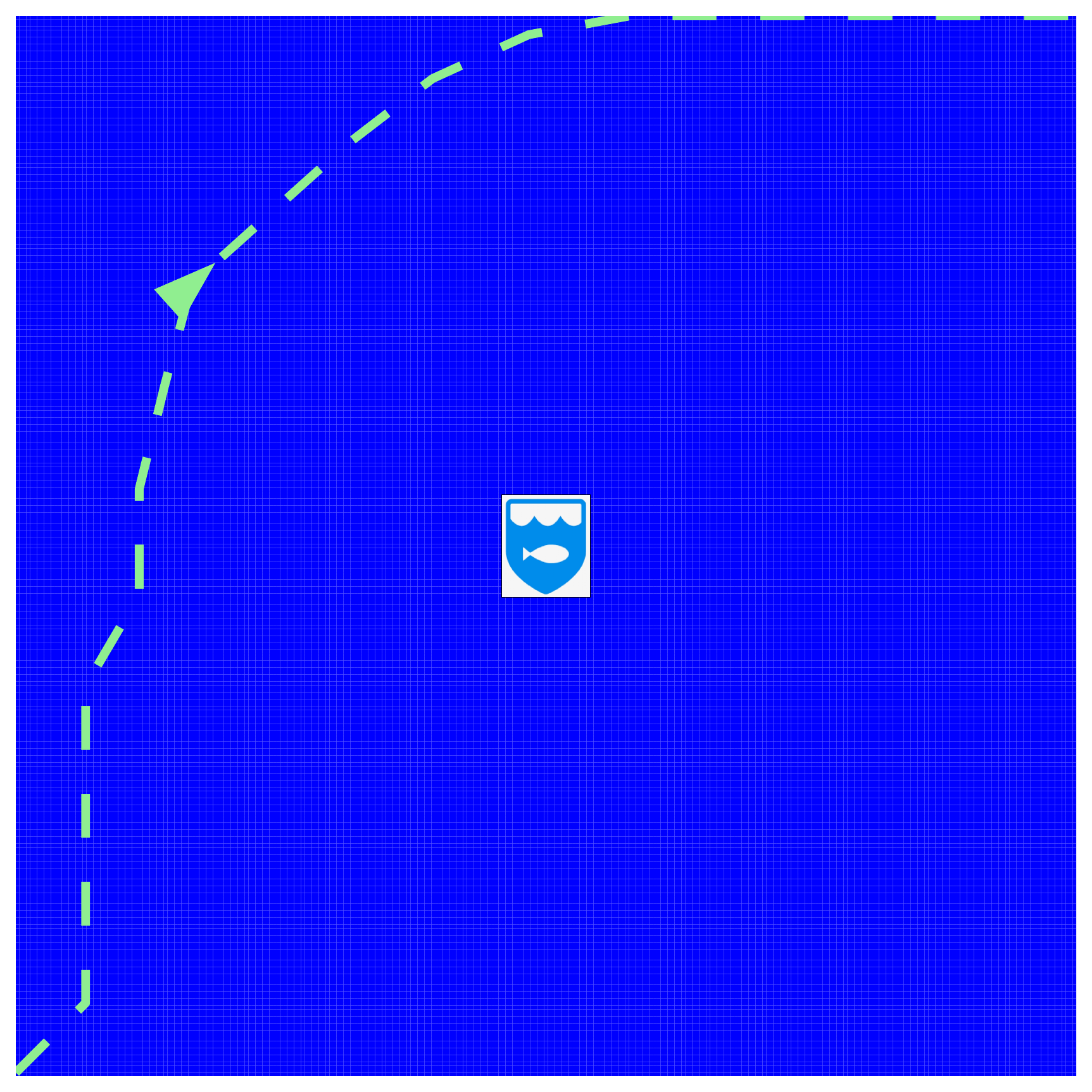}
\end{minipage}
\begin{minipage}[t]{0.45\linewidth}
    \centering
    \vspace{0cm}
    \includegraphics[height=6cm,width=6cm]{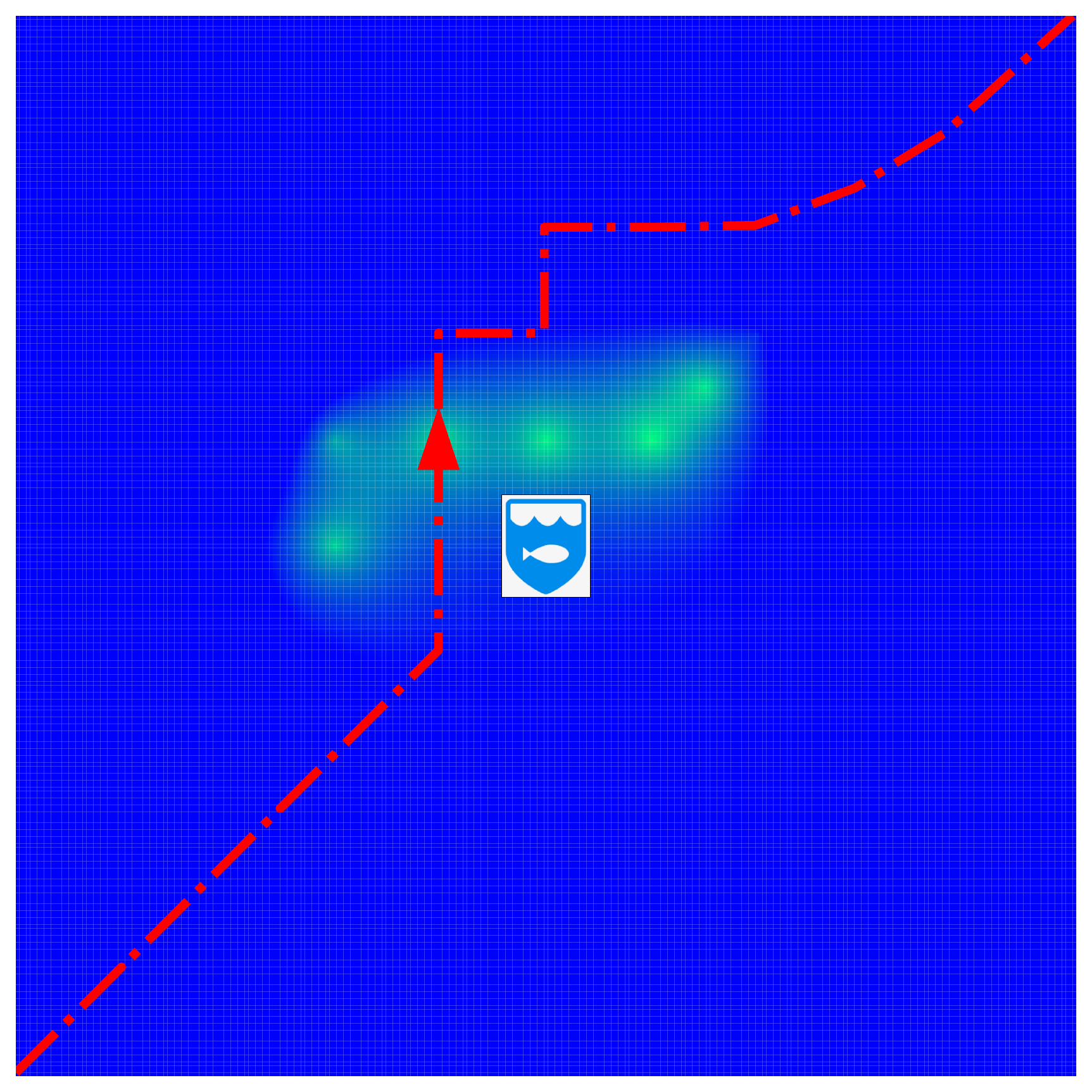}
\end{minipage}
\begin{minipage}[t]{0.08\linewidth}
    \centering
    \vspace{0cm}
    \includegraphics[height=6.1cm,width=0.7625cm]{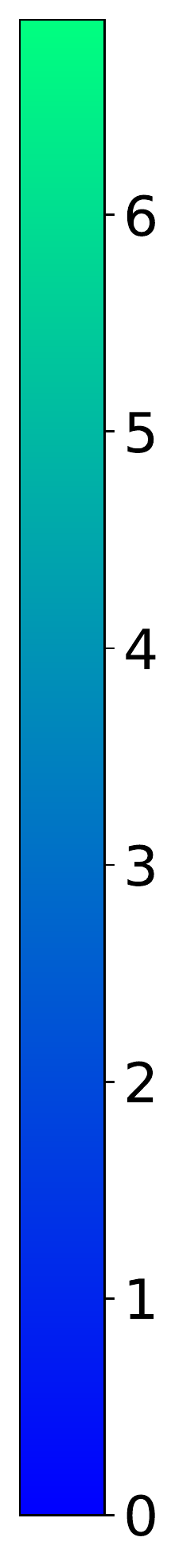}
\end{minipage}
    \caption{(Ship Route Planning) Visualization of routes and violation of constraints using SI-CPPO (left) and naive discretization with $N_{\text{baseline}}=1000$ (right). The heat refers to the number $5(V^{\hat\pi}_{c_y}(\mu)-u_y)_+$. Larger numbers mean a more serious violation of constraints. The green dashed line represents a feasible route induced by the SI-CPPO policy, while the red dash-dot line represents an infeasible route induced by the baseline policy. The blue icons in the center represent the ecologically critical points.}
    \label{Figure_Route_Heat}    

\begin{figure}[H]
\begin{minipage}[htb]{0.48\linewidth}
    \centering
    \includegraphics[height=70mm, width=70mm]{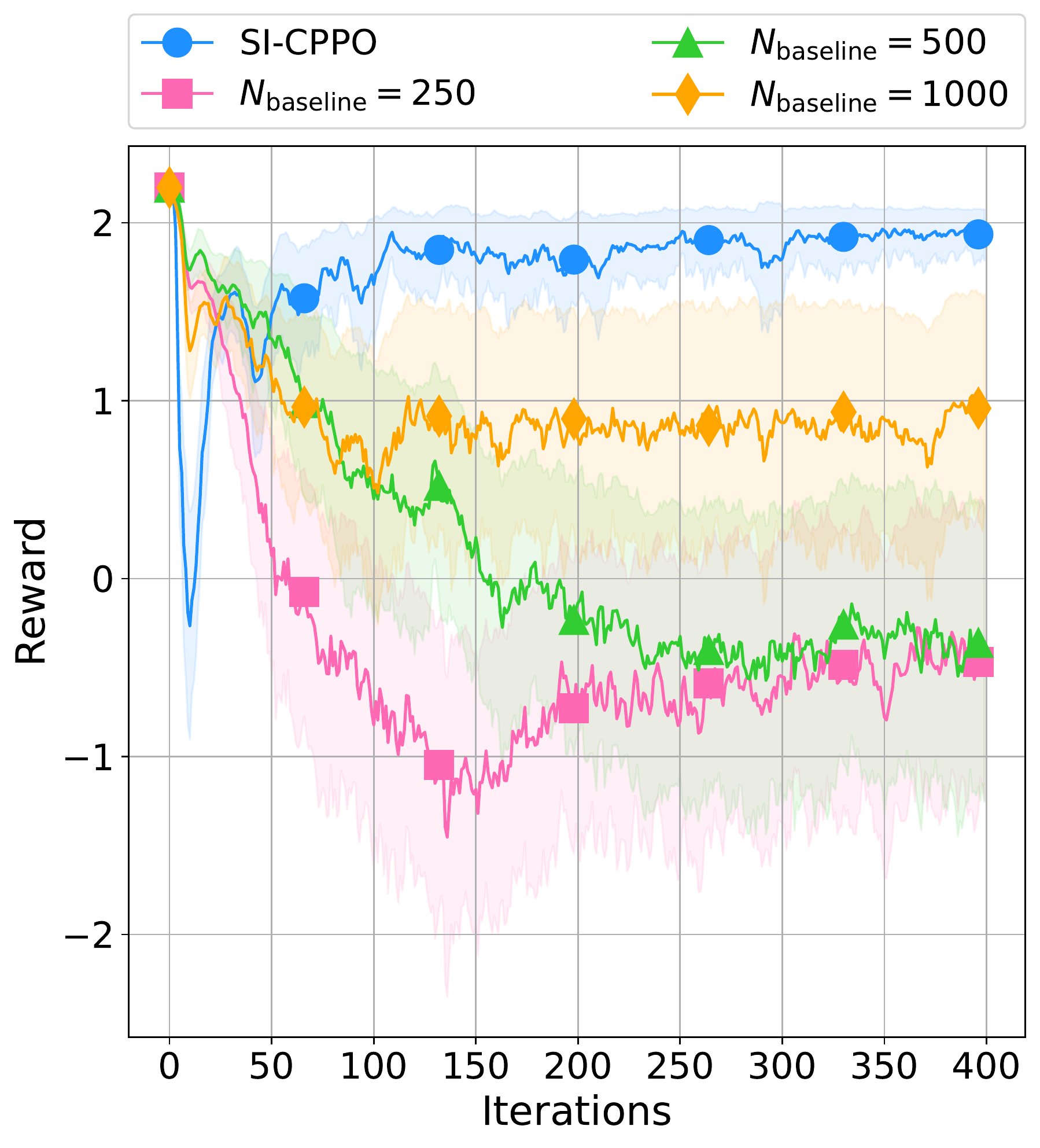}
    \caption{(Ship Route Planning) Cumulative reward of SI-CPPO and baselines versus the number of iterations.
    The solid line is the cumulative reward averaged over 20 random seeds.
    And we also provide the according error bars.}
    \label{Figure_Route_Reward}  
\end{minipage}
\hspace{.15in}
\begin{minipage}[htb]{0.48\linewidth}
    \centering
    \includegraphics[height=70mm, width=70mm]{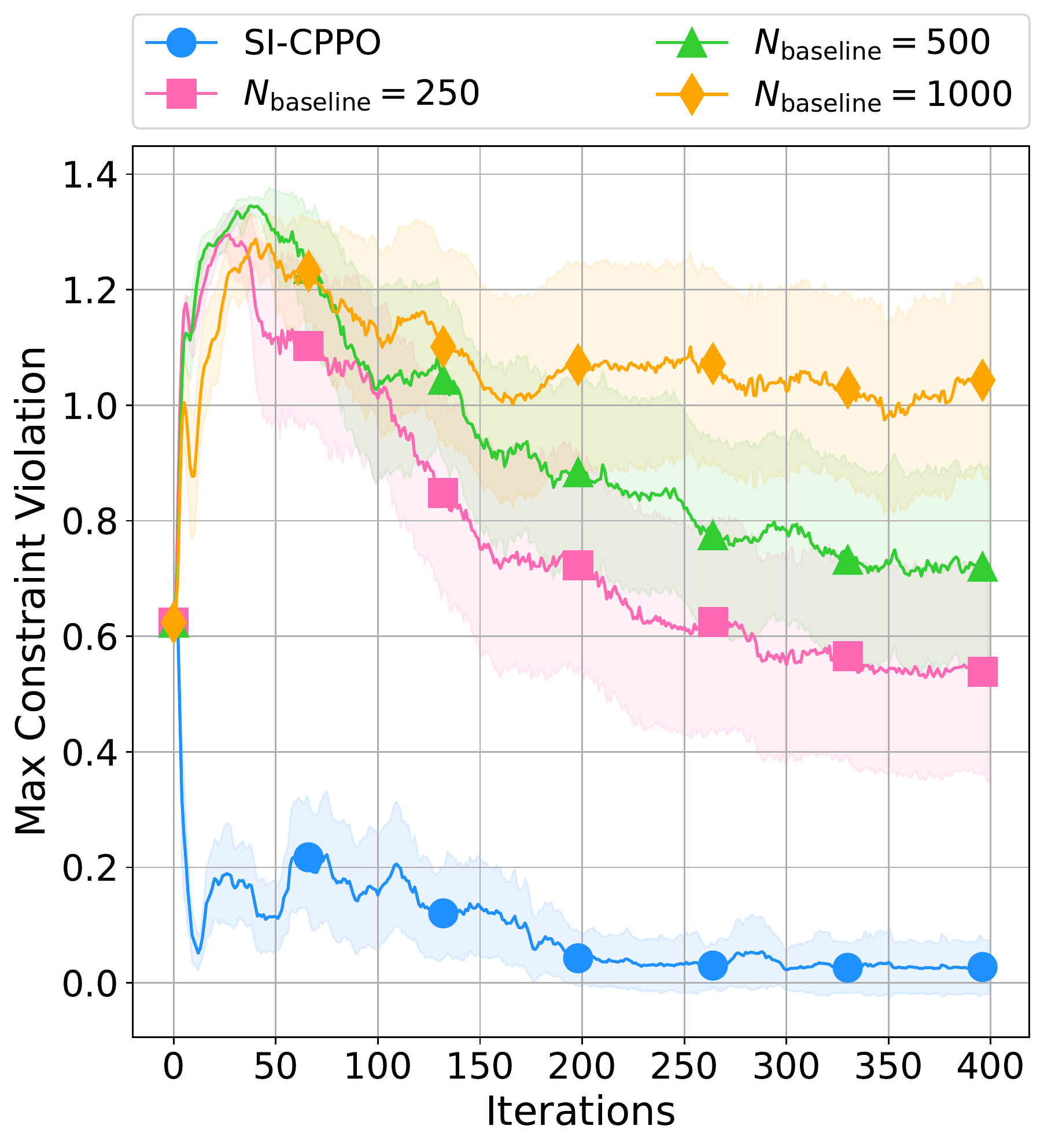}
    \caption{(Ship Route Planning) Maximum constraint violation of SI-CPPO and baselines versus the number of iterations.
    The solid line is the maximum constraint violation averaged over 20 random seeds.
    And we also provide the according error bars.}
    \label{Figure_Route_Violat}  
\end{minipage}
\end{figure}
\end{figure}

\section{Conclusion}
We have studied a novel generalization of CMDP that we have called SICMDP. In particular, we have considered a continuum of constraints rather than a finite number of constraints.
We have devised two reinforcement learning algorithms SI-CRL and SI-CPO to solve SICMDP problems.
Furthermore, we have presented theoretical analysis for our proposed algorithms, establishing the iteration complexity bounds as well as the sample complexity bounds.
We have also performed the extensive numerical experiments to show the efficacy of our proposed methods and its advantage over traditional CMDPs.

\section{Acknowledgement}
The authors would like to thank Mr. Hao Jin for helpful discussions.

\bibliography{ref}
\bibliographystyle{abbrvnat}
\newpage

\newpage
\newpage
\appendix
\section{Omitted Proofs in Section \ref{Section_SICRL}}
\label{Appendix_Proofs_4}
\proof{Proof of Theorem \ref{Theorem_Feasible}.}
By Lemma \ref{Theorem_Empirical_Bernstein}, 
$$
\PB\paren{|P(s^\prime|s,a)-\hat P(s^\prime|s,a)|\leq \sqrt{\frac{2\hat P(s^\prime|s,a)(1-\hat P(s^\prime|s,a))\log4/\delta}{n}}+\frac{4\log 4/\delta}{n}}\geq 1-\delta.
$$
By Lemma \ref{Theorem_Hoeffding_Inequality}, 
$$
\PB\paren{|P(s^\prime|s,a)-\hat P(s^\prime|s,a)|\leq \sqrt{\frac{\log 2/\delta}{2n}}}\geq  1-\delta.
$$
Combining the two inequalities in a union bound, we have:
$$
\PB\paren{|P(s^\prime|s,a)-\hat P(s^\prime|s,a)|\leq d_\delta(s,a,s^\prime)}\geq  1-2\delta.
$$
Again we apply the union bound argument to get:
$$
\PB(M\in M_\delta)=\PB\paren{|P(s^\prime|s,a)-\hat P(s^\prime|s,a)|\leq d_\delta(s,a,s^\prime),\forall s,s^\prime\in \gS,a\in \gA}\geq  1-2|\gS|^2|\gA|\delta.
$$
Finally, Problem (\ref{Problem_Optimistic}) is feasible as long as $P\in M_\delta$ because of Assumption $\ref{Assumption_Feasible}$.
\endproof
\section{Omitted Proofs in Section \ref{Section_Theory_SICRL}}
\label{Appendix_Proofs_SICRL}
First, we define some additional notations. 
Given a stationary policy $\pi$, we define the value function $V_\diamond^\pi(s)=\EB\paren{\sum_{t=0}^\infty \gamma^t r(s_t,a_t)|s_0=s}$, $V_\diamond^\pi=(V_\diamond^\pi(s_1), \ldots, V_\diamond^\pi(s_{|\gS|}))^\top\in \RB ^{|\gS|}$.
Thus we have $V^\pi_\diamond(\mu)=\mu^\top V_\diamond^\pi$.
Here $\diamond$ represents either the reward $r$ or cost $c_y$.
We use $Q_\diamond^\pi(s,a):=\EB\paren{\sum_{t=0}^\infty \gamma^t \diamond(s_t,a)}$ and $Q_\diamond^\pi=(Q_\diamond^\pi(s_1,a_1),...,Q_\diamond^\pi(s_{\gS},a_{|\gA|}))\in\RB^{|\gS|\cdot|\gA|}$ to denote the state-action value function. 
The local variance is defined as $\Var_P(V_\diamond^\pi)(s,a)=\EB_{s^\prime\sim P(\cdot|s,a)}(V_\diamond^\pi(s^\prime)-P(\cdot|s,a)V_\diamond^\pi)^2$.
We view $\Var_P(V^\pi)$ as vectors of length $|\gS|\cdot|\gA|$. 
We overload notation and let $P$ also refer to a matrix of size $(|\gS|\cdot |\gA|)\times |\gS|$, where the entry $P_{(s, a), s^{\prime}}$ is equal to $P(s^\prime|s,a)$. 
We also define $P^\pi$ to be the transition matrix on state-action pairs induced by a stationary policy $\pi$, namely:
$$P_{(s, a),\left(s^{\prime}, a^{\prime}\right)}^{\pi}:=P\left(s^{\prime}| s, a\right) \pi\left(a^{\prime} |s^{\prime}\right).
$$
We use $\widetilde V_\diamond^{\pi}(s), \widetilde Q_\diamond^\pi(s,a), {\Var_{\widetilde P}}(\widetilde{V}_\diamond^\pi)(s,a), \widetilde V_\diamond^\pi, \widetilde Q_\diamond^\pi,{\Var_{\widetilde P}}(\widetilde{V}^\pi), \widetilde P, \widetilde P^\pi$ to denote the value function, state-action value function, local variance, vector of the value function, vector of the state-action value function, vector of local variance, transition matrix, transition matrix on state-action pairs w.r.t. SICMDP $\widetilde M$, respectively.

\begin{lemma}\label{Lemma_Iteration_Comlexity_General}
    Suppose for all $t\in\{1,...,T\}$, 
    $$
    \frac{1}{1-\gamma}\sum_{s, a,s^\prime}z^{(t)}(s,a,s^\prime)c_{y^{(t)}}(s,a)- u_{y^{(t)}}\geq \max_{y\in Y}\left[\frac{1}{1-\gamma}\sum_{s, a,s^\prime}z^{(t)}(s,a,s^\prime)c_{y}(s,a)- u_{y}\right]-\epsilon,
    $$
    Then if we set $\eta=\epsilon$ and $T=O\left(\left[\frac{\mathrm{diam}(Y)|\gS|^2|\gA|}{(1-\gamma)\epsilon}\right]^m \right)$,
    then SI-CRL is guaranteed to output a $2\epsilon$-optimal solution of Problem~\ref{Problem_Optimistic_ELSIP}.
\end{lemma}
\proof{Proof of Lemma~\ref{Lemma_Iteration_Comlexity_General}}
For the convenience of presentation, then the problem can be written as
$$
\begin{aligned}
    \max_{z\in Z}\ &z^\top r\\
    \text{s.t.}\ &z^\top c_y\leq u_y,\ \forall y\in Y.
\end{aligned}
$$
Here $r,c\in[0,1]^{|\gS|^2|\gA|}$, $Z\subset\RB^{|\gS|^2|\gA|}$ is a feasible set defined by constraints other than the semi-infinite one.
Let $f(y,z)=z^\top c_y-u_y$, we note $f(y,z)$ is Lipschitz w.r.t. $y$ and the Lipschitz coefficient is $\beta:=\frac{2|\gS|^2|\gA|L_y}{1-\gamma}$.
WLOG, we also assume $Y$ is contained in a $\|\cdot\|_\infty$ ball with radius $R$ with $R\leq \frac{\mathrm{diam}(Y)}{2}$.
At iteration $t<T$, if we have $f(y^{(t)}, z^{(t)})\leq\epsilon$, then the algorithm terminates and we obtain a $2\epsilon$-optimal solution of Problem~\ref{Problem_Optimistic_ELSIP}.
Else we have $f(y^{(t)},z^{(t)})>\epsilon$.
Since $f(z,y)$ is $\beta$-Lipshitz in y, we can conclude $\forall z$, if $f(z,y)>\epsilon$ and $f(z,y^\prime)<0$, then $\|y-y^\prime\|_\infty>\epsilon/\beta$.
Define $B^{(t)}=\{\|y-y^{(t)}\|_\infty\leq \epsilon/2\beta\}$, as $f(y^{(t)},z^{(t)})>\epsilon$ and $f(y^{(t^\prime)},z^{(t)})\leq 0$, $t^\prime=1,...,t-1$, we have $B^{(t)}\cap\left(\cup_{t^\prime=1}^{t-1} B^{(t^\prime)}\right)=\emptyset$. 
Then by induction one may conclude $\{B^{(t^\prime)},t^\prime=1,...,t\}$ forms a $\epsilon/2\beta$-packing of $Y$.
Noting the fact that the $\epsilon/2\beta$-packing number of $Y$ is less than $\left(\frac{2R\beta}{\epsilon}\right)^m$, we complete the proof.

\endproof

\proof{Proof of Theorem~\ref{Theorem_Iteration_Complexity_Random_Search}.}
Since we have Lemma~\ref{Lemma_Iteration_Comlexity_General}, we only need to ensure that with probability at least $1-\delta$, 
for all $t\in\{1,...,T\}$, 
    $$
    \frac{1}{1-\gamma}\sum_{s, a,s^\prime}z^{(t)}(s,a,s^\prime)c_{y^{(t)}}(s,a)- u_{y^{(t)}}\geq \max_{y\in Y}\left[\frac{1}{1-\gamma}\sum_{s, a,s^\prime}z^{(t)}(s,a,s^\prime)c_{y}(s,a)- u_{y}\right]-\epsilon.
    $$
We adopt the notations introduced in the proof of Lemma~\ref{Lemma_Iteration_Comlexity_General}.
At the $t$ th iteration, let $y^*:=\argmax_{y\in Y}\left[\frac{1}{1-\gamma}\sum_{s, a,s^\prime}z^{(t)}(s,a,s^\prime)c_{y}(s,a)- u_{y}\right]$.
As $f(y,z)$ is $\beta$-Lipschitz w.r.t. $y$, then it suffices to ensure that with probability at least $1-\delta/T$ there exist $i\in\{1,...,M\}$ such that $\|y_i-y^*\|_\infty\leq \epsilon/\beta$.
As long as $\epsilon/\beta\leq \epsilon_0$, we simply need
$$
\PB\left(\exists i\in\{1,...,M\}, \|y_i-y^*\|_\infty\leq \epsilon/\beta\right)=1-\left(1-\left(\frac{\epsilon}{\beta R}\right)^m\right)^M\geq 1-\frac{\delta}{T}.
$$
The proof can be completed by basic algebra operations.
\endproof

\proof{Proof of Theorem~\ref{Theorem_Iteration_Complexity_Projected_GD}.}
Since we have Lemma~\ref{Lemma_Iteration_Comlexity_General}, we only need to ensure that for all $t\in\{1,...,T\}$, 
    $$
    \frac{1}{1-\gamma}\sum_{s, a,s^\prime}z^{(t)}(s,a,s^\prime)c_{y^{(t)}}(s,a)- u_{y^{(t)}}\geq \max_{y\in Y}\left[\frac{1}{1-\gamma}\sum_{s, a,s^\prime}z^{(t)}(s,a,s^\prime)c_{y}(s,a)- u_{y}\right]-\epsilon.
    $$
We adopt the notations introduced in the proof of Lemma~\ref{Lemma_Iteration_Comlexity_General}.
By Theorem 3.2 in \cite{bubeck2015convex}, the statement above is satisfied as long as $T_{PGA}\geq \frac{\beta^2R^2}{\epsilon^2}$.
The proof can be completed by basic algebra operations.
\endproof

\begin{lemma}\label{Lemma_Crude_Bound}
If Assumption \ref{Assumption_Two_Nonzero} is true and $M\in M_\delta$, we have 
$$
\left\|Q_r^\pi-\widetilde Q_r^\pi\right\|_\infty\leq \frac{2\gamma}{(1-\gamma)^2}\sqrt{\frac{\log 2/\delta}{2n}}
$$
\end{lemma}
\proof{Proof of Lemma~\ref{Lemma_Crude_Bound}.}
Given a stationary policy $\pi$, if Assumption \ref{Assumption_Two_Nonzero} is true and $M\in M_\delta$, 
$$\left\|\widetilde P(\cdot|s,a) -P(\cdot|s,a)\right\|_1 \leq 2\sqrt{\frac{\log 2/\delta}{2n}},\forall s\in \gS,a\in \gA,
$$
which implies
$$
\left\|(P-\widetilde P)V_r^\pi\right\|_\infty\leq \frac{2}{1-\gamma}\sqrt{\frac{\log 2/\delta}{2n}}.
$$
Then we have
$$\begin{aligned}
    \left\|Q_r^\pi-\widetilde Q_r^\pi\right\|_\infty&= \left\|\gamma\left(I-\gamma \widetilde{P}^{\pi}\right)^{-1}(P-\widetilde{P}) V_r^{\pi}\right\|_\infty\\
    &\leq \frac{\gamma}{1-\gamma}\left\|(P-\widetilde P)V_r^\pi\right\|_\infty\\
    &\leq \frac{2\gamma}{(1-\gamma)^2}\sqrt{\frac{\log 2/\delta}{2n}}
\end{aligned}
$$
\endproof

\begin{lemma}\label{Lemma_Bound_Variance}
Given a stationary policy $\pi$, when Assumption \ref{Assumption_Two_Nonzero} is true and $M\in M_\delta$, we have
$$\Var_{P}(V_r^\pi)\leq 2\Var_{\widetilde P}(\widetilde{V}_r^\pi) + \frac{6}{(1-\gamma)^2}\sqrt{\frac{\log 2/\delta}{2n}}+\frac{8\gamma^2}{(1-\gamma)^4}\frac{\log 2/\delta}{2n}.
$$
\end{lemma}
\proof{Proof of Lemma~\ref{Lemma_Bound_Variance}.}
For simplicity of notation, we drop the dependence on $\pi$.
By definition,
$$\begin{aligned}
\Var_P(V_r) &= \Var_P(V_r)-\Var_{\widetilde P}(V_r)+\Var_{\widetilde P}(V_r)\\
&= P(V_r)^2-(PV_r)^2-\widetilde P(V_r)^2 + (\widetilde P V_r)^2+\Var_{\widetilde P}(V_r)\\
&=(P-\widetilde P)(V_r)^2-\left[(PV_r)^2-(\widetilde PV_r)^2\right]+\Var_{\widetilde P}(V_r),
\end{aligned}
$$
where $(\cdot)^2$ means element-wise squares.
When Assumption \ref{Assumption_Two_Nonzero} is true and $M\in M_\delta$, by Lemma~\ref{Lemma_Crude_Bound},
$$
\begin{aligned}
\|(P-\widetilde P)(V_r)^2\|_\infty&\leq \frac{2}{(1-\gamma)^2}\sqrt{\frac{\log 2/\delta}{2n}}\\
\left\|\left[(PV_r)^2-(\widetilde PV_r)^2\right]\right\|_\infty&\leq \|PV_r +\widetilde PV_r\|_\infty\|PV_r -\widetilde PV_r\|_\infty\\
&\leq \frac{2}{1-\gamma}\left\|PV_r -\widetilde PV_r\right\|_\infty\\
&\leq \frac{4}{(1-\gamma)^2}\sqrt{\frac{\log 2/\delta}{2n}}.
\end{aligned}
$$
We also have
$$
\begin{aligned}
\Var_{\widetilde P}(V_r)&=\Var_{\widetilde P}(V_r-\widetilde{V_r}+\widetilde V_r)\\
&\leq 2\Var_{\widetilde P}(V_r-\widetilde{V_r}) + 2\Var_{\widetilde P} (\widetilde V_r)\quad \text{(AM–GM inequality)}\\
&\leq 2\left\|V_r-\widetilde{V_r}\right\|_\infty^2+2\Var_{\widetilde P} (\widetilde V_r)\\
&\leq \frac{8\gamma^2}{(1-\gamma)^4}\frac{\log 2/\delta}{2n}+2\Var_{\widetilde P} (\widetilde V_r)\quad \text{(Lemma \ref{Lemma_Crude_Bound})}.
\end{aligned}
$$
Therefore, we can get
$$\Var_{P}(V_r^\pi)\leq 2\Var_{\widetilde P}(\widetilde{V_r^\pi}) + \frac{6}{(1-\gamma)^2}\sqrt{\frac{\log 2/\delta}{2n}}+\frac{8\gamma^2}{(1-\gamma)^4}\frac{\log 2/\delta}{2n}.
$$
\endproof

\begin{lemma}\label{Lemma_Distance_between_P_tilde_P}
Let $p,\tilde p,\hat p\in[0,1]$ satisfy
$$
\begin{aligned}
|p-\hat p|&\leq \min\brc{\sqrt{\frac{2 \hat p(1-\hat p) \log 4/\delta}{n}}+\frac{4\log 4/\delta}{n},\sqrt{\frac{ \log 2/\delta}{2 n}}}\\
|\tilde p-\hat p|&\leq \min\brc{\sqrt{\frac{2 \hat p(1-\hat p) \log 4/\delta}{n}}+\frac{4\log 4/\delta}{n},\sqrt{\frac{ \log 2/\delta}{2 n}}}.
\end{aligned}
$$
Then
$$
|p-\tilde p|\leq \sqrt{\frac{8 p(1- p) \log 4/\delta}{n}}+4\paren{\frac{ \log 4/\delta}{n}}^{3/4}+\frac{8\log 4/\delta}{n}
$$
\end{lemma}

\proof{Proof of Lemma~\ref{Lemma_Distance_between_P_tilde_P}.}
Assume WLOG that $\hat p\geq p$.
Therefore,
$$\begin{aligned}
    |p-\hat p|&\leq \sqrt{\frac{2 p(1- p) \log 4/\delta}{n}}+\sqrt{\frac{2 (\hat p- p)(1- p) \log 4/\delta}{n}}+\frac{4\log 4/\delta}{n}\\
    &\leq \sqrt{\frac{2 p(1- p) \log 4/\delta}{n}}+\sqrt{\frac{2 \sqrt{\frac{ \log 2/\delta}{2 n}} \log 4/\delta}{n}}+\frac{4\log 4/\delta}{n}\\
    &\leq \sqrt{\frac{2 p(1-p) \log 4/\delta}{n}}+2^{1/4}\paren{\frac{ \log 4/\delta}{n}}^{3/4}+\frac{4\log 4/\delta}{n}.
\end{aligned}
$$
Similarly, we have
$$|\tilde p-\hat p|\leq \sqrt{\frac{2 p(1-p) \log 4/\delta}{n}}+2^{1/4}\paren{\frac{ \log 4/\delta}{n}}^{3/4}+\frac{4\log 4/\delta}{n}.
$$
Thus we may complete the proof using triangular inequality.
\endproof

\begin{lemma}\label{Lemma_Quasi_Bernstein}
Given a stationary policy $\pi$, suppose Assumption \ref{Assumption_Two_Nonzero} is true and $M\in M_\delta$, then 
$$
|(P-\widetilde P)V_r^\pi|\preceq \sqrt{\frac{8\Var_P(V_r^\pi)\log 4/\delta}{n}}+\frac{4}{1-\gamma}\paren{\frac{\log 4/\delta}{n}}^{3/4}+\frac{8\log 4/\delta}{n(1-\gamma)},
$$
where $\preceq$ means every element of LHS is less than or equal to the its counterpart in RHS.
\end{lemma}

\proof{Proof of Lemma~\ref{Lemma_Quasi_Bernstein}.}
Let $p=P(sa^+|s,a),\tilde p=\tilde P(sa^+|s,a)$. Applying Lemma \ref{Lemma_Distance_between_P_tilde_P} yields 
$$|p-\tilde p|\leq \sqrt{\frac{8 p(1-p) \log 4/\delta}{n}}+4\paren{\frac{ \log 4/\delta}{n}}^{3/4}+\frac{8\log 4/\delta}{n}.
$$
Assume WLOG that $V_r^\pi(sa^+)\geq V_r^\pi(sa^-) $.
Therefore we have
$$
\begin{aligned}
|(P(\cdot|s,a)-\tilde{P}(\cdot|s,a))^\top V_r^\pi|\leq &\sqrt{\frac{8 p(1- p) \log 4/\delta}{n}}(V_r^\pi(sa^+)-V_r^\pi(sa^-))+\frac{4}{1-\gamma}\paren{\frac{ \log 4/\delta}{n}}^{3/4}\\
&+\frac{8\log 4/\delta}{n(1-\gamma)}.
\end{aligned}
$$
Since
$$
\begin{aligned}
p(1- p)(V_r^\pi(sa^+)- V_r^\pi(sa^-))^2&=[ p V_r^\pi(sa^+)^2+(1- p) V_r^\pi(sa^-)^2]-[ p V_r^\pi(sa^+)+(1- p) V_r^\pi(sa^-)]^2\\
&=\Var_P(V_r^\pi)
\end{aligned}
$$
We may get
$$
|(P(\cdot|s,a)-\widetilde{P}(\cdot|s,a))^\top  V_r^\pi|\leq \sqrt{\frac{8\Var_P(V_r^\pi)(s,a)\log 4/\delta}{n}}+\frac{4}{1-\gamma}\paren{\frac{\log 4/\delta}{n}}^{3/4}+\frac{8\log 4/\delta}{n(1-\gamma)},
$$
which completes the proof.
\endproof

\begin{lemma}\label{Lemma_Bound_on_V_Same_Pi}
Given a stationary policy $\pi$, suppose Assumption \ref{Assumption_Two_Nonzero} is true and $M\in M_\delta$, then we have
$$\left\|V_r^\pi -\widetilde{V_r}^\pi\right\|_\infty\leq \frac{4}{(1-\gamma)^{3/2}}\sqrt{\frac{\log 4/\delta}{n}} +\frac{4\sqrt{6}}{(1-\gamma)^2}\left(\frac{\log 4/\delta}{n}\right)^{3/4}+ \frac{8}{(1-\gamma)^4}\left(\frac{\log 4/\delta}{n}\right)^{3/2}
$$
\end{lemma}
\proof{Proof of Lemma~\ref{Lemma_Bound_on_V_Same_Pi}.}
From Lemma \ref{Lemma_Simulation_Lemma}, Lemma \ref{Lemma_Norm_of_Inf_Horizon_Expectation}, Lemma \ref{Lemma_Bound_of_Weighted_Variance}, Lemma \ref{Lemma_Quasi_Bernstein} and the fact that $\left(I-\gamma \widetilde{P}^{\pi}\right)^{-1}$ has positive entries, we know
$$
\begin{aligned}
\left\|Q^\pi-\widetilde {Q}^\pi\right\|_\infty&=\gamma\left\|\left(I-\gamma \widetilde{P}^{\pi}\right)^{-1}(P-\widetilde{P}) V_r^{\pi}\right\|_\infty\\
&\leq \sqrt{\frac{8\log 4/\delta}{n}}\left\|\left(I-\gamma\widetilde{P}^{\pi}\right)^{-1}\sqrt{\Var_{P}(V_r^\pi)}\right\|_\infty+\frac{4}{(1-\gamma)^2}\left(\frac{\log 4/\delta}{n}\right)^{3/4}+\frac{8}{(1-\gamma)^2}\left(\frac{\log 4/\delta}{n}\right)\\
&\leq \sqrt{\frac{16\log 4/\delta}{n}}\left\|\left(I-\gamma\widetilde{P}^{\pi}\right)^{-1}\sqrt{\Var_{\widetilde P}(\widetilde{V_r}^\pi)}\right\|_\infty +\frac{4\sqrt{6}}{(1-\gamma)^2}\left(\frac{\log 4/\delta}{n}\right)^{3/4}+ \frac{8}{(1-\gamma)^3}\left(\frac{\log 4/\delta}{n}\right)\\
&\leq \frac{4}{(1-\gamma)^{3/2}}\sqrt{\frac{\log 4/\delta}{n}} +\frac{4\sqrt{6}}{(1-\gamma)^2}\left(\frac{\log 4/\delta}{n}\right)^{3/4}+ \frac{8}{(1-\gamma)^3}\left(\frac{\log 4/\delta}{n}\right).
\end{aligned}
$$
The proof is completed since $\left\|V_r^\pi -\widetilde{V_r}^\pi\right\|_{\infty}\leq\left\|Q^\pi -\widetilde{Q}^\pi\right\|_{\infty}$ by definitions.
\endproof

\begin{lemma}\label{Lemma_Bound_on_V_Same_Pi_Leading}
Suppose Assumption \ref{Assumption_Two_Nonzero} is true and $n>\frac{6\log 4/\delta}{(1-\gamma)^{5/2}}$, then with probability at least $1-2|\gS|^2|\gA|\delta$, we have
$$
\begin{aligned}
\left\|V_r^\pi -\widetilde{V_r}^\pi\right\|_\infty&\leq 12\sqrt{\frac{\log 4/\delta}{n(1-\gamma)^3}}\\
\left\|C^\pi -\widetilde{C}_y^\pi\right\|_\infty&\leq 12\sqrt{\frac{\log 4/\delta}{n(1-\gamma)^3}},\forall y\in Y\\
\end{aligned}
$$
\end{lemma}

\proof{Proof of Lemma~\ref{Lemma_Bound_on_V_Same_Pi_Leading}.}
When Assumption \ref{Assumption_Two_Nonzero} is true and $M\in M_\delta$, it follows from Lemma \ref{Lemma_Bound_on_V_Same_Pi} that
$$
\left\|V_r^\pi -\widetilde{V_r}^\pi\right\|_\infty\leq \frac{4}{(1-\gamma)^{3/2}}\sqrt{\frac{\log 4/\delta}{n}} +\frac{4\sqrt{6}}{(1-\gamma)^2}\left(\frac{\log 4/\delta}{n}\right)^{3/4}+ \frac{8}{(1-\gamma)^3}\left(\frac{\log 4/\delta}{n}\right).
$$
And by setting $n>\max\left\{\frac{36\log4/\delta}{(1-\gamma)^2},\frac{4\log4/\delta}{(1-\gamma)^3}\right\}$ we will get
$$\left\|V_r^\pi -\widetilde{V_r}^\pi\right\|_\infty\leq 12\sqrt{\frac{\log 4/\delta}{n(1-\gamma)^3}}.
$$
Similar arguments can be applied to bound $\left\|C_y^\pi-\widetilde{C}_y^\pi\right\|_\infty$. Since by Theorem \ref{Theorem_Feasible} we have 
$$\PB(M\in M_\delta)\geq 1-2|\gS|^2|\gA|\delta,
$$
the proof is completed.
\endproof

\proof{Proof of Theorem \ref{Lemma_Bound_on_V}.}
By Lemma \ref{Lemma_Bound_on_V_Same_Pi}, we know that with probability $1-2|\gS|^2|\gA|\delta$,
$$\begin{aligned}
\left\|V_r^{\tilde \pi}-\tilde V_r^{\tilde \pi}\right\|_\infty&\leq12\sqrt{\frac{\log 4/\delta}{n(1-\gamma)^3}}\\
\left\|V_r^{\pi^*}-\tilde V_r^{\pi^*}\right\|_\infty&\leq 12\sqrt{\frac{\log 4/\delta}{n(1-\gamma)^3}}.
\end{aligned}
$$
Thus
$$\begin{aligned}
|V_r^{\tilde \pi}(\mu)-\tilde V_r^{\tilde \pi}(\mu)|&\leq12\sqrt{\frac{\log 4/\delta}{n(1-\gamma)^3}}\\
|V_r^{\pi^*}(\mu)-\tilde V_r^{\pi^*}(\mu)|&\leq12\sqrt{\frac{\log 4/\delta}{n(1-\gamma)^3}}.
\end{aligned}
$$ 
Noting that $\tilde V_r^{\tilde\pi}(\mu)\geq\tilde V_r^{\pi^*}(\mu)$, we may get
$$
\begin{aligned}
V_r^{\pi^*}(\mu)-V_r^{\tilde \pi}(\mu)&\leq V_r^{\pi^*}(\mu)-\tilde V_r^{\pi^*}(\mu)+\tilde V_r^{\tilde \pi}(\mu)-V_r^{\tilde \pi}(\mu)\\
&\leq|V_r^{\pi^*}(\mu)-\tilde V_r^{\pi^*}(\mu)| + |\tilde V_r^{\tilde \pi}(\mu)-V_r^{\tilde \pi}(\mu)|\\
&\leq 24\sqrt{\frac{\log 4/\delta}{n(1-\gamma)^3}}.
\end{aligned}
$$

Similarly, when
$$|C_y^{\tilde \pi}(\mu)-\tilde C_y^{\tilde \pi}(\mu)|\leq12\sqrt{\frac{\log 4/\delta}{n(1-\gamma)^3}},\forall y\in Y,
$$
we may get 
$$C_y^{\tilde \pi}(\mu) - u_y \leq 12\sqrt{\frac{\log 4/\delta}{n(1-\gamma)^3}},\forall y\in Y.
$$
since $\tilde C_y^{\tilde \pi}(\mu)\leq u_y$.
\endproof

\proof{Proof of Theorem \ref{Theorem_Sample_Complexity_General}.}
The proof is nearly identical to the proof of Theorem 3 in \cite{LATTIMORE2014125}. The idea is to augment each state/action pair of the original MDP with $|\gS|-2$ states in the form of a binary tree as pictured in the diagram below. 

\begin{figure}[!htb]
    \centering
    \includegraphics[width=0.3\textwidth]{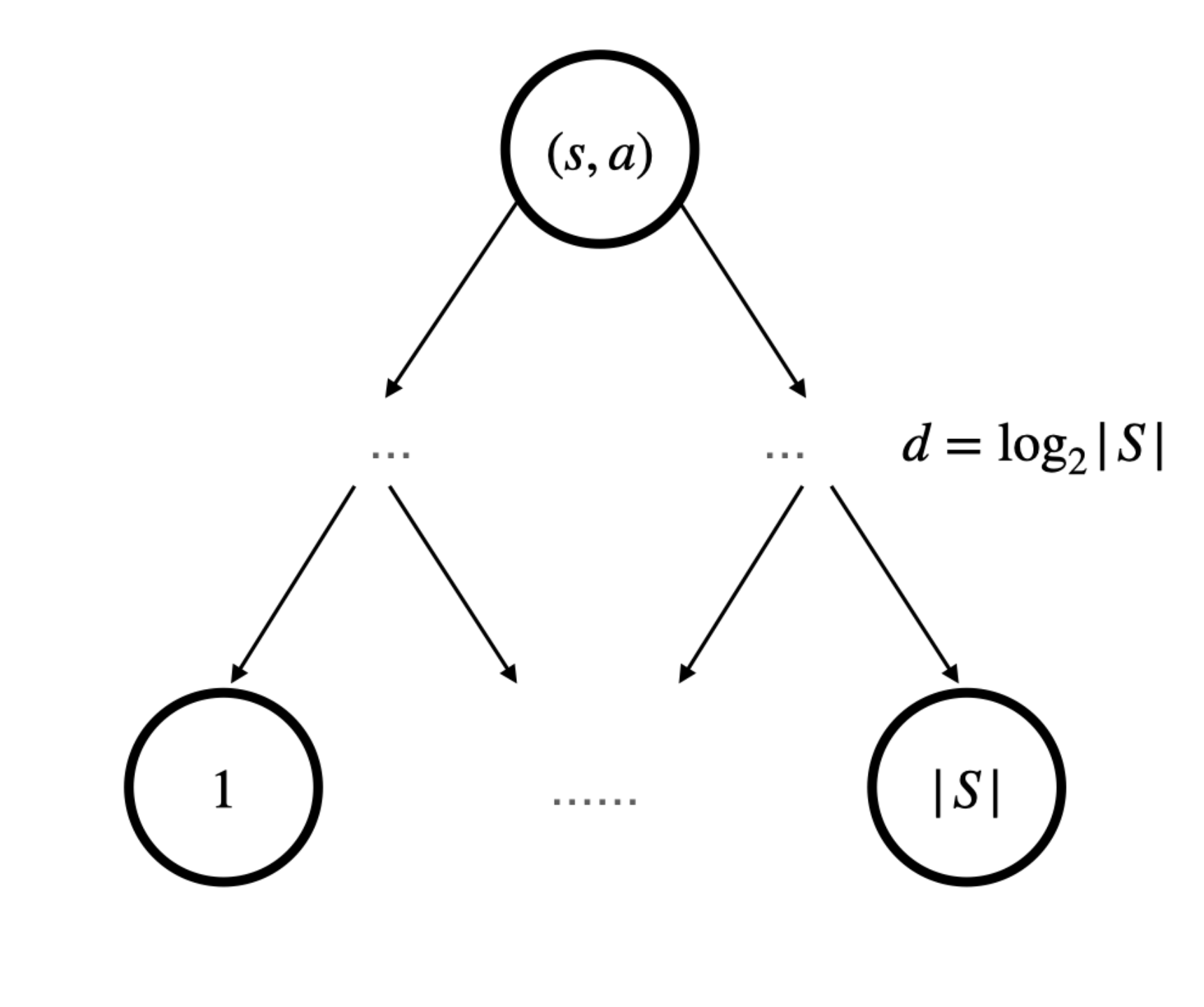}
    \label{Figure_Binary_Tree}
\end{figure}

The intention of the tree is to construct a SICMDP, $\bar M=\langle \bar \gS,\gA,Y,\bar P,\bar r,\bar c,u,\mu,\bar\gamma\rangle$ that with appropriate transition probabilities is functionally equivalent to $M$ while satisfying Assumption \ref{Assumption_Two_Nonzero}.
The rewards and costs in the added states are set to zero.
Since the tree has depth $d=O(\log_2|\gS|)$, it now takes $d$ time steps in the augmented SICMDP to change states once in the original SICMDP.
Therefore we must also rescale the discount factor $\bar \gamma$ by setting $\bar\gamma<\gamma^d$.
Now we have
$$
\begin{aligned}
|\bar \gS|&= O(|\gS|^2|\gA|)\\
\frac{1}{1-\bar \gamma}&=\frac{\log |\gS|}{1-\gamma}.
\end{aligned}
$$
Then we complete the proof by applying results in Theorem \ref{Theorem_Sample_Complexity}.
\endproof

\proof{Proof of Theorem \ref{Theorem_Sample_Complexity_General_Measure}.}
By Theorem \ref{Theorem_Chernoff_Inequality}, we have for any fixed $(s,a)\in \gS\times \gA$
$$
\begin{aligned}
\PB(n(s,a)<m\nu_{\min}/2)&\leq \PB(n(s,a)<m\nu(s,a)/2)\\
&\leq e^{-m\nu(s,a)}\left(\frac{em\nu(s,a)}{em\nu(s,a)/2}\right)^{\frac{m\nu(s,a)}{2}}\\
&=\left(\sqrt{\frac{e}{2}}\right)^{-\nu(s,a)m}\\
&\leq \left(\sqrt{\frac{e}{2}}\right)^{-\nu_{\min}m}
\end{aligned}
$$
\endproof
Let $m=\frac{2}{1-\log 2}\frac{\log 2|\gS||\gA|/\delta}{\nu_{\min}}$, we have $\PB(n(s,a)\geq m\nu_{\min}))\geq 1-\delta/2|\gS||\gA|$.
Therefore, with probability at least $1-\delta/2$, we can get
$$
n(s,a)>m\nu_{\min}, \forall (s,a)\in \gS\times \gA.
$$
Then our problem is reduced to the case that the offline dataset is generated by generative models.
The proof is completed by using results in Theorem \ref{Theorem_Sample_Complexity_General}.

\section{Omitted Proofs in Section \ref{Section_Theory_SICPO}}
\label{Appendix_Proofs_SICPO}

\begin{lemma}\label{Lemma_basic_expansion}
We have
$$
\begin{aligned}
&(1-\gamma)\alpha\sum_{t\in \gB} (V_r^*(\mu)-V_r^{(t)}(\mu))+(1-\gamma)\alpha\sum_{t\in\gN}\left(\eta-\left|\widehat V_{c^{(t)}}^{(t)}(\mu)-V_{c^{(t)}}^{(t)}(\mu)\right|\right)\\
&\leq \log|\gA|+\alpha\sum_{t\in\gB}\sqrt{E^{\nu^*}\left(r, \theta^{(t)},\hat w^{(t)}\right)}+\alpha\sum_{t\in\gN}\sqrt{E^{\nu^*}\left(c^{(t)}, \theta^{(t)},\hat w^{(t)}\right)}+\frac{\beta\alpha^2}{2}\sum_{t=0}^{T-1}\|\hat w^{(t)}\|_2^2.
\end{aligned}
$$
\end{lemma}
\proof{Proof of Lemma~\ref{Lemma_basic_expansion}}
Note that using Assumption~\ref{Assumption_smooth} and Taylor's expansion we have
$$
\log \frac{\pi_t(a | s)}{\pi_{t+1}(a | s)}+\nabla_{\theta} \log \pi_{t}(a | s)^\top\left(\theta_{t+1}-\theta_t\right) \leq \frac{\beta}{2}\left\|\theta_{t+1}-\theta_t\right\|^{2}.
$$
and $\theta^{(t+1)}-\theta^{(t)}=\alpha\hat w^{(t)}$.
As a result, suppose $t\in\gB$,
$$
\begin{aligned}
&\EB_{s\sim d^{\pi^*}}(D_{\operatorname{KL}}(\pi^*(\cdot|s)\|\pi^{(t)}(\cdot|s))-D_{\operatorname{KL}}(\pi^*(\cdot|s)\|\pi^{(t+1)}(\cdot|s)))\\
&=-\EB_{(s,a)\sim\nu^*}\log\frac{\pi^{(t)}(a|s)}{\pi^{(t+1)}(a|s)}\\
&\geq\alpha\EB_{(s,a)\sim\nu^*}[\nabla_{\theta} \log \pi^{(t)}(a | s)^\top \hat w^{(t)}]-\frac{\beta\eta_\theta^2}{2}\|\hat w^{(t)}\|^2_2\\
&=\alpha\EB_{(s,a)\sim\nu^*} A_r^{(t)}(s,a)+\alpha \EB_{(s,a)\sim\nu^*}\left[\nabla_\theta\log \pi^{(t)}(a|s)^\top \hat w^{(t)}-A_r^{(t)}(s,a)\right]-\frac{\beta\alpha^2}{2}\|\hat w^{(t)}\|^2_2\\
&\geq (1-\gamma)\alpha(V_r^*(\mu)-V_r^{(t)}(\mu))-\alpha\sqrt{E^{\nu^*}\left(r, \theta^{(t)},\hat w^{(t)}\right)}
-\frac{\beta\alpha^2}{2}\|\hat w^{(t)}\|^2_2\\
\end{aligned}
$$

The second last inequality is true due to the performance difference lemma, the last inequality is true due to Jensen's inequality and definition of the transferred function approximation error.
Rearranging terms yields
$$
\begin{aligned}
&(1-\gamma)\alpha(V_r^*(\mu)-V_r^{(t)}(\mu))\\
&\leq \EB_{s\sim d^{\pi^*}}(D_{\operatorname{KL}}(\pi^*(\cdot|s)\|\pi^{(t)}(\cdot|s))-D_{\operatorname{KL}}(\pi^*(\cdot|s)\|\pi^{(t+1)}(\cdot|s)))+\alpha\sqrt{E^{\nu^*}\left(r, \theta^{(t)},\hat w^{(t)}\right)}
+\frac{\beta\alpha^2}{2}\|\hat w^{(t)}\|^2_2.
\end{aligned}
$$
Similarly, suppose $t\in\gN$, we have $\theta^{(t+1)}-\theta^{(t)}=-\alpha\hat w^{(t)}$
$$
\begin{aligned}
&(1-\gamma)\alpha(V_{c^{(t)}}^{(t)}(\mu)-V_{c^{(t)}}^*(\mu))\\
&\leq \EB_{s\sim d^{\pi^*}}(D_{\operatorname{KL}}(\pi^*(\cdot|s)\|\pi^{(t)}(\cdot|s))-D_{\operatorname{KL}}(\pi^*(\cdot|s)\|\pi^{(t+1)}(\cdot|s)))+\alpha\sqrt{E^{\nu^*}\left(c^{(t)}, \theta^{(t)},\hat w^{(t)}\right)}
+\frac{\beta\alpha^2}{2}\|\hat w^{(t)}\|^2_2.
\end{aligned}
$$
Then we may get
$$
\begin{aligned}
&(1-\gamma)\alpha\sum_{t\in \gB} (V_r^*(\mu)-V_r^{(t)}(\mu))+(1-\gamma)\alpha\sum_{t\in \gN}\left(V_{c^{(t)}}^{(t)}(\mu)-V_{c^{(t)}}^*(\mu)\right)\\
&\leq \sum_{t=0}^{T-1} \EB_{s\sim d^{\pi^*}}(D_{\operatorname{KL}}(\pi^*(\cdot|s)\|\pi^{(t)}(\cdot|s))-D_{\operatorname{KL}}(\pi^*(\cdot|s)\|\pi^{(t+1)}(\cdot|s)))\\
&\quad+\alpha\sum_{t\in\gB}\sqrt{E^{\nu^*}\left(r, \theta^{(t)},\hat w^{(t)}\right)}+\alpha\sum_{t\in\gN}\sqrt{E^{\nu^*}\left(c^{(t)}, \theta^{(t)},\hat w^{(t)}\right)}+\frac{\beta\alpha^2}{2}\sum_{t=0}^{T-1}\|\hat w^{(t)}\|_2^2\\
&\leq \log|\gA|+\alpha\sum_{t\in\gB}\sqrt{E^{\nu^*}\left(r, \theta^{(t)},\hat w^{(t)}\right)}+\alpha\sum_{t\in\gN}\sqrt{E^{\nu^*}\left(c^{(t)}, \theta^{(t)},\hat w^{(t)}\right)}+\frac{\beta\alpha^2}{2}\sum_{t=0}^{T-1}\|\hat w^{(t)}\|_2^2.
\end{aligned}
$$
Since for $t\in\gN$
$$
\begin{aligned}
V_{c^{(t)}}^{(t)}(\mu)-V_{c^{(t)}}^*(\mu)&\geq \widehat V_{c^{(t)}}^{(t)}(\mu)-V_{c^{(t)}}^*(\mu)-\left|\widehat V_{c^{(t)}}^{(t)}(\mu)-V_{c^{(t)}}^{(t)}(\mu)\right|\\
&\geq u_{y_*^{(t)}}+\eta-u_{y_*^{(t)}}-\left|\widehat V_{c^{(t)}}^{(t)}(s)-V_{c^{(t)}}^{(t)}(\mu)\right|\\
&=\eta-\left|\widehat V_{c^{(t)}}^{(t)}(s)-V_{c^{(t)}}^{(t)}(\mu)\right|
\end{aligned}
$$
we may obtain the conclusion.
\endproof

\begin{lemma}\label{Lemma_SGD_convergence}
We have for $t\in\gB$, $\forall\delta\in(0,1)$, with probability at least $1-\delta$,
for $t\in\gB$
$$
\begin{aligned}
    & E^{\nu^*}(r,\theta^{(t)}, \hat w^{(t)})\\
    &\leq \frac{1}{1-\gamma}\left\|\frac{\nu^*}{\nu_0}\right\|_\infty\left(\epsilon_{bias}+C\frac{(4L_\pi^2W+8L_\pi/(1-\gamma))^2}{(1-\gamma)^2\mu_F}\frac{\log(T/\delta)}{K_{sgd}}+\frac{4\gamma^H}{1-\gamma}\left(\frac{1}{1-\gamma}+WL_\pi\right)\right),
\end{aligned}
$$
for $t\in\gN$,
$$
\begin{aligned}
     &E^{\nu^*}(c^{(t)},\theta^{(t)}, \hat w^{(t)})\\
     &\leq \frac{1}{1-\gamma}\left\|\frac{\nu^*}{\nu_0}\right\|_\infty\left(\epsilon_{bias}+C\frac{(4L_\pi^2W+8L_\pi/(1-\gamma))^2}{(1-\gamma)^2\mu_F}\frac{\log(T/\delta)}{K_{sgd}}+\frac{4\gamma^H}{1-\gamma}\left(\frac{1}{1-\gamma}+WL_\pi\right)\right),
\end{aligned}
$$
where $\nu_0$ is the uniform distribution on $\gS\times\gA$.
\end{lemma}

\proof{Proof of Lemma~\ref{Lemma_SGD_convergence}}
First we show for $t\in\gB$, with high probability
$$
\begin{aligned}
    &E^{\nu^*}(r,\theta^{(t)}, \hat w^{(t)})\\
    &\leq \frac{1}{1-\gamma}\left\|\frac{\nu^*}{\nu_0}\right\|_\infty\left(\epsilon_{bias}+C\frac{(4L_\pi^2W+8L_\pi/(1-\gamma))^2}{(1-\gamma)^2\mu_F}\frac{\log(T/\delta)}{K_{sgd}}+\frac{4\gamma^H}{1-\gamma}\left(\frac{1}{1-\gamma}+WL_\pi\right)\right).
\end{aligned}
$$
We have:
$$
\begin{aligned}
&E^{\nu^*}(r,\theta^{(t)}, \hat w^{(t)})\\
&\leq \left\|\frac{\nu^{*}}{\nu^{(t)}}\right\|_{\infty} E^{\nu^{(t)}}(r,\theta^{(t)}, \hat w^{(t)})\\
&\leq \frac{1}{1-\gamma} \left\|\frac{\nu^{*}}{\nu_{0}}\right\|_{\infty} E^{\nu^{(t)}}(r,\theta^{(t)}, \hat w^{(t)})\\
&=\frac{1}{1-\gamma}\left\|\frac{\nu^{*}}{\nu_{0}}\right\|_{\infty}\left( \min_w E^{\nu^{(t)}}(r,\theta^{(t)},w)+E^{\nu^{(t)}}\left(r,\theta^{(t)}, \hat w^{(t)}\right)-\min_w E^{\nu^{(t)}}(r,\theta^{(t)},w)\right)\\
&\leq \frac{1}{1-\gamma}\left\|\frac{\nu^{*}}{\nu_{0}}\right\|_{\infty}\left(\epsilon_{bias}+E^{\nu^{(t)}}\left(r, \theta^{(t)}, \hat w^{(t)}\right)-\min_{w\in B(0,W,\|\cdot\|_2)} E^{\nu^{(t)}}(r,\theta^{(t)},w)\right),
\end{aligned}
$$
the last step is due to Assumption~\ref{Assumption_func_approx_err} and Assumption~\ref{Assumption_est_err}.
Now we define a proximal loss function:
$$
\widetilde{E}^{\nu^{(t)}}(r,\theta^{(t)},w):=\EB_{(s,a)\sim\nu}(\widetilde{A}^{\pi^{(t)}}_r(s,a)-w^\top\nabla_\theta\log\pi_\theta(a|s))^2,
$$
where
$$
\begin{aligned}
    \widetilde{A}^{(t)}_r(s,a):&=\widetilde{Q}^{\pi^{(t)}}_r(s,a)-\widetilde{V}^{\pi^{(t)}}_r(s),\\
    \widetilde{Q}^{\pi^{(t)}}_r(s,a):&=\EB\left(\sum_{t=0}^H\gamma^t r(s_t,a_t)\mid (s_0,a_0)=(s,a)\right),\\
    \widetilde{V}^{\pi^{(t)}}_r(s):&=\EB\left(\sum_{t=0}^H\gamma^t r(s_t,a_t)\mid s_0=s\right).\\
\end{aligned}
$$
Recall that
$$
\begin{aligned}
&\widetilde{E}^{\nu^{(t)}}(r,\theta^{(t)},w)\\
&=\EB_{(s,a)\sim\nu^{(t)}}(\widetilde{A}^{(t)}(s,a)-w^\top\nabla_\theta\log\pi_{\theta^{(t)}}(a|s))^2\\    &=w^\top F(\theta^{(t)})w-2\sum_{s,a}\nu^{(t)}(s,a)w^\top \nabla_\theta\log\pi_{\theta^{(t)}}(a|s)\widetilde{A}^{(t)}(s,a)+\sum_{s,a}\nu^{(t)}(s,a)\widetilde{A}^{(t)}(s,a)^2.
\end{aligned}
$$
According to Assumption~\ref{Assumption_PSD_Fisher}, $E^{\nu^{(t)}}(r,\theta^{(t)},w)$ is $(1-\gamma)^2\mu_F$-strongly convex.
The full gradient is
$$G(w)=2F(\theta^{(t)})w-2\sum_{s,a}\nu^{(t)}(s,a)\widetilde{A}^{(t)}(s,a)\nabla_\theta\log\pi_{\theta^{(t)}}(a|s),
$$
and the stochastic gradient we use is 
$$\widehat G(w)=2\nabla_\theta\log\pi_\theta(A|S)\nabla_\theta\log\pi_\theta(A|S)^\top w-2\widehat A^{\pi_\theta}(S,A)\nabla_\theta\log\pi_\theta(A|S),
$$
where $(S,A)\sim \nu^{(t)}$.
Then we have $\EB\widehat G(w)=G(w)$ and
$$
\|G(w)-\widehat G(w)\|_2\leq 2L_\pi^2W+\frac{4L_\pi}{1-\gamma}.
$$
Set $\eta_k=\frac{2}{(1-\gamma)^2\mu_F(k+1)}$, $\gamma_k=\frac{2k}{K(K+1)}$, by Theorem C.3 in \cite{Harvey2019Simple}, for any $\delta\in(0,1)$, with probability at least $1-\delta$ we have,
$$
\widetilde{E}^{\nu^{(t)}}\left(r, \theta^{(t)}, \hat w^{(t)}\right)-\min_{w\in B(0,W,\|\cdot\|_2)} \widetilde{E}^{\nu^{(t)}}(r,\theta^{(t)},w)\leq C\frac{(4L_\pi^2W+8L_\pi/(1-\gamma))^2}{(1-\gamma)^2\mu_F}\frac{\log(1/\delta)}{K_{sgd}}.
$$
Here $C$ is a universal constant.
We may finish the bound by noting that
$$
\sup_{w\in B(0,W,\|\cdot\|_2)}\left|\widetilde{E}^{\nu^{(t)}}\left(r, \theta^{(t)}, \hat w^{(t)}\right)-E^{\nu^{(t)}}\left(r, \theta^{(t)}, \hat w^{(t)}\right)\right|\leq \frac{2\gamma^H}{1-\gamma}\left(\frac{1}{1-\gamma}+WL_\pi\right).
$$
For $t\in\gN$, the inequality holds by similar arguments as above.
And we would like to emphasize that $c^{(t)}$ is determined a priori and thus can be viewed as a fixed cost.
Our final conclusion can be obtained by a union bound argument.
\endproof

\begin{lemma}\label{Lemma_whp_basic_expansion}
For any $\delta\in(0,1)$, let $A$ denote the event
$$
\begin{aligned}
&(1-\gamma)\alpha\sum_{t\in \gB} (V_r^*(\mu)-V_r^{(t)}(\mu))\\
&+(1-\gamma)\alpha|\gN|\left(\eta-\frac{1}{(1-\gamma)\sqrt{2K_{eval}}}\left(1+\sqrt{\log{(4T/\delta)}+m\log(4\mathrm{diam}(Y)L_y\sqrt{2K_{eval}}+\mathrm{diam}(Y))}\right)-\frac{\gamma^H}{1-\gamma}\right)\\
&\leq \log|\gA|+\frac{\beta\alpha^2 TW^2}{2(1-\gamma)^2}\\
&+\alpha T\sqrt{\frac{1}{1-\gamma}\left\|\frac{\nu^*}{\nu_0}\right\|_\infty\left(\epsilon_{bias}+C\frac{(4L_\pi^2W+8L_\pi/(1-\gamma))^2}{(1-\gamma)^2\mu_F}\frac{\log(T/\delta)}{K_{sgd}}+\frac{4\gamma^H}{1-\gamma}\left(\frac{1}{1-\gamma}+WL_\pi\right)\right)}.
\end{aligned}
$$
Then we have $\PB(A)\geq 1-\delta$.
\end{lemma}
\proof{Proof of Lemma~\ref{Lemma_whp_basic_expansion}}
Let $A_1$ denote the event that 
for $t\in\gB$
$$
\begin{aligned}
     &E^{\nu^*}(r,\theta^{(t)}, \hat w^{(t)})\\
     &\leq \frac{1}{1-\gamma}\left\|\frac{\nu^*}{\nu_0}\right\|_\infty\left(\epsilon_{bias}+C\frac{(4L_\pi^2W+8L_\pi/(1-\gamma))^2}{(1-\gamma)^2\mu_F}\frac{\log(2T/\delta)}{K_{sgd}}+\frac{4\gamma^H}{1-\gamma}\left(\frac{1}{1-\gamma}+WL_\pi\right)\right),
\end{aligned}
$$
for $t\in\gN$,
$$
\begin{aligned}
     &E^{\nu^*}(c^{(t)},\theta^{(t)}, \hat w^{(t)})\\
     &\leq \frac{1}{1-\gamma}\left\|\frac{\nu^*}{\nu_0}\right\|_\infty\left(\epsilon_{bias}+C\frac{(4L_\pi^2W+8L_\pi/(1-\gamma))^2}{(1-\gamma)^2\mu_F}\frac{\log(2T/\delta)}{K_{sgd}}+\frac{4\gamma^H}{1-\gamma}\left(\frac{1}{1-\gamma}+WL_\pi\right)\right).
\end{aligned}
$$
By Lemma~\ref{Lemma_SGD_convergence}, $\PB(A_1)\geq 1-\delta/2$.
Also note that we always have $\|\hat w^{(t)}\|_2^2\leq \frac{W^2}{(1-\gamma)^2}$, together with Lemma~\ref{Lemma_basic_expansion} lead to that conditioned on $A_1$,
$$
\begin{aligned}
&(1-\gamma)\alpha\sum_{t\in \gB} (V_r^*(\mu)-V_r^{(t)}(\mu))+(1-\gamma)\alpha\sum_{t\in\gN}\left(\eta-\left|\widehat V_{c^{(t)}}^{(t)}(\mu)-V_{c^{(t)}}^{(t)}(\mu)\right|\right)\\
&\leq \log|\gA|+\frac{\beta\alpha^2 TW^2}{2(1-\gamma)^2}\\
&\alpha T\sqrt{\frac{1}{1-\gamma}\left\|\frac{\nu^*}{\nu_0}\right\|_\infty\left(\epsilon_{bias}+C\frac{(4L_\pi^2W+8L_\pi/(1-\gamma))^2}{(1-\gamma)^2\mu_F}\frac{\log(2T/\delta)}{K_{sgd}}+\frac{4\gamma^H}{1-\gamma}\left(\frac{1}{1-\gamma}+WL_\pi\right)\right)}.
\end{aligned}
$$
It remains to bound $\left|\widehat V_{c^{(t)}}^{(t)}(\mu)-V_{c^{(t)}}^{(t)}(\mu)\right|$.
Let $A_2$ denote the event that $\forall t\in\{0,...,T-1\}$,
 $$
    \left|\widehat V_{c^{(t)}}^{(t)}(\mu)-V_{c^{(t)}}^{(t)}(\mu)\right|\leq \frac{1}{(1-\gamma)\sqrt{2K_{eval}}}\left(1+\sqrt{\log{(4T/\delta)}+m\log(4\mathrm{diam}(Y)L_y\sqrt{2K_{eval}}+\mathrm{diam}(Y))}+\frac{\gamma^H}{1-\gamma}\right)
    $$
By Lemma~\ref{Lemma_WHP_Bound_Evaluation_Uniform_y}, $\PB(A_2)\geq \delta/2$.
Thus conditioned on the event $A_1\cap A_2$,
$$
\begin{aligned}
&(1-\gamma)\alpha\sum_{t\in \gB} (V_r^*(\mu)-V_r^{(t)}(\mu))\\
&+(1-\gamma)\alpha|\gN|\left(\eta-\frac{1}{(1-\gamma)\sqrt{2K_{eval}}}\left(1+\sqrt{\log{(4T/\delta)}+m\log(4\mathrm{diam}(Y)L_y\sqrt{2K_{eval}}+\mathrm{diam}(Y))}\right)-\frac{\gamma^H}{1-\gamma}\right)\\
&\leq \log|\gA|+\frac{\beta\alpha^2 TW^2}{2(1-\gamma)^2}\\
&+\alpha T\sqrt{\frac{1}{1-\gamma}\left\|\frac{\nu^*}{\nu_0}\right\|_\infty\left(\epsilon_{bias}+C\frac{(4L_\pi^2W+8L_\pi/(1-\gamma))^2}{(1-\gamma)^2\mu_F}\frac{\log(2T/\delta)}{K_{sgd}}+\frac{4\gamma^H}{1-\gamma}\left(\frac{1}{1-\gamma}+WL_\pi\right)\right)}.
\end{aligned}
$$
We complete the proof by noting $\PB(A_1\cap A_2)\geq 1-\delta$.
\endproof

\begin{lemma}\label{Lemma_well_define_B}
    Suppose 
    $$\begin{aligned}
        &\frac{T}{2}\left(\eta-\frac{1}{(1-\gamma)\sqrt{2K_{eval}}}\left(1+\sqrt{\log{(4T/\delta)}+m\log(4\mathrm{diam}(Y)L_y\sqrt{2K_{eval}}+\mathrm{diam}(Y))}\right)-\frac{\gamma^H}{1-\gamma}\right)\\
        &>\frac{\log|\gA|}{(1-\gamma)\alpha}+\frac{\beta\alpha TW^2}{2(1-\gamma)^3}\\
        &+\frac{ T}{1-\gamma}\sqrt{\frac{1}{1-\gamma}\left\|\frac{\nu^*}{\nu_0}\right\|_\infty\left(\epsilon_{bias}+C\frac{(4L_\pi^2W+8L_\pi/(1-\gamma))^2}{(1-\gamma)^2\mu_F}\frac{\log(2T/\delta)}{K_{sgd}}+\frac{4\gamma^H}{1-\gamma}\left(\frac{1}{1-\gamma}+WL_\pi\right)\right)},
    \end{aligned}
    $$
Then $\gB\not=\emptyset$. Then for any $\delta\in(0,1)$, conditioned on event $A$, either a). $|\gB|\geq T/2$ or b). $\sum_{t\in\gB}(V_r^*(\mu)-V_r^{(t)}(\mu))< 0$.
\end{lemma}
\proof{Proof of Lemma~\ref{Lemma_well_define_B}}
    First we set $\sum_{t\in\gB}(V_r^*(\mu)-V_r^{(t)}(\mu))= 0$ if $\gB=\emptyset$. Now we would show $|\gB|\geq T/2$ as long as $\sum_{t\in\gB}(V_r^*(\mu)-V_r^{(t)}(\mu))\geq 0$, which also precludes the possibility that $\gB=\emptyset$.
    Note that when $\sum_{t\in\gB}(V_r^*(\mu)-V_r^{(t)}(\mu))\geq 0$ and $A$ happens,
    $$
    \begin{aligned}
    &|\gN|\left(\eta-\frac{1}{(1-\gamma)\sqrt{2K_{eval}}}\left(1+\sqrt{\log{(4T/\delta)}+m\log(4\mathrm{diam}(Y)L_y\sqrt{2K_{eval}}+\mathrm{diam}(Y))}\right)-\frac{\gamma^H}{1-\gamma}\right)\\
    &\leq\frac{\log|\gA|}{(1-\gamma)\alpha}+\frac{\beta\alpha TW^2}{2(1-\gamma)^3}\\
    &+\frac{ T}{1-\gamma}\sqrt{\frac{1}{1-\gamma}\left\|\frac{\nu^*}{\nu_0}\right\|_\infty\left(\epsilon_{bias}+C\frac{(4L_\pi^2W+8L_\pi/(1-\gamma))^2}{(1-\gamma)^2\mu_F}\frac{\log(2T/\delta)}{K_{sgd}}+\frac{4\gamma^H}{1-\gamma}\left(\frac{1}{1-\gamma}+WL_\pi\right)\right)}.
    \end{aligned}
    $$
    Since
    $$
    \begin{aligned}
    &\frac{T}{2}\left(\eta-\frac{1}{(1-\gamma)\sqrt{2K_{eval}}}\left(1+\sqrt{\log{(4T/\delta)}+m\log(4\mathrm{diam}(Y)L_y\sqrt{2K_{eval}}+\mathrm{diam}(Y))}\right)-\frac{\gamma^H}{1-\gamma}\right)\\
    &>\frac{\log|\gA|}{(1-\gamma)\alpha}+\frac{\beta\alpha TW^2}{2(1-\gamma)^3}\\
    &+\frac{ T}{1-\gamma}\sqrt{\frac{1}{1-\gamma}\left\|\frac{\nu^*}{\nu_0}\right\|_\infty\left(\epsilon_{bias}+C\frac{(4L_\pi^2W+8L_\pi/(1-\gamma))^2}{(1-\gamma)^2\mu_F}\frac{\log(2T/\delta)}{K_{sgd}}+\frac{4\gamma^H}{1-\gamma}\left(\frac{1}{1-\gamma}+WL_\pi\right)\right)}.
    \end{aligned}
    $$ we can conclude $|\gN|\leq T/2$ and $|\gB|\geq T/2$.
\endproof

\begin{lemma}\label{Lemma_reward_bound}
    Let $\alpha=1/\sqrt{T}$. Suppose we choose
    $$K_{eval}=\widetilde{O}\left(\frac{1}{\eta^2(1-\gamma)^2}\right)
    ,~H=O\left(\log\left(\frac{\eta(1-\gamma)}{\gamma}\right)\right)$$ and
    $$\begin{aligned}
    \eta&> \frac{4\log |\gA|}{\sqrt{T}(1-\gamma)}+\frac{2\beta W^2}{\sqrt{T}(1-\gamma)^3}\\
    &+\frac{4}{1-\gamma}\sqrt{\frac{1}{1-\gamma}\left\|\frac{\nu^*}{\nu_0}\right\|_\infty\left(\epsilon_{bias}+C\frac{(4L_\pi^2W+8L_\pi/(1-\gamma))^2}{(1-\gamma)^2\mu_F}\frac{\log(2T/\delta)}{K_{sgd}}+\frac{4\gamma^H}{1-\gamma}\left(\frac{1}{1-\gamma}+WL_\pi\right)\right)}
    \end{aligned}
    $$,
    then we have for any $\delta\in(0,1)$, conditioned on event $A$,
    $$
    \begin{aligned}
        &\frac{1}{|\gB|}\sum_{t\in\gB}(V_r^*(\mu)-V_r^{(t)}(\mu)) \leq \frac{2\log|\gA|}{\sqrt{T}(1-\gamma)}+\frac{\beta W^2}{\sqrt{T}(1-\gamma)^3}\\
       &+\frac{2}{1-\gamma}\sqrt{\frac{1}{1-\gamma}\left\|\frac{\nu^*}{\nu_0}\right\|_\infty\left(\epsilon_{bias}+C\frac{(4L_\pi^2W+8L_\pi/(1-\gamma))^2}{(1-\gamma)^2\mu_F}\frac{\log(2T/\delta)}{K_{sgd}}+\frac{4\gamma^H}{1-\gamma}\left(\frac{1}{1-\gamma}+WL_\pi\right)\right)}.
    \end{aligned}
    $$
    Here $\widetilde{O}$ means we discard any logarithmic terms. 
\end{lemma}
\proof{Proof of Lemma~\ref{Lemma_reward_bound}}
    When we choose $K_{eval}=\widetilde{O}\left(\frac{m}{\eta^2(1-\gamma)^2}\right)$ and $H=O\left(\log\left(\frac{\eta(1-\gamma)}{\gamma}\right)\right)$, we have
    $$
    \eta-\frac{1}{(1-\gamma)\sqrt{2K_{eval}}}\left(1+\sqrt{\log{(4T/\delta)}+m\log(4\mathrm{diam}(Y)L_y\sqrt{2K_{eval}}+\mathrm{diam}(Y))}\right)-\frac{\gamma^H}{1-\gamma}
    >\frac{\eta}{2}$$
    Setting
    $$\begin{aligned}
    \eta&> \frac{4\log |\gA|}{\sqrt{T}(1-\gamma)}+\frac{2\beta W^2}{\sqrt{T}(1-\gamma)^3}\\
    &+\frac{4}{1-\gamma}\sqrt{\frac{1}{1-\gamma}\left\|\frac{\nu^*}{\nu_0}\right\|_\infty\left(\epsilon_{bias}+C\frac{(4L_\pi^2W+8L_\pi/(1-\gamma))^2}{(1-\gamma)^2\mu_F}\frac{\log(2T/\delta)}{K_{sgd}}+\frac{4\gamma^H}{1-\gamma}\left(\frac{1}{1-\gamma}+WL_\pi\right)\right)}
    \end{aligned}
    $$
    results in
   $$\begin{aligned}
        &\frac{T}{2}\left(\eta-\frac{1}{(1-\gamma)\sqrt{2K_{eval}}}\left(1+\sqrt{\log{(4T/\delta)}+m\log(4\mathrm{diam}(Y)L_y\sqrt{2K_{eval}}+\mathrm{diam}(Y))}\right)-\frac{\gamma^H}{1-\gamma}\right)\\
        &>\frac{\log|\gA|}{(1-\gamma)\alpha}+\frac{\beta\alpha TW^2}{2(1-\gamma)^3}\\
        &\frac{ T}{1-\gamma}\sqrt{\frac{1}{1-\gamma}\left\|\frac{\nu^*}{\nu_0}\right\|_\infty\left(\epsilon_{bias}+C\frac{(4L_\pi^2W+8L_\pi/(1-\gamma))^2}{(1-\gamma)^2\mu_F}\frac{\log(2T/\delta)}{K_{sgd}}+\frac{4\gamma^H}{1-\gamma}\left(\frac{1}{1-\gamma}+WL_\pi\right)\right)},
    \end{aligned}
    $$
    Then we may use Lemma~\ref{Lemma_well_define_B} to conclude either a). $|\gB|\geq T/2$ or b). $\sum_{t\in\gB}(V_r^*(\mu)-V_r^{(t)}(\mu))< 0$.
    If $a)$, the conclusion holds trivially.
    Else via the combination of b) and the fact that we are in event $A$ we may obtain
$$
    \begin{aligned}
        &\frac{1}{|\gB|}\sum_{t\in\gB}(V_r^*(\mu)-V_r^{(t)}(\mu)) \leq \frac{2\log|\gA|}{\sqrt{T}(1-\gamma)}+\frac{\beta W^2}{\sqrt{T}(1-\gamma)^3}\\
       &+\frac{2}{1-\gamma}\sqrt{\frac{1}{1-\gamma}\left\|\frac{\nu^*}{\nu_0}\right\|_\infty\left(\epsilon_{bias}+C\frac{(4L_\pi^2W+8L_\pi/(1-\gamma))^2}{(1-\gamma)^2\mu_F}\frac{\log(2T/\delta)}{K_{sgd}}+\frac{4\gamma^H}{1-\gamma}\left(\frac{1}{1-\gamma}+WL_\pi\right)\right)}.
    \end{aligned}
    $$
    
\endproof

\begin{lemma}\label{Lemma_constraint_violation_random_search}
    Suppose Assumption~\ref{Assumption_regular_maxima} holds and we use random search (Algorithm~\ref{Algorithm_random_search}) to solve the inner-loop problem.
    Let $B$ denote the event that for any $t\in\{1,...,T\}$, 
    $$\begin{aligned}
    &\max_{y\in Y} \left[V_{c_y}^{(t)}(\mu)-u_y\right]-\left[\widehat V_{c_{y^{(t)}}}^{(t)}(\mu)-u^{(t)}\right]\\
    &\leq \frac{1}{(1-\gamma)\sqrt{2K_{eval}}}\left(1+\sqrt{\log{(4T/\delta)}+m\log(4\mathrm{diam}(Y)L_y\sqrt{2K_{eval}}+\mathrm{diam}(Y))}\right)+\frac{\gamma^H}{1-\gamma}+2\epsilon_{con},
    \end{aligned}$$
    We have $\PB(B\cap A_2)\geq 1-\delta$ as long as $M>\log(2T/\delta)/\log(1-\epsilon_{con}^m/[\operatorname{vol}(Y)(2L_y)^m(1-\gamma)^m])    
    $
\end{lemma}
\proof{Proof of Lemma~\ref{Lemma_constraint_violation_random_search}}
    According to Lemma~\ref{Lemma_WHP_Bound_Evaluation_Uniform_y}, conditioned on event $A_2$, we have $\forall t\in\{1,...,T\}$
     $$
    \sup_{y\in Y}\left|\widehat V_{c_y}^{(t)}(\mu)-V_{c_y}^{(t)}(\mu)\right|\leq \frac{1}{(1-\gamma)\sqrt{2K_{eval}}}\left(1+\sqrt{\log{(4T/\delta)}+m\log(4\mathrm{diam}(Y)L_y\sqrt{2K_{eval}}+\mathrm{diam}(Y))}\right)+\frac{\gamma^H}{1-\gamma}
    $$
    as long as $K_{eval}$ is large enough.
    Therefore, let $y^*\in\arg\max_{y\in Y} \left[V_{c_y}^\pi(\mu)-u_y\right]$, 
    $$
    \left|V_{c_{y^*}}^\pi(\mu)-\widehat{V}_{c_{y^*}}^\pi(\mu)\right|\leq \frac{1}{(1-\gamma)\sqrt{2K_{eval}}}\left(1+\sqrt{\log{(4T/\delta)}+m\log(4\mathrm{diam}(Y)L_y\sqrt{2K_{eval}}+\mathrm{diam}(Y))}\right)+\frac{\gamma^H}{1-\gamma}
    $$
    By Assmption~\ref{Assumption_Lipschitz}, as long as $\exists i\in\{1,...,M\}$, $\|y_i-y^*\|_\infty\leq \epsilon_{con}(1-\gamma)/2L_y$,
    we have 
    $$\left[\widehat V_{c_{y^*}}^{(t)}(\mu)-u_{ y^*}\right]-\max_{i\in\{1,...,M\}}\left[\widehat V_{c_{y_i}}^\pi(\mu)-u_{y_i}\right]\leq \epsilon_{con}.$$
     According to Assumption~\ref{Assumption_regular_maxima}, when $\epsilon_{con}(1-\gamma)/2L_y<\epsilon_0$
    $$
    \PB(\exists i\in\{1,...,M\}, \|y_i-\hat y\|_\infty\leq \epsilon_{con}(1-\gamma)/2L_y)\geq 1-\left(1-\frac{\epsilon_{con}^m(1-\gamma)^m}{\operatorname{vol}(Y)(2L_y)^m}\right)^M.
    $$
    If $M\geq M(\epsilon_{con},\beta)$, we have
    $$\PB\left(\left[\widehat V_{c_{y^*}}^{(t)}(\mu)-u_{ y^*}\right]-\max_{i\in\{1,...,M\}}\left[\widehat V_{c_{y_i}}^\pi(\mu)-u_{y_i}\right]\leq \epsilon\right)\geq 1-\beta,$$ 
    where we define
    $$M(\epsilon_{con},\beta)=\frac{\log\beta}{\log\left(1-\frac{\epsilon_{con}^m(1-\gamma)^m}{\operatorname{vol}(Y)(2L_y)^m}\right)}.
    $$
    Setting $M=M(\epsilon_{con},\delta/2T)$ and applying the union bound argument yields the result.
\endproof

\begin{lemma}\label{Lemma_constraint_violation_PGA}
    Suppose Assumption~\ref{Assumption_concave_constraint} holds and we use projected subgradient ascent (Algorithm~\ref{Algorithm_projected_GD}) to solve the inner-loop problem.
    Then 
    $$\begin{aligned}
    &\max_{y\in Y} \left[V_{c_y}^{(t)}(\mu)-u_y\right]-\left[\widehat V_{c_{y^{(t)}}}^{(t)}(\mu)-u^{(t)}\right]\\
    &\leq \frac{1}{(1-\gamma)\sqrt{2K_{eval}}}\left(1+\sqrt{\log{(4T/\delta)}+m\log(4\mathrm{diam}(Y)L_y\sqrt{2K_{eval}}+\mathrm{diam}(Y))}\right)+\frac{\gamma^H}{1-\gamma}+\epsilon_{con},
    \end{aligned}$$
    as long as $T_{PGA}>\frac{4[\mathrm{diam}(Y)]^2L_y^2}{\epsilon_{con}^2(1-\gamma)^2}
    $.
\end{lemma}
\proof{Proof of Lemma~\ref{Lemma_constraint_violation_PGA}}
Noting that the objective is $\frac{2L_y}{1-\gamma}$-Lipschitz, we may complete the proof by directly applying Theorem 3.2 in \cite{bubeck2015convex}.
\endproof
\proof{Proof of Theorem~\ref{Theorem_random_search_SICPO}}
Conditioned on event $A=A_1\cap A_2$, setting $K_{sgd}, H, K_{eval}, T, \alpha, \eta$ as above yields
$$\begin{aligned}
    \eta&> \frac{4\log |\gA|}{\sqrt{T}(1-\gamma)}+\frac{2\beta W^2}{\sqrt{T}(1-\gamma)^3}\\
    &+\frac{4}{1-\gamma}\sqrt{\frac{1}{1-\gamma}\left\|\frac{\nu^*}{\nu_0}\right\|_\infty\left(\epsilon_{bias}+C\frac{(4L_\pi^2W+8L_\pi/(1-\gamma))^2}{(1-\gamma)^2\mu_F}\frac{\log(2T/\delta)}{K_{sgd}}+\frac{4\gamma^H}{1-\gamma}\left(\frac{1}{1-\gamma}+WL_\pi\right)\right)}.
    \end{aligned}
$$
Now we use Lemma~\ref{Lemma_reward_bound} to get
$$
    \begin{aligned}
        &\frac{1}{|\gB|}\sum_{t\in\gB}(V_r^*(\mu)-V_r^{(t)}(\mu)) \leq \frac{2\log|\gA|}{\sqrt{T}(1-\gamma)}+\frac{\beta W^2}{\sqrt{T}(1-\gamma)^3}\\
       &+\frac{2}{1-\gamma}\sqrt{\frac{1}{1-\gamma}\left\|\frac{\nu^*}{\nu_0}\right\|_\infty\left(\epsilon_{bias}+C\frac{(4L_\pi^2W+8L_\pi/(1-\gamma))^2}{(1-\gamma)^2\mu_F}\frac{\log(2T/\delta)}{K_{sgd}}+\frac{4\gamma^H}{1-\gamma}\left(\frac{1}{1-\gamma}+WL_\pi\right)\right)}\\
       &\leq \epsilon+\frac{1}{(1-\gamma)^{3/2}}\sqrt{\left\|\frac{\nu^*}{\nu_0}\right\|_\infty\epsilon_{bias}}.
    \end{aligned}
    $$
According to Lemma~\ref{Lemma_constraint_violation_random_search}, conditioned on event $A_2\cap B$, as long as we set $M$ as in the statement of the theorem, we have $\forall t\in\{1,...,T\}$
$$\begin{aligned}
    &\sup_{y\in Y} \left[V_{c_y}^{(t)}(\mu)-u_y\right]-\left[\widehat V_{c_{y^{(t)}}}^{(t)}(\mu)-u^{(t)}\right]\\
    &\leq \frac{1}{(1-\gamma)\sqrt{2K_{eval}}}\left(1+\sqrt{\log{(4T/\delta)}+m\log(4\mathrm{diam}(Y)L_y\sqrt{2K_{eval}}+\mathrm{diam}(Y))}\right)+\frac{\gamma^H}{1-\gamma}+\epsilon/2,
    \end{aligned}$$
Thus for $t\in \gB$, 
$$ \sup_{y\in Y}\left[V_{c_y}^{(t)}(\mu)-u_y\right]\leq 2\epsilon+\frac{1}{(1-\gamma)^{3/2}}\sqrt{\left\|\frac{\nu^*}{\nu_0}\right\|_\infty\epsilon_{bias}}.
$$
We complete the proof by noting that
$$
P(A\cap B)\geq 1-(1-P(A))-(1-P(A_2\cap B))\geq 1-2\delta.
$$
\endproof

\proof{Proof of Theorem~\ref{Theorem_PGA_SICPO}}
Conditioned on event $A=A_1\cap A_2$, setting $K_{sgd}, H, K_{eval}, T, \alpha, \eta$ as above yields
$$\begin{aligned}
    \eta&> \frac{4\log |\gA|}{\sqrt{T}(1-\gamma)}+\frac{2\beta W^2}{\sqrt{T}(1-\gamma)^3}\\
    &+\frac{4}{1-\gamma}\sqrt{\frac{1}{1-\gamma}\left\|\frac{\nu^*}{\nu_0}\right\|_\infty\left(\epsilon_{bias}+C\frac{(4L_\pi^2W+8L_\pi/(1-\gamma))^2}{(1-\gamma)^2\mu_F}\frac{\log(2T/\delta)}{K_{sgd}}+\frac{4\gamma^H}{1-\gamma}\left(\frac{1}{1-\gamma}+WL_\pi\right)\right)}.
    \end{aligned}
$$
Now we use Lemma~\ref{Lemma_reward_bound} to get
$$
    \begin{aligned}
        &\frac{1}{|\gB|}\sum_{t\in\gB}(V_r^*(\mu)-V_r^{(t)}(\mu)) \leq \frac{2\log|\gA|}{\sqrt{T}(1-\gamma)}+\frac{\beta W^2}{\sqrt{T}(1-\gamma)^3}\\
       &+\frac{2}{1-\gamma}\sqrt{\frac{1}{1-\gamma}\left\|\frac{\nu^*}{\nu_0}\right\|_\infty\left(\epsilon_{bias}+C\frac{(4L_\pi^2W+8L_\pi/(1-\gamma))^2}{(1-\gamma)^2\mu_F}\frac{\log(2T/\delta)}{K_{sgd}}+\frac{4\gamma^H}{1-\gamma}\left(\frac{1}{1-\gamma}+WL_\pi\right)\right)}\\
       &\leq \epsilon+\frac{1}{(1-\gamma)^{3/2}}\sqrt{\left\|\frac{\nu^*}{\nu_0}\right\|_\infty\epsilon_{bias}}.
    \end{aligned}
    $$
According to Lemma~\ref{Lemma_constraint_violation_PGA}, conditioned on event $A_2$, as long as we set $T_{PGA}$ as in the statement of the theorem, we have $\forall t\in\{1,...,T\}$
$$\begin{aligned}
    &\sup_{y\in Y} \left[V_{c_y}^{(t)}(\mu)-u_y\right]-\left[\widehat V_{c_{y^{(t)}}}^{(t)}(\mu)-u^{(t)}\right]\\
    &\leq \frac{1}{(1-\gamma)\sqrt{2K_{eval}}}\left(1+\sqrt{\log{(4T/\delta)}+m\log(4\mathrm{diam}(Y)L_y\sqrt{2K_{eval}}+\mathrm{diam}(Y))}\right)+\frac{\gamma^H}{1-\gamma}+\epsilon/2,
    \end{aligned}$$
Thus for $t\in \gB$, 
$$ \sup_{y\in Y}\left[V_{c_y}^{(t)}(\mu)-u_y\right]\leq 2\epsilon+\frac{1}{(1-\gamma)^{3/2}}\sqrt{\left\|\frac{\nu^*}{\nu_0}\right\|_\infty\epsilon_{bias}}.
$$
\endproof

\begin{lemma}\label{Lemma_WHP_Bound_Evaluation_Uniform_y}
    For any $\delta\in(0,1)$, with probability at least $1-\delta$,
    $$
    \sup_{y\in Y}\left|\widehat V_{c_y}^{(t)}(\mu)-V_{c_y}^{(t)}(\mu)\right|\leq \frac{1}{(1-\gamma)\sqrt{2K_{eval}}}\left(1+\sqrt{\log{(2/\delta)}+m\log(4\mathrm{diam}(Y)L_y\sqrt{2K_{eval}}+\mathrm{diam}(Y))}\right)+\frac{\gamma^H}{1-\gamma}
    $$
    as long as $L_y\sqrt{2K_{eval}}>1$.
\end{lemma}
\proof{Proof of Lemma~\ref{Lemma_WHP_Bound_Evaluation_Uniform_y}}
Assumption~\ref{Assumption_Lipschitz} implies
$$
\left|\widehat V_{c_y}^{(t)}(\mu)-\widetilde{V}_{c_y}^{(t)}(\mu)\right|-\left|\widehat V_{c_{y^\prime}}^{(t)}(\mu)-\widetilde{V}_{c_{y^\prime}}^{(t)}(\mu)\right|\leq \frac{2L_y}{1-\gamma}\|y-y^\prime\|_{\infty}.
$$
Let $N_{\epsilon}:=\{y_1,...,y_N\}$ be a $\epsilon$ cover w.r.t. $\|\cdot\|_{\infty}$ of $Y$, Example 5.8 in \cite{wainwright2019high}. We have
$$
\log N\leq m\log(2\mathrm{diam}(Y)/\epsilon+\mathrm{diam}(Y)).
$$
Combining Theorem~\ref{Theorem_Hoeffding_Inequality} and the arguments of union bound we may have that for any $\delta\in(0,1)$, with probability at least $1-\delta$,
 $$
    \sup_{y\in N_{\epsilon/2}}\left|\widehat V_{c_y}^{(t)}(\mu)-\widetilde{V}_{c_y}^{(t)}(\mu)\right|\leq \frac{1}{1-\gamma}\sqrt{\frac{\log(2N/\delta)}{2K_{eval}}}.
    $$
Then we may get
$$
\begin{aligned}
    \sup_{y\in Y}\left|\widehat V_{c_y}^{(t)}(\mu)-\widetilde{V}_{c_y}^{(t)}(\mu)\right|&\leq
    \frac{2L_y\epsilon}{1-\gamma}+
    \frac{1}{1-\gamma}\sqrt{\frac{\log(2/\delta)+m\log(2\mathrm{diam}(Y)/\epsilon+\mathrm{diam}(Y))}{2K_{eval}}}.
\end{aligned}
    $$
Here we set $\epsilon=\frac{1}{2L_y\sqrt{2K_{eval}}}$ and get
$$
    \sup_{y\in Y}\left|\widehat V_{c_y}^{(t)}(\mu)-\widetilde{V}_{c_y}^{(t)}(\mu)\right|\leq \frac{1}{(1-\gamma)\sqrt{2K_{eval}}}\left(1+\sqrt{\log{(2/\delta)}+m\log(4\mathrm{diam}(Y)L_y\sqrt{2K_{eval}}+\mathrm{diam}(Y))}\right).
$$
Since
$$
\sup_{y\in Y}\left|V_{c_y}^{(t)}(\mu) -\widetilde{V}_{c_y}^{(t)}(\mu)\right|\leq \frac{\gamma^H}{1-\gamma},
$$
we complete the proof.
\endproof
\section{Auxiliary Lemmas}

\begin{lemma}[Empirical Bernstein Inequality]
\label{Theorem_Empirical_Bernstein}
Suppose $n\geq 3$, $\{X_1,...,X_n\}$ be $n$ i.i.d. random variables with values in $[0,1]$. 
Let $\delta>0$. 
Then with probability at least $1-\delta$ we have
$$
\left|\EB X_1-\frac{\sum_{i=1}^n X_i}{n}\right|\leq\sqrt{\frac{2\mathbb{V}_n(X_{1:n})\log4/\delta}{n}}+\frac{4\log 4/\delta}{n},
$$
where $\mathbb{V}_n(X_{1:n}):=\frac{1}{n(n-1)}\sum_{i,j}\frac{(X_i-X_j)^2}{2}$ denotes the empirical variance of the dataset $\{X_1,...,X_n\}$.
\end{lemma}
\proof{Proof of Lemma~\ref{Theorem_Empirical_Bernstein}.}
See Theorem 11 in \cite{maurer2009empirical}.
\endproof

\begin{lemma}[Hoeffding's Inequality]
\label{Theorem_Hoeffding_Inequality}
Suppose $\{X_1,...,X_n\}$ be $n$ i.i.d. random variables with values in $[0,1]$.
Let $\delta>0$. 
Then with probability at least $1-\delta$ we have
$$
\left|\EB X_1-\frac{\sum_{i=1}^n X_i}{n}\right|\leq\sqrt{\frac{\log 2/\delta}{2n}}
$$
\end{lemma}
\proof{Proof of Lemma~\ref{Theorem_Hoeffding_Inequality}.}
See Theorem 2.2.6 in \cite{vershynin_2018}.
\endproof

\begin{lemma}\label{Lemma_Simulation_Lemma}
For any policy $\pi$ and transition probabilities $P$, $\widetilde{P}$, we have that
$$Q^{\pi}-\widetilde{Q}^{\pi}=\gamma\left(I-\gamma \widetilde{P}^{\pi}\right)^{-1}(P-\widetilde{P}) V^{\pi}
$$
\end{lemma}
\proof{Proof of Lemma~\ref{Lemma_Simulation_Lemma}.}
See Lemma 2 in \cite{pmlr-v125-agarwal20b}.
\endproof

\begin{lemma}\label{Lemma_Norm_of_Inf_Horizon_Expectation}
For any policy $\pi$, any transition probability $P$ and any vector $v\in \RB^{|\gS|\cdot|\gA|}$, we have 
$$\left\|\left(I-\gamma P^{\pi}\right)^{-1} v\right\|_{\infty} \leq\|v\|_{\infty} /(1-\gamma).
$$
\end{lemma}
\proof{Proof of Lemma~\ref{Lemma_Norm_of_Inf_Horizon_Expectation}.}
See Lemma 3 in \cite{pmlr-v125-agarwal20b}.
\endproof

\begin{lemma}\label{Lemma_Bound_of_Weighted_Variance}
For any policy $\pi$ and any transition probability $P$, we have
$$\left\|(I-\gamma P^\pi)^{-1} \sqrt{\Var_P^\pi}\right\|_\infty\leq\sqrt{\frac{2}{(1-\gamma)^3}},
$$
where $\sqrt{\cdot}$ is defined as the element-wise square root.
\end{lemma}
\proof{Proof of Lemma~\ref{Lemma_Bound_of_Weighted_Variance}.}
See Lemma 4 in \cite{pmlr-v125-agarwal20b}.
\endproof

\begin{theorem}[Chernoff's Inequality]
\label{Theorem_Chernoff_Inequality}
Let $X_i$ be independent Bernoulli random variables with parameter $p_i$. 
Consider their sum $S_N=\sum_{i=1}^N X_i$ and denote its mean by $\mu=\EB S_N$. 
Then, for any $t<\mu$, we have
$$
\PB\paren{S_N<t}\leq e^{-\mu}\paren{\frac{e\mu}{t}}^t.
$$
\end{theorem}
\proof{Proof of Theorem~\ref{Theorem_Chernoff_Inequality}.}
See \cite{vershynin_2018}.
\endproof
\section{Experimental Setup}
\label{Appendix_Detials_of_Experiments}
We use PyTorch, RLlib, and Gurobi framework to implement the algorithms in our work, and the codes are run on Nvidia RTX Titan GPUs and Intel Xeon Gold 6132 CPUs with 252GB memory. The code of our algorithms and construction of the corresponding environments has been released on
\href{https://github.com/pengyang7881187/SICMDP-new}{https://github.com/pengyang7881187/SICMDP-new}.




\subsection{Details of Construction of Discharge of Sewage}
We generate this environment randomly with $|\gS|=8$, $|\gA|=4$ and $\gamma=0.9$. In each experiment, we sample the environments several times and report the average result.

Positions of sewage outfalls, transition dynamics and rewards are sampled uniformly on $Y$, probability simplex and $[0, 1]$ respectively. The state-occupancy measure $d$ is then generated by the uniform policy and $\Delta$ is set to $10^{-6}$.

The optimal policy $\pi^*$ used is obtained by solving a corresponding linear programming with true transition dynamic $P$ and a fine grid of $Y$ of size $10^6$. 



\subsection{Implementation of SI-CRL}
We would like to clarify how to solve the maximization problem to generate new $y_t$ in $t$th iteration in SI-CRL. Since the $u_y$ can be non-convex in $y$ and evaluating $\widehat V^\pi_{c_y}-u_y$ for multiple $y$ is much cheaper in the model-based setting with small $|\gS|$ and $|\gA|$ than solving the linear programming, we choose to solve the maximization problem by brute force.
Specifically, we first create a grid of $Y$ of size $10^5$, and then find the grid point with max objective.
This method works well since in the problems we consider $Y$ is of low dimensions.

When solving the linear programming, we do not force Gurobi to use dual simplex method in SI-CRL algorithm, and find that it achieves an even better re-optimization performance in practice. 

\subsection{Implementation of SI-CPO}
The parameterized policy class is chosen as the softmax policy: $\pi_\theta(a|s)=e^{\theta_{s,a}}/\sum_{a^\prime\in \gA}e^{\theta_{s,a^\prime}}$, where $\theta\in\RB^{|\gS||\gA|}$ is initialized as $0$. This policy class satisfies the assumptions of the theoretical analysis of SI-CPO.

We use learning rate $\alpha=1$, tolerance $\eta=0.013$, maximum iteration
number $T=10000$ for both SI-CPO and baseline. SI-CPO and baseline use sample-based NPG, Algorithm \ref{Algorithm_sample_based_NPG}, as the policy optimization subroutine, and share the same hyper-parameters: number of evaluation paths $K_{eval}=100000$, number of training paths $K_{sgd}=1000$, fixed horizon $H=100$, upper bound of parameters' norm $W=1000$, constant learning rate $1$ and weight $\gamma_k=2k/K_{sgd}(K_{sgd}+1)$.

In $t$th iteration in SI-CPO, we sample $100$ points uniformly in $Y$ and find the best one to solve the maximization problem approximately as Algorithm \ref{Algorithm_random_search}. In the model-free setting, evaluating $\widehat V^\pi_{c_y}$ is much more computationally expensive than the model-based setting, hence brute force is impractical. Additionally, the random search method can yield a better policy than the one utilizing constraint optimization algorithm in practice. 

\subsection{Details of Construction of Ship Route Planning}
We fix the outset $O=(0, 0)$, destination $D=(1, 1)$, environmentally critical point $MPA=(\frac{1}{2}, \frac{1}{2})$ and we assume the ship sails with constant speed $0.1$. $c_y$ and $u_y$ are designed to make sure the trajectory along the curve $y=x^4$ satisfies the constraint.

\subsection{Implementation of SI-CPPO}
The actor, constraint critic and reward critic networks have two hidden layers of size 512 with tanh non-linearities without sharing layers for both SI-CPPO and baseline. We update the networks using the Adam optimizer with learning rate $10^{-4}$. SI-CPPO and baseline are modified from the standard implementation of PPO in RLlib and share the same hyper-parameters. See more details in Algorithm \ref{Algorithm_SICPPO}, note that we use the generalized advantage estimation in practice instead of the one-step estimation in the pseudo-code.

In $t$th iteration in SI-CPPO, we use a trust-region method to solve the maximization problem based on the trajectories collected by $4$ roll-out workers. We call scipy.optimize.minimize to implement it, if the optimization process fails to converge, we switch to use random search with $100$ points uniformly in $Y$ temporarily. Empirically, the optimization process seldom fails and this strategy outperforms random search.

\begin{algorithm}[htb!]
   \caption{SI-CPPO}
   \label{Algorithm_SICPPO}
\begin{algorithmic}
   \STATE {\bfseries Input:} state space $\gS$, action space $\gA$, reward function $r$, a continuum of cost function $c$, index set $Y$, value for constraints $u$, discount factor $\gamma$, batch size $B$, tolerance $\eta$, maximum iteration number $T$.
   \STATE Initialize policy network $\pi^{(0)}$, reward critic network $V_r^{(0)}$ and constraint critic network $V^{(0)}_c$.
   \FOR{$t=0,...,T-1$}
   \STATE Sample $B$ trajectories $\gB^{(t)}$ using policy network $\pi^{(t)}$.
   \STATE Obtain Monte-Carlo estimator $\widehat V_{c_y}^{\pi^{(t)}}(\mu)$ based on $\gB^{(t)}$.
   \STATE Use an optimization subroutine to solve ${\max_y\ \widehat V_{c_y}^{\pi^{(t)}}(\mu)-u_y}$, and set ${y^{(t)}\approx\argmax_y \widehat V_{c_y}^{\pi^{(t)}}(\mu)-u_y}$, $c^{(t)}=c_{y^{(t)}}$.
   \IF {$\widehat V_{c^{(t)}}^{\pi^{(t)}}(\mu)-u_{y^{(t)}}\leq \eta$}
   \STATE  Obtain advantage estimation of reward using reward critic network: 
   
   $\hat{A}^{(t)}(s_\tau,a_\tau)=\left(r_\tau+\gamma V_r^{(t)}(s_{\tau+1}) \right)-V_r^{(t)}(s_{\tau})$, $\forall (s_\tau, a_\tau, r_\tau, s_{\tau+1})\in \gB^{(t)}$.
   \ELSE 
   \STATE  Obtain advantage estimation of constraint at $y^{(t)}$ using constraint critic network: 
   
   $\hat{A}^{(t)}(s_\tau,a_\tau)=V^{(t)}_{c^{(t)}}(s_{\tau})-\left(c_{y^{(t)},\tau}+\gamma V^{(t)}_{c^{(t)}}(s_{\tau+1}) \right)$,  $\forall(s_\tau, a_\tau, r_\tau, s_{\tau+1})\in \gB^{(t)}$.
   \ENDIF
   \STATE Update policy network $\pi^{(t)}$, reward critic network $V_r^{(t)}$ and constraint critic network $V^{(t)}_c$ to $\pi^{(t+1)}$, $V_r^{(t+1)}$ and $V^{(t+1)}_c$ respectively with the above advantage estimation via the standard proximal policy optimization algorithm.
   \ENDFOR
   \STATE {\bfseries RETURN} $\hat\pi=\pi^{(T)}$.
\end{algorithmic}
\end{algorithm}



\section{Omitted Algorithms}\label{Appendix_Algorithm}

\begin{algorithm}[htb!]
   \caption{Random Search}
   \label{Algorithm_random_search}
\begin{algorithmic}
   \STATE {\bfseries Input:} Objective function $f\colon Y\to\RB$, where $Y$ is a compact subset of $\RB^m$.
   \STATE Sample $y_1,...,y_M\stackrel{i.i.d.}{\sim}\mathrm{Unif}(Y)$.
   \STATE {\bfseries RETURN} $\hat y=y_{i_0}$, $i_0=\argmax_{i\in\{1,...,M\}}f(y_i)$.
\end{algorithmic}
\end{algorithm}

\begin{algorithm}[htb!]
   \caption{Projected Subgradient Ascent}
   \label{Algorithm_projected_GD}
\begin{algorithmic}
   \STATE {\bfseries Input:} Objective function $f\colon Y\to\RB$, where $Y$ is a compact subset of $\RB^m$, the maximum number of iterations $T_{PGA}$.
   \STATE Initialize: set $y_0$ as an arbitrary element of $Y$, learning rate $\alpha=\frac{\mathrm{diam}(Y)}{L_y\sqrt{T}}$.
   \FOR {$t=0,...,T_{PGA}-1$}
   \STATE $t_{t+0.5}=y_t+\alpha g_t$, where $g_t$ is a subgradient of $f$ at $y_t$.
   \STATE $y_{t+1}=\argmin_{y\in Y} \|y-y_{t+0.5}\|$.
   \ENDFOR
   \STATE {\bfseries RETURN} $\hat y=\frac{1}{T_{PGA}}\sum_{t=1}^{T_{PGA}} y_t$.
\end{algorithmic}
\end{algorithm}

\begin{algorithm}[htb!]
   \caption{Sample-based NPG}
   \label{Algorithm_sample_based_NPG}
\begin{algorithmic}
   \STATE {\bfseries Input:} state space $\gS$, action space $\gA$, a criterion function $b$ (Can be the reward function $r$ or cost function $c_y$ for some fixed $y$), discount factor $\gamma$, policy $\pi_\theta$, number of paths $K_{sgd}$, fixed horizon $H$, upper bound of parameters' norm $W$, learning rate $\{\eta_k\}$, weight $\{\gamma_k\}$
   \FOR{$k=0$ {\bfseries to} $K_{sgd}-1$}
      \STATE Draw $(s,a)\sim \nu$, with $\nu(s,a)=d^{\pi_\theta}(s)\pi_\theta(a|s)$.
   \STATE Execute policy $\pi_\theta$ from $(s,a)$ for $H$ steps, then construct the estimators as
   $$
   \begin{aligned}
   \widehat Q^{\pi_\theta}(s,a)=\sum^{H-1}_{k=0} \gamma^k b(s_k,a_k),\ \text{where } (s_0,a_0)=(s,a).
   \end{aligned}
   $$
   \STATE Execute policy $\pi_\theta$ from $s$ for $H$ steps, then construct the estimators as
   $$
   \widehat V^{\pi_\theta}(s,a)=\sum^{H-1}_{k=0} \gamma^k b(s_k,a_k),\ \text{where } s_0=s.
   $$
   \STATE Set $\widehat A^{\pi_\theta}(s,a)=\widehat Q^{\pi_\theta}(s,a)-\widehat V^{\pi_\theta}(s)$.
   \STATE Perform an iteration of projected SGD: $w^{(k+1)}=\operatorname{Proj}_{B(0,W,\|\cdot\|_2)}(w^{(k)}-\eta_k G^{(k)})$ with
   $$
   \begin{aligned}
      G^{(k)}&=2({w^{(k)}}^\top\nabla_\theta\log\pi_\theta(a|s)-\widehat A^{\pi_\theta}(s,a))\nabla_\theta\log\pi_\theta(a|s),\\
   \end{aligned}
   $$
   and $B(0,W,\|\cdot\|_2):=\{w\in\RB^d|\|w\|_2\leq W\}$.
   \ENDFOR
   \STATE {\bfseries RETURN} $\sum_{k=1}^K \gamma_k w^{(k)}$ as a NPG update direction at $\pi_\theta$ w.r.t. criterion function $b$.
\end{algorithmic}
\end{algorithm}

\end{document}